\documentclass[pdflatex,sn-mathphys-num]{sn-jnl}% Math and Physical Sciences Numbered Reference Style
\usepackage{graphicx}%
\usepackage{multirow}%
\usepackage{amsmath,amssymb,amsfonts}%
\usepackage{amsthm}%
\usepackage{mathrsfs}%
\usepackage[title]{appendix}%
\usepackage{xcolor}%
\usepackage{textcomp}%
\usepackage{manyfoot}%
\usepackage{booktabs}%
\usepackage{algorithm}%
\usepackage{algorithmicx}%
\usepackage{algpseudocode}%
\usepackage{listings}%
\usepackage{caption}
\usepackage{multirow}
\theoremstyle{thmstyleone}%
\theoremstyle{thmstyletwo}%

\theoremstyle{thmstylethree}%

\begin{document}

%\title[Article Title]{AI-Driven Diabetic Retinopathy Diagnosis Enhancement through Image Processing and Salp Swarm Algorithm-Optimized Ensemble Network}

\title[Article Title]{Temperature-Driven Robust Disease Detection in Brain and Gastrointestinal Disorders via Context-Aware Adaptive Knowledge Distillation}


\author*[1,3]{\fnm{Saif Ur Rehman} \sur{Khan}}\email{saif\_ur\_rehman.khan@dfki.de}

\author*[1,2]{\fnm{Muhammad Nabeel} \sur{Asim}}\email{muhammad\_nabeel.asim@dfki.de}
\author[1,2]{\fnm{Sebastian} \sur{ Vollmer}}\email{sebastian.vollmer@dfki.de}
\author[1,2,3]{\fnm{Andreas} \sur{Dengel}}\email{andreas.dengel@dfki.de}

\affil[1]{\orgdiv{German Research Center for Artificial Intelligence}, \orgaddress{ \city{Kaiserslautern}, \postcode{67663}, \country{Germany}}}
\affil[2]{\orgdiv{Intelligentx GmbH (intelligentx.com)}, \orgaddress{ \city{Kaiserslautern}, \country{Germany}}}
\affil[3]{\orgdiv{Department of Computer Science}, \orgname{Rhineland-Palatinate Technical University of Kaiserslautern-Landau} \orgaddress{ \city{Kaiserslautern}, \postcode{67663}, \country{Germany}}}
%%\textcolor{red}{easily} 
\abstract{Medical disease prediction, particularly through imaging, remains a challenging task due to the complexity and variability of medical data, including noise, ambiguity, and differing image quality. Recent deep learning models, including Knowledge Distillation (KD) methods, have shown promising results in brain tumor image identification but still face limitations in handling uncertainty and generalizing across diverse medical conditions. Traditional KD methods often rely on a context-unaware temperature parameter to soften teacher model predictions, which does not adapt effectively to varying uncertainty levels present in medical images. To address this issue, we propose a novel framework that integrates Ant Colony Optimization (ACO) for optimal teacher-student model selection and a novel context-aware predictor approach for temperature scaling. The proposed context-aware framework adjusts the temperature based on factors such as image quality, disease complexity, and teacher model confidence, allowing for more robust knowledge transfer. Additionally, ACO is preferred for teacher-student model selection due to its efficiency in exploring complex solution spaces. It outperforms PSO and Grid Search by achieving 96.33\% accuracy with only 47 evaluations, demonstrating faster convergence and lower computational cost. The proposed framework is evaluated using three publicly available benchmark datasets, each corresponding to a distinct medical imaging task. The results demonstrate that the proposed framework significantly outperforms current state-of-the-art methods, achieving top accuracy rates: 98.01\% on the MRI brain tumor (Kaggle) dataset, 92.81\% on the Figshare MRI dataset, and 96.20\% on the GastroNet dataset. This enhanced performance represents a percentage increase of 0.77\% over the existing benchmark of 97.24\% (Kaggle), 1.38\% over 91.43\% (Figshare), and 1.20\% over 95.00\% (GastroNet).}


\maketitle
\section{Introduction}\label{sec1}
Classifying multiple medical conditions through diverse imaging modalities involves significant complexity, and problems in medical field. A tumor is a collection of cells that are dividing out of control, and thus can be non-cancerous or cancerous. A brain tumor is a growth of irregular cells that develop within the brain and disrupt the normal functions of brain \cite{boutry2022evolution,khan2025optimizing}. The brain is one of the most complex organs with complex functions, controlling the nervous system and organizing billions of cells. Brain tumors pose a substantial health risk globally because they can disrupt crucial body functions and life-threatening disorders can occur if not diagnosed early and overlooked. \cite{alemany2021late,hekmat2025differential}. The accurate detection of these tumors is crucial for improving survival rates of patients. Tumors are classified as primary, which originate in the brain, or secondary tumor that originate from other parts of the body and spread towards the brain by blood flow \cite{rathi2022influence}. The common symptoms that cause these tumors are seizures , headaches problem, aphasia, vision changes, dizziness, weakness, and mental complications depending on their size and location \cite{lee2021neurologic}. There are different categories of brain tumors but the most frequently occurring types are meningiomas, gliomas, and pituitary gland tumors. Malignant tumors consist of 29.7\% cases, while non-malignant types are consisting of 70.3\% cases \cite{ostrom2018epidemiology}. Meningiomas and gliomas are the most fatal tumors because gliomas are very severe with a low survival rate, representing 45\% of all brain tumors, while pituitary tumors and meningiomas representing only 15\% for each \cite{swati2019content}. This makes brain tumors as one of the extremely harmful type of cancer. The brain is the eighth and utmost important organ of the body. It faces major health challenge due to brain tumors, which have the highest cancer death rate worldwide. The origin and growth rate of these tumors remain an unidentified \cite{ganesh2024multi}.
Analysis of tumors is challenging because of their varying sizes, locations, and appearance in clinical settings. Accurate classification of tumors is critical for diagnosis but conventional imaging methods such as CT, MRI, and PET images can be time-consuming and prone to errors \cite{jyothi2023deep}. These methods depend on radiologists’ expertise who review the images manually to find any abnormalities. However, conventional methods for spotting brain tumors come with their own challenges, such as how different people interpret these images and the difficulty of infectious early and subtle signs of a tumor spread. 
To overcome these challenges, transfer learning (TL) which is a sub-field of machine learning (ML) has been employed to improve the early diagnosis and classification of brain tumors \cite{malakouti2024machine}. The task of classifying brain tumors is difficult because of the shape and contrast variations. Recent advances in the field of artificial intelligence (AI), especially convolutional neural networks (CNNs) with deep learning (DL), have improved tumor diagnostic accuracy by automatically extracting features from medical images \cite{saboor2024ddfc}. However, training these models demands large datasets and extensive time, making transfer learning a viable solution with pre-trained models such as ImageNet \cite{rajput2024transfer}.To further enhance the performance, feature fusion technique is employed which combines different features to enhance the  performance of model, but it often lacks due to the varying scales and complexities of the features \cite{salih2024fusion,ag2024robust}. Multiscale feature fusion addresses this by integrating features at multiple scales by capturing both fine and coarse details. This approach improves the accuracy of the model by increasing the overall representation of the data, taking advantage of a larger set of information \cite{zhou2024research}.

\subsection{Problem formalization}
Medical image classification, particularly for disease prediction, is hindered by several challenges \cite{khan2025optimized} , including noise, ambiguity, and varying image quality, which complicate accurate diagnosis. Traditional deep learning models, such as KD, have made strides in improving model performance by transferring knowledge from a teacher model to a student model. In traditional KD methods, the teacher model’s predictions are softened using a temperature scaling parameter \( \text{T} \) to make the output probabilities more informative for the student model. This scaling helps to smooth the logits, providing the student with a softened version of the teacher’s hard predictions. However, this approach assumes that the uncertainty across all data samples is uniform, which does not reflect the actual variability in real-world data, particularly in the case of medical images. The need for robustness to uncertainty and reproducible behavior under variable imaging conditions aligns with prior work on capsule-enhanced neural networks and reproducibility \cite{zhang2024cenn}.

\bmhead{Example Medical: MRI Brain Tumor Detection}
MRI images used for brain tumor detection can vary significantly in terms of quality. A high-quality MRI scan with a clear distinction between tumor and surrounding tissue might present very little uncertainty for the model. However, a low-quality MRI scan with motion artifacts, distortion, or noise might introduce substantial uncertainty, making it difficult for the model to accurately classify the tumor or predict its type. In a traditional KD approach with a static temperature, the KD process applies a uniform softening of the class probabilities across all images, regardless of their quality. This approach treats both high-quality and noisy MRI images the same, which is not ideal since the noisy images may require more robust learning mechanisms to deal with the uncertainty.

In contrast, a our dynamic temperature approach could adjust the temperature based on the uncertainty level of the MRI scan. For high-quality MRI images, the temperature could be adjust lower to preserve the sharp distinctions between tumor and non-tumor tissues. However, for lower-quality or noisy images, the temperature could be increased to allow the model to learn from a broader range of class probabilities and make better-informed decisions in the presence of uncertainty. This dynamic adjustment enhances the model ability to handle variability in image quality, leading to more accurate tumor detection, even in challenging scenarios with noise or artifacts.

To illustrate (Fig 1), let consider the task of training a model on brain cancer detection. The teacher model may confidently detect tumors \cite{habib2024comprehensive} in clear MRI with high certainty, while the predictions for early-stage tumors or scans with artifacts will carry greater uncertainty. Using the same degree of softening for both types of predictions (confident) and uncertain (limit) the student model ability to learn from the more challenging cases, as the information provided by the teacher in uncertain areas may not be appropriately weighted or represented.
Mathematically, the traditional KD loss function can be expressed as:
\[
L_{\text{KD}} = \alpha \cdot L_{\text{CE}}(y, \hat{y}) + \beta \cdot T^2 \cdot D_{\text{KL}}(q_{\text{teacher}}(T), q_{\text{student}}(T))
\]

Where \( L_{\text{CE}} \) is the cross-entropy loss, \( D_{\text{KL}} \) is the Kullback-Leibler divergence, and \( T \) controls the amount of smoothing applied. However, the application of the same level of smoothing across all predictions ignores the natural variability in uncertainty, which reduces the effectiveness of distilling complex knowledge from the teacher model, especially for uncertain and ambiguous cases.
\begin{figure}[h]
    \centering
    \includegraphics[width = 12cm]{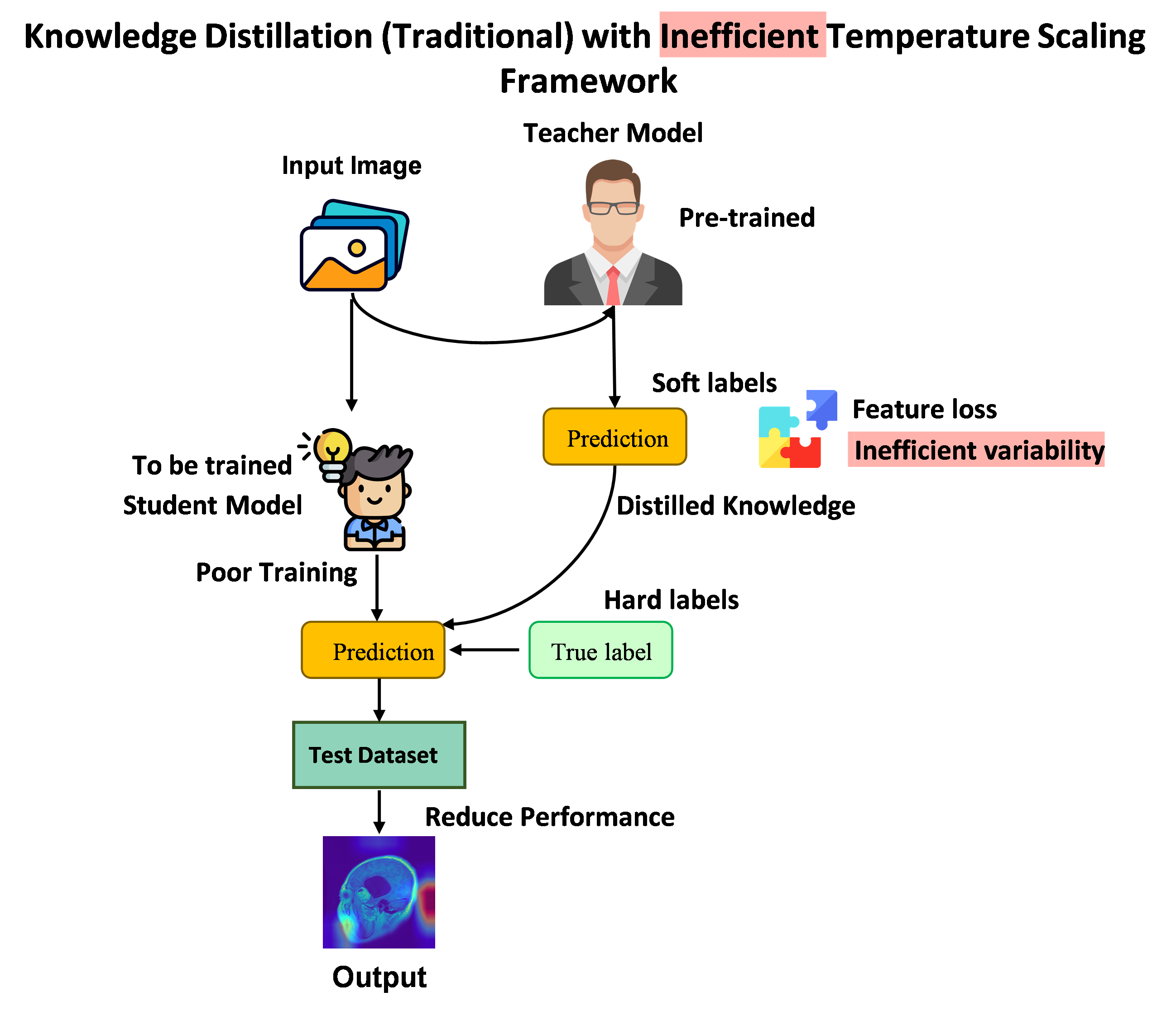}
    \caption{Visualization of Inefficient Temperature Scaling in Traditional KD Framework}
    \label{fig:se.png}
\end{figure}
Main contribution of this work as follows:
\begin{itemize}
   \item \textbf{Enhancing Adaptability to Uncertainty:} Traditional KD methods apply a uniform scaling factor, which limits their ability to adapt to the varying uncertainty present in medical images. Our novel context-aware predictor system dynamically adjusts this scaling mechanism, allowing for improved handling of noisy and ambiguous data, leading to enhanced performance.

   \item \textbf{Context-Aware Temperature Adjustment Mechanism:} The novel context-aware prediction framework enhances knowledge transfer by adapting the temperature value based on dynamic factors such as image quality, disease complexity, and teacher model confidence. This adaptive mechanism fine-tunes the learning process for each instance, enabling the model to better capture subtle patterns and variations in the data, thereby improving the overall performance and robustness of the system.

   \item \textbf{Handling Complex and Diverse Medical Datasets:} Our approach is tested on multiple medical datasets, including Figshare MRI, and gastro data, ensuring robustness. Conventional KD methods apply uniform smoothing, which fails to account in medical images. We resolve this by vigorously scaling the transfer process, enhancing accuracy and generalization, particularly for uncertain regions.
    
    \item \textbf{Improvement in Model Performance:} The Adaptive Knowledge-Driven Student model achieves exceptional performance, with 98.01\% accuracy on the Kaggle MRI brain tumor dataset, 92.81\% on the Figshare MRI dataset, and 96.20\% on the GastroNet dataset. These results surpass the existing benchmarks of 97.24\% (Kaggle), 91.43\% (Figshare), and 95.00\% (GastroNet). The improvement is driven by Context-Aware Rule-Based temperature scaling, enhancing both accuracy and generalization.
    
\end{itemize}

\section{Related work}\label{sec2}
The field of DL in image processing is advancing more rapidly to involve tasks such as image segmentation, classification and interpretation. Traditional methods for the classification of medical image such as Kidney Stone \cite{khan2025multi,khan2025ai} are based on the detail of color, texture, and shape of images. Baloch et al \cite{baloch2007flexible} proposed a versatile skew-symmetric shape model to capture hidden variation within certain neighborhoods. Meanwhile, Koitka et al. \cite{koitka2016traditional} employed manually extracted visual features to classify medical images. A drawback of these conventional methods is their dependence on manual feature extraction, which can be laborious and not able to capture full complexity of image variations. The manual inspection of medical images is frequently prone to human-errors and time-consuming task, specifically with increasing number of cases \cite{shahzad2024enhancing}.The evolution of CNNs with DL addresses these challenges by automatically extracting more intricate features from MRI images \cite{singha2021deep,ghiasi2022simple}. The operation of CNNs is highly effective at identifying intricate details and subtle patterns that are difficult for human observers to distinguish. Their application in TL enhances image processing tasks such as image enhancement, augmentation, denoising, and classification to significantly advance accurate and efficient tumor classification and diagnostic accuracy \cite{archana2024deep}.
Recent studies have investigated DL feature fusion methods for brain tumor classification using MRI images. These methods normally include pre-processing steps such as normalization of data and augmenting the data, followed by feature extraction using pre-trained CNNs such as ResNet, AlexNet, and GoogleNet. Zahid et al. \cite{zahid2022brainnet} introduce an automated method for brain tumor multi-classification task using normalized MRI data by employing a fine-tuned ResNet101 model. They address redundant features through differential evaluation, particle swarm optimization, and PCA. This approach significantly enhancing efficiency and performance compared to earlier methods by achieving an accuracy of 94.4\%. Kibriya et al. \cite{kibriya2022novel} proposed a multiclassification method by employing deep feature fusion with MRI images. The features obtained from AlexNet, GoogLeNet, and ResNet18 are merged into a single vector, which is then classified by SVM and KNN classifiers. Although this method achieves remarkable accuracy, but it increases the computational cost due to the integration of multiple DL architectures. Öksüz et al. \cite{oksuz2022brain} present brain tumor classification method that combines both the deep and shallow features to identify different types of brain tumors. They use pre-trained networks like AlexNet, ResNet-18, GoogLeNet, and ShuffleNet to extract features and a shallow network for extracting low-level features. Fusing these features and classifying with SVM and k-NN improves the sensitivity by 11.72\%, though the method's complexity and computational costs are significant drawbacks. Khan et al. \cite{khan2023deep} propose a multimodal brain tumor classification method using two-way DL and hybrid feature optimization. They fine-tune NasNet-Mobile on original and enhanced MRI images. They subsequently employ Haze-CNN for contrast enhancement and then fuse features through multiset canonical correlation analysis. Despite the increased computational cost, this method achieved 94.8\% and 95.7\% accuracy on the BraTs2018 and BraTs2019 datasets, respectively. Agarwal et al. \cite{agrawal2024multifenet} introduced a MultiFeNet model using multiscale feature scaling for brain tumor classification evaluated on 3,064 MRI images. The model achieved 96.4\% accuracy on the Figshare dataset. Despite its high accuracy, the method faces challenges due to its complexity and computational requirements. Liu et al. \cite{liu2024multi} propose MultiGeneNet, which is a multi-scale fusion network for predicting IDH1 gene mutations in gliomas. By using two feature extractors and bilinear pooling, the authors achieved 83.57\% accuracy on 296 pathological sections. This approach outperforms single-scale methods and offers a more accurate, cost-effective diagnostic solution. Shahin et al.\cite{shahin2023mbtfcn} developed a modular deep fully CNN method to detect tumor from MRI images. The network incorporates modules for feature extraction, residual strip pooling attention etc. This method has tested on 9,581 images from four public datasets, demonstrating superior performance but still faces challenges with large intra-class variations and small datasets. Sobhaninia et al.\cite{sobhaninia2023brain} proposed a Multiscale Cascaded Multitask Network to segment and classify brain tumors in MRI images using multitask learning with a feature aggregation module. The method achieved 97.988\% accuracy in classification but still faces challenges due to the complexity of tumor shapes and locations. Table ~\ref{T0} below outlines recent methods proposed by prior works in the literature.
\begin{table}[ht]
\centering
\caption{Comparison of distillation methods and their limitations}
\label{T0}
\renewcommand{\arraystretch}{1.3}
\begin{tabular}{cp{5.2cm}p{6.5cm}}
\hline
\textbf{Reference} & \textbf{Approach} & \textbf{Limitation} \\
\hline
R. Miles et al \cite{miles2024understanding} , 2024 &
Prior to distillation, the logits are standardized. &
This approach treats both high-confidence and unclear disease locations uniformly since it ignores the different levels of data uncertainty. \\
\hline
M. Yuan et al, \cite{yuan2024student} 2024 &
Adds noise to the logits to smooth them out for improved uncertainty learning. &
Noise can lower the quality of distillation by obscuring crucial information, particularly in complicated diseases. \\
\hline
Y. Ding et al, \cite{ding2024generous} 2024 &
Divides the loss from logit distillation into several phrases. &
Model performance may suffer if high-confidence areas are not given enough priority when the loss terms are split. \\
\hline
Z. Hao et al, \cite{hao2023revisit} 2023 &
Lessens the distance between students and teachers &
Different levels of feature uncertainty are not accommodated by fixed temperature solutions for gap reduction. \\
\hline
\end{tabular}
\end{table}
\section{Methods and materials}\label{sec3}
In this section, we will explain the selection of the teacher-student model using the ACO algorithm, which is key to identifying the optimal teacher-student pair for efficient knowledge transfer. Following this, we will provide an architectural overview of the selected model, highlighting its core components and their interactions in the KD process. We will also address the challenges associated with traditional knowledge distillation approaches, where the temperature scaling mechanism remains constant, often resulting in suboptimal performance. To overcome this, we propose an adaptive temperature adjustment mechanism, leveraging context-aware rules to dynamically modify the temperature during the distillation process. This approach allows for a more flexible and effective transfer of knowledge, enhancing model performance by addressing the limitations of static temperature configurations.
\subsection{Dataset Description}\label{subsec2}
It is crucial to anticipate brain cancers accurately since it may allow for early detection and treatment. Negative effects, such as increased tumor growth and metastasis, can arise from improperly diagnosing brain tumors, which can ultimately lead to more serious health issues and fewer treatment options. In order to do this significant task, we used a dataset of 7023 MRI scans that was taken from the publicly accessible Kaggle repository \cite{khan2025detection}. The MRI in this dataset fall into four different categories such as 2000 MRI related to normal brain, 1645 MRI belong to meningioma tumor sample, 1757 MRI samples shows pituitary tumor, and 1621 MRI associated to glioma brain samples. With a common resolution of 512 x 512 pixels, Fig 2 provides an example of brain MR pictures to give an idea of the dataset. The dataset distribution consists of 80\% of the MRI images being used for training, 10\% allocated for validation, and 20\% reserved for evaluating the model performance.
\begin{figure}[h]
    \centering
    \includegraphics[width = 12cm]{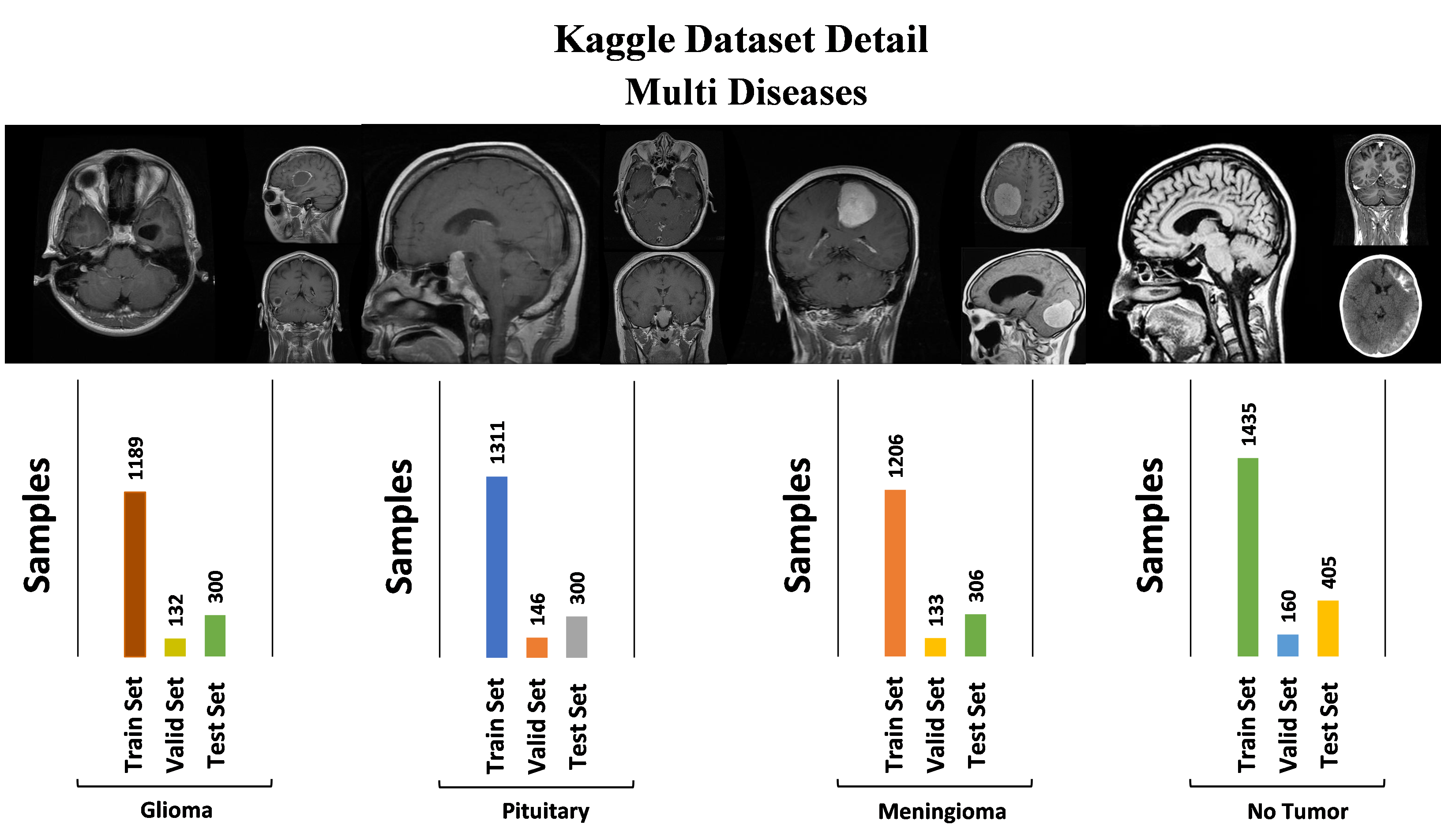}
    \caption{Kaggle multiclass dataset sample visualization}
    \label{fig:se.png}
\end{figure}
\subsection{Ant Colony-Based Selection of Teacher and Student Models from Pool of Pre-trained Networks}\label{subsubsec2}
The ACO \cite{wu2024application} has used for teacher and student model selection for knowledge distillation offers the gain of intelligently exploring and balancing the trade-offs between different models, leveraging pheromone-guided exploration to find the best combination. the simpler approach of running all models (sixteen) one by one and selecting the top two, ACO adapts dynamically over iterations, improving the search process by considering both model performance and past successes, leading to a more efficient and optimal model selection. This method helps conserve computational resources and enhances model selection accuracy, particularly when handling a large number of models, as demonstrated in our case where we utilized 16 pre-trained models initially.
Fig 3 illustrate the ACO approach of selection of teacher and student models.
\begin{figure}[h]
    \centering
    \includegraphics[width = 12cm]{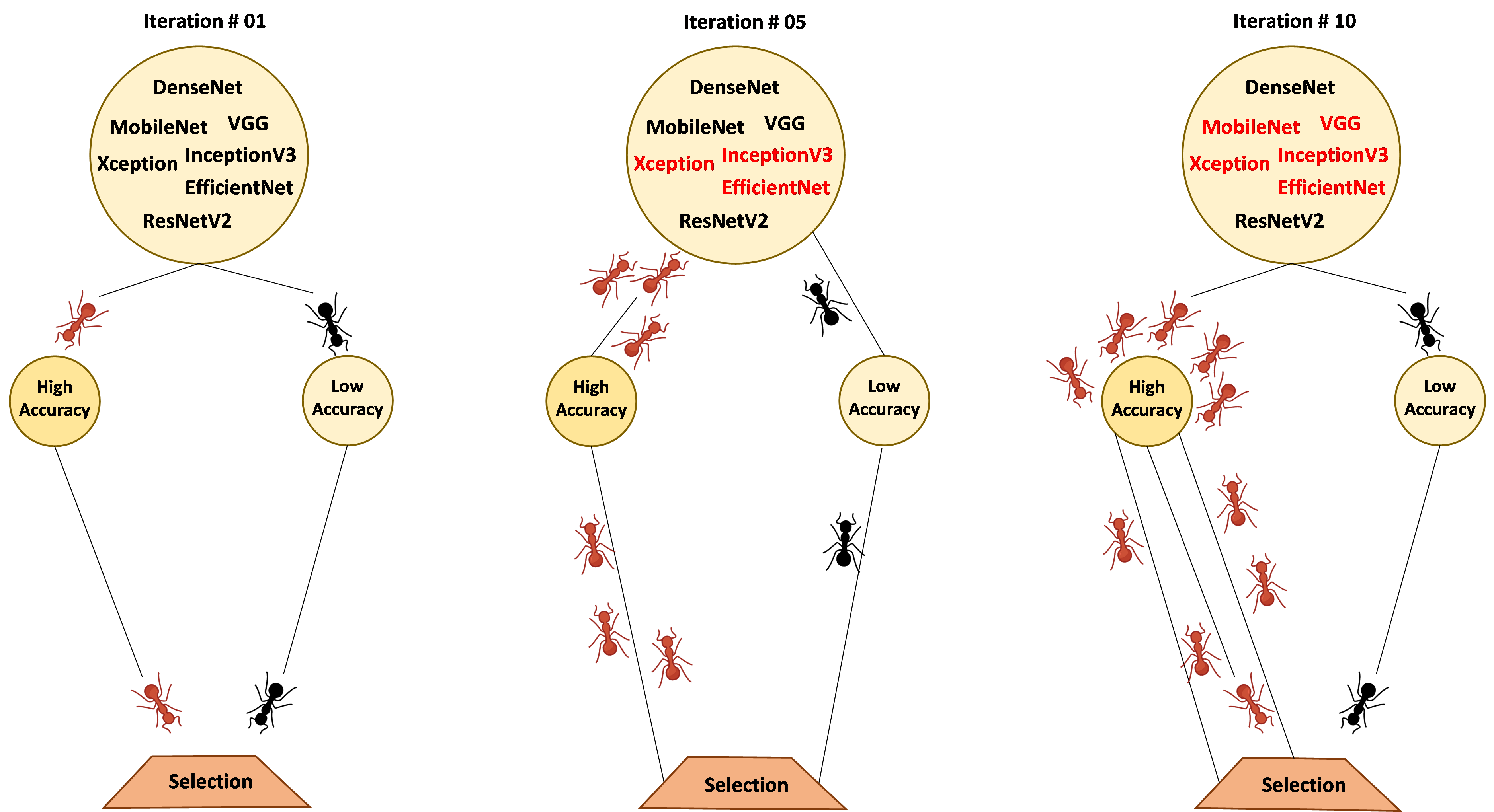}
    \caption{Ant Colony-Based selection of teacher and student models}
    \label{fig:se.png}
\end{figure}
The ACO \cite{wu2024application} algorithm for teacher-student model selection mimics ant foraging behavior to efficiently identify optimal pre-trained teacher-student models for knowledge distillation. Starting with uniform pheromone levels and heuristic values (e.g., validation accuracy) for all models, artificial "ants" probabilistically select models using a weighted combination of pheromones and heuristics, governed by parameters (pheromone influence) and heuristic bias. Each ant evaluates its chosen model on validation data, after which pheromones undergo evaporation (reduced by rate to avoid local optima) and deposition (enhanced for high-performing models). Over iterations, this dynamic balances exploration (testing diverse models) and exploitation (reinforcing top candidates), converging toward the best teacher-student model—either the highest-pheromone or highest-validation-performance candidate. The approach scales efficiently across large model pools, adaptively prioritizes promising options, and resists overfitting through pheromone decay, making it ideal for automated, resource-conscious model selection in distillation pipelines. Fig 3 illustrate the ACO approach of selection of teacher and student models. While Agorithm 1 presents the mathematical formula.

\bmhead{Impact of Parameters}: The parameters $\alpha$ (pheromone effect), $\beta$ (heuristic effect), $\rho$ (pheromone evaporation rate), and $q_0$ (exploration-exploitation factor) are crucial in balancing exploration and exploitation in the ACO algorithm, directly influencing its performance. The value of $\alpha$ determines how strongly ants are influenced by pheromone trails, helping to exploit previously found solutions, while $\beta$ governs the importance of heuristic information, guiding the search towards promising regions. The evaporation rate $\rho$ ensures that older solutions gradually lose their influence, maintaining search diversity and preventing stagnation. Lastly, $q_0$ controls the exploration-exploitation trade-off, with values close to 1 favoring exploitation and values near 0 promoting exploration. Proper tuning of these parameters is essential for achieving a balance between exploring new solutions and exploiting known efficient solutions, thereby optimizing the teacher-student model selection process. These parameter settings have been chosen based on empirical results and domain-specific insights to ensure effective convergence towards optimal solutions.
\begin{algorithm}
\caption{ACO for Selection of Teacher and Student Model}
\begin{algorithmic}[1]
\State \textbf{Input:}
\State \quad A set of pre-trained (sixteen) $M = \{\text{DenseNet, ResNetV2, VGG, MobileNet, \dots}\}$
\State \quad Performance evaluation $P_m$ (validation accuracy)
\State \quad Total ants $N_{\text{ants}}$ (Solutions generated per iteration)
\State \quad Number of iterations $N_{\text{iterations}}$
\State \quad Pheromone matrix $\Phi$, initially set to ones for each model
\State \quad Heuristic matrix $H$, based on model performance measure $P_m$
\State \quad Parameters:
\State \quad \quad $\alpha$ (pheromone effect)
\State \quad \quad $\beta$ (heuristic effect)
\State \quad \quad $\rho$ (pheromone evaporation rate)
\State \quad \quad $q_0$ (exploration-exploitation factor)

\State \textbf{Output:} Best teacher-student model based on highest performance (validation accuracy)

\State \textbf{Initialize Variables:}
\State Initialize pheromone matrix $\Phi = [1, 1, \dots]$ for each model in $M$
\State Initialize heuristic matrix $H$ using model outcome $P_m$ for each model

\For {$i = 1$ to $N_{\text{iterations}}$}
    \For {$k = 1$ to $N_{\text{ants}}$}
        \State Calculate the probability $P_m$ of selecting each model $m$ based on pheromone and heuristic:
        \[
P_{a,b}^{m'} = \frac{\Phi_m^\alpha \times H_m^\beta}{\sum_{m'} \Phi_{m'}^\alpha \times H_{m'}^\beta}
\]
        \State Select a high performing model $m$ based on the calculated probabilities using the roulette wheel selection method (randomly choose with probabilities proportional to: \[P_{a,b}^{m'}\]
        \State Train the selected model on the training dataset.
        \State Evaluate the model on the validation dataset to compute its performance (e.g., validation accuracy).
        \State Store the performance score for the selected model.
    \EndFor
    \State \textbf{Update Best Performance:}
    \State Track the best performance across all ants. If the performance of an ant's selected model exceeds the current best, update the best performance and store the corresponding model as the best solution.
    \State \textbf{Update Pheromone Matrix:}
    \State After all ants have selected and evaluated models, update the pheromone matrix as:
    \[
    \Phi_m = (1 - \rho) \cdot \Phi_m + \sum_{k=1}^{N_{\text{ants}}} \delta(\text{model}_m, \text{Solution}_k) \cdot \text{performance}_k
    \]
    where $\delta(\text{model}_m, \text{Solution}_k)$ is 1 if model $m$ was selected by ant $k$, and 0 otherwise. The performance of each selected model is added to the pheromone value.
\EndFor

\State \textbf{Return Best Teacher-Student Model:}
\State After completing all iterations, select the model with the highest performance as the best Teacher-Student model.
\State Output the best Teacher-Student model and its corresponding performance.
\end{algorithmic}
\end{algorithm}
\subsubsection{Running Example}
 Probability Calculation: 
\[
P_{a,b}^{m'} = \frac{\Phi_m^\alpha \times H_m^\beta}{\sum_{m'} \Phi_{m'}^\alpha \times H_{m'}^\beta}
\]
\textbf{Number of models = 3}\\
Assume iterations \(a\) and \(b\) fixed (just calculation of probabilities for these models).
\\
\textbf{Parameters:}

\[
\begin{array}{cccc}
\hline
\text{Model {m'}} & \Phi_m \ (\text{pheromone}) & H_m \ (\text{heuristic}) & \Phi_m^\alpha \times H_m^\beta \ (\text{with } x = 1, y = 2) \\
\hline
1 & 2 & 3 & 2^1 \times 3^2 = 2 \times 9 = 18 \\
2 & 1 & 5 & 1^1 \times 5^2 = 1 \times 25 = 25 \\
3 & 4 & 2 & 4^1 \times 2^2 = 4 \times 4 = 16 \\
\hline
\end{array}
\]
Sum over all models:
\[
\sum_{{m'} = 1}^3 \Phi_{m'}^\alpha \times H_{m'}^\beta = 18 + 25 + 16 = 59
\]
Calculate probability for each model:
\[
P_{(a,b)}^1 = \frac{18}{59} \approx 0.305
\]
\[
P_{(a,b)}^2 = \frac{25}{59} \approx 0.424
\]
\[
P_{(a,b)}^3 = \frac{16}{59} \approx 0.271
\]
So, the probability that the ant selects model 2 is highest (42.4\%).\\
\textbf{Pheromone Matrix Update}
 \[
    \Phi_m = (1 - \rho) \cdot \Phi_m + \sum_{k=1}^{N_{\text{ants}}} \delta(\text{model}_m, \text{Solution}_k) \cdot \text{performance}_k
    \]
Where:
\begin{itemize}
    \item \(\rho\) = evaporation rate (e.g., 0.1)
    \item \(\delta(\text{model}_m, \text{solution}_k) = 1\) if ant \(k\) selected model \(m\), else 0
    \item \(\text{performance}_k\) = performance of model selected by ant \(k\)
\end{itemize}

\textbf{Example:}
\begin{itemize}
    \item Number of ants: \( N_{\text{ants}} = 3 \)
    \item Evaporation rate: \( \rho = 0.1 \)
    \item Current pheromone values for models:
\end{itemize}
\[
\begin{array}{cc}
\hline
\text{Model m} & \text{Current } \Phi_m \\
\hline
1 & 2 \\
2 & 1 \\
3 & 4 \\
\hline
\end{array}
\]
\textbf{Ants selected models and their performance:}
\[
\begin{array}{ccc}
\hline
\text{Ant k} & \text{Selected Model} & \text{Performance} \\
\hline
1 & \text{Model 2} & 0.8 \\
2 & \text{Model 1} & 0.9 \\
3 & \text{Model 2} & 0.7 \\
\hline
\end{array}
\]
\textbf{Calculation updates for each model:}
\[
\Phi_1 = (1 - 0.1) \times 2 + (\delta(1,1) \times 0.9 + \delta(1,2) \times 0.0 + \delta(1,3) \times 0.0) = 1.8 + 0.9 = 2.7
\]
\[
\Phi_2 = (1 - 0.1) \times 1 + (\delta(2,1) \times 0.8 + \delta(2,2) \times 0.0 + \delta(2,3) \times 0.7) = 0.9 + 0.8 + 0.7 = 2.4
\]
(Ant 1 and Ant 3 selected model 2)
\[
\Phi_3 = (1 - 0.1) \times 4 + (0 + 0 + 0) = 3.6 + 0 = 3.6
\]\\
No ant selected model 3.
\\

\begin{itemize}
    \item \textbf{Model 1}: The initial pheromone value was \( \Phi_1 = 2 \). After considering the performances of the ants (Ant 1 with 0.9 and Ant 2 with 0.9), the updated pheromone value becomes \( \Phi_1 = 2.7 \).
    
    \item \textbf{Model 2}: The initial pheromone value was \( \Phi_2 = 1 \). Ants 1 and 3 selected Model 2 with performance values of 0.8 and 0.7, respectively. Therefore, the updated pheromone value becomes \( \Phi_2 = 2.4 \).
    
    \item \textbf{Model 3}: The initial pheromone value was \( \Phi_3 = 4 \). Since no ants selected Model 3, its pheromone value remains unchanged, resulting in \( \Phi_3 = 3.6 \).
\end{itemize}
Thus, after the pheromone update, the final pheromone values are:
\[
\begin{array}{cccc}
\hline
\text{Model} & \text{Old } \Phi_m & \text{Update Calculation} & \text{New } \Phi_m \\
\hline
1 & 2 & 1.8 + 0.9 & 2.7 \\
2 & 1 & 0.9 + 0.8 + 0.7 & 2.4 \\
3 & 4 & 3.6 + 0 & 3.6 \\
\hline
\end{array}
\]
\subsubsection{Teacher model}\label{sec4}
DenseNet201 \cite{huang2017densely} is a CNN architecture that has gained significant attention in the medical image classification field due to its unique design and effective performance. It utilizes a DL framework as a pretrained model, where each layer is directly connected to every other layer by allowing for efficient feature propagation and reducing the risk of vanishing gradients. It comprises 201 layers by enabling it to capture intricate patterns and detailed features in medical images that are often crucial for precise diagnosis. One of the key benefits of using DenseNet201 in medical image classification is its capability to leverage transfer learning. By initiating a pre-trained model, researchers can fine-tune the architecture on specific medical datasets by significantly improving classification accuracy even when the training data is limited. Additionally, the increased depth and improved feature extraction lead to better performance in image classification tasks, although it comes with a slightly higher computational cost, where computational cost is a crucial factor in resources constrained environment. Overall, it stands out as a powerful tool for enhancing the precision and efficiency of medical image analysis. In the proposed methodology, we consider DenseNet201 as the teacher model for knowledge transfer. The overall architectural overview of DenseNet201 is detailed in Fig 4.
\begin{figure}[h]
    \centering
    \includegraphics[width = 8cm]{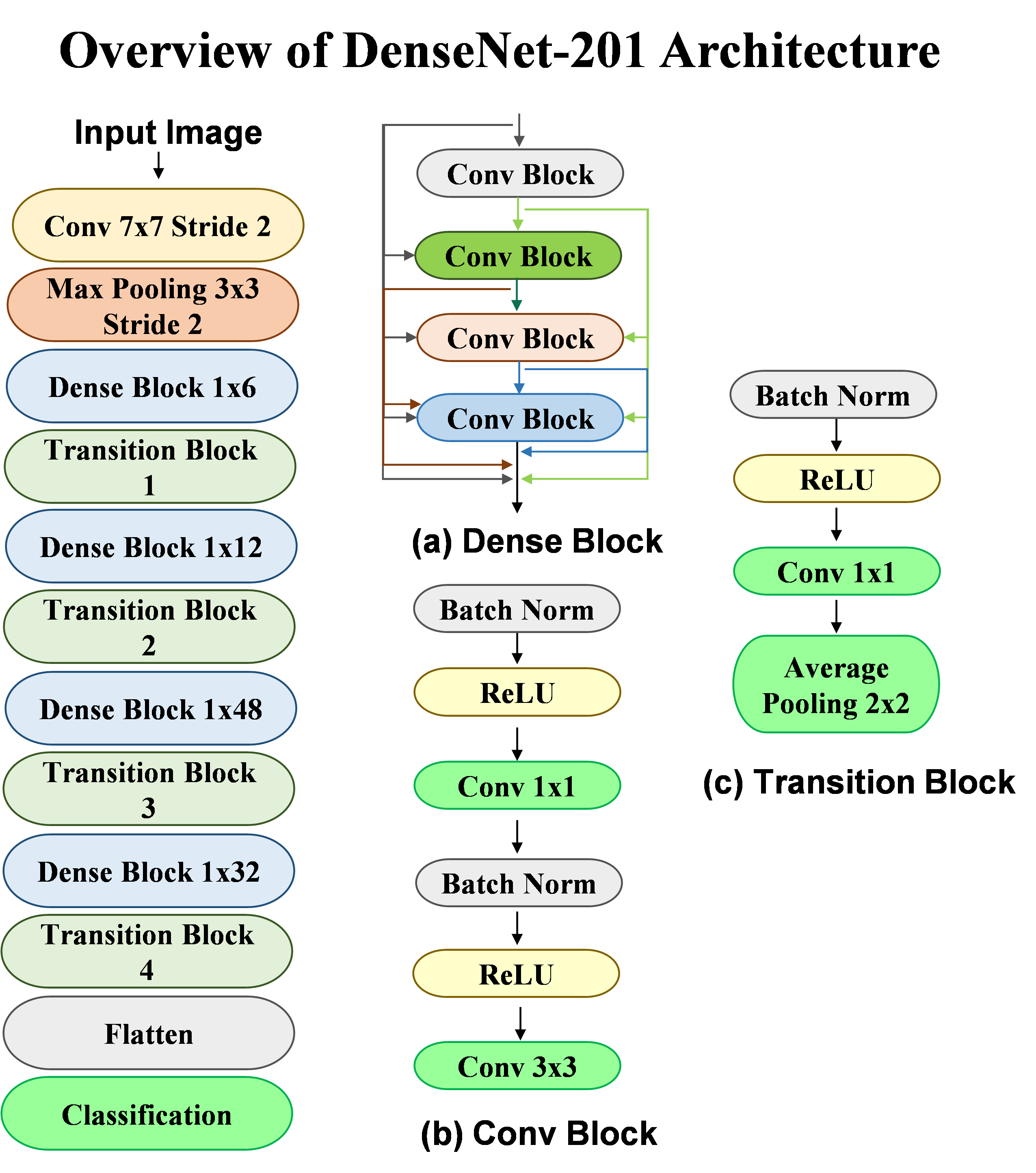}
    \caption{Architecture overview of teacher model}
    \label{fig:se.png}
\end{figure}
\subsubsection{Student model}
ResNet152V2 is an advanced deep residual network introduced by He et al. \cite{he2016identity} as an improved version of the earlier ResNet152 model. It is well-known architecture due to its extensive depth and remarkable performance capabilities. We leveraged ResNet152V2 model due to its superior accuracy within the ResNet family, as evidenced by its performance metrics. Despite having 152 layers, this model is less computationally complex compared to several other architectures, such as VGG16 family models and ConvNext series models. The increased depth of the network facilitates enhanced feature extraction by facilitating improved recognition of complex patterns. However, training deeper networks presents challenges related to backpropagation, particularly the issue of vanishing gradients. ResNet addresses these concerns by incorporating residual connections, which help to mitigate the effects of vanishing gradients and enable smoother training. A key distinction between ResNet-V1 and ResNet-V2 is the implementation of batch normalization and ReLU activation prior to each weight layer. This modification contributes to improved training efficiency and overall model performance by considering it a compelling choice for a variety of DL applications. In the proposed methodology, ResNet152V2 is considered the student model for learning from the teacher model through knowledge distillation. Fig 5 presents the architecture overview of ResNetV2-152 model.
\begin{figure}[h]
    \centering
    \includegraphics[width = 8cm]{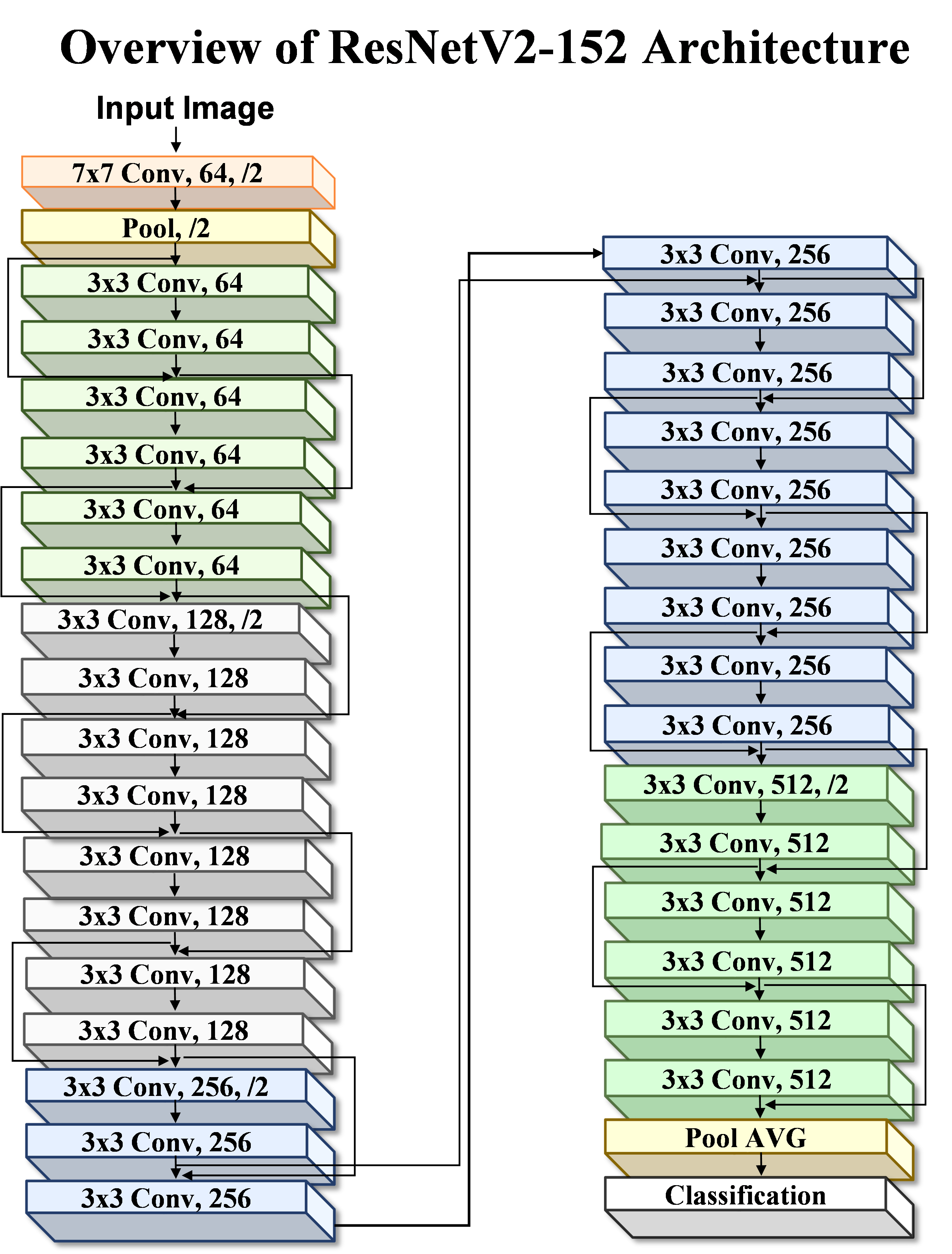}
    \caption{Architecture overview of student model}
    \label{fig:se.png}
\end{figure}
\subsection{Context-unaware Temperature Mechanism for Uncertainty-Aware Distillation}
In traditional KD \cite{habib2024comprehensive}, a less variability temperature parameter \( T \) is used to soften the predictions made by the teacher model. This temperature value is constant throughout the training process, as illustrated in Fig 6. While this approach works well in general (Eq-1), it has a limitation in scenarios where there is varying uncertainty in the data. Specifically, in medical image analysis, the uncertainty across different image regions can differ significantly. For example, certain areas of a medical image may contain highly clear and easily identifiable features, while others may be ambiguous or noisy, requiring more nuanced interpretation. The use of a constanted temperature does not account for these varying levels of uncertainty. As a result, it may not effectively adapt to the complexity and intricacy present in medical images, potentially leading to suboptimal performance in areas with high uncertainty.
\begin{equation}
X_{\text{KD}} = (1 - t) \cdot X_{\text{Entropy}} + t \cdot Y^2 \cdot \text{KL}(\sigma\left(\frac{V_Y}{Y}\right), \sigma\left(\frac{V_s}{Y}\right))
\end{equation}

Where:
\begin{itemize}
    \item \( X_{\text{Entropy}} \) denotes the cross-entropy loss, which is employed to train the student model using the actual labels.
    \item \( \text{KL} \) is the Kullback-Leibler divergence, which measures how different the softened probability distributions produced by the teacher and the student are.
    \item \( \sigma\left(\frac{V_Y}{Y}\right), \sigma\left(\frac{V_s}{Y}\right) \) represent the softmax functions applied to the teacher and student logits after they have been scaled by the temperature \( Y \), respectively. These functions yield the probability distributions for the teacher and student models.
    \item \( t \) represents a weighting factor that adjusts the relative contributions of the cross-entropy loss and the distillation loss.
    \item \( V_Y \) and \( V_s \) are the logits from the teacher and student models, respectively.
    \item By scaling the logits, the temperature parameter \( Y \) regulates how smooth the probability distributions are.
\end{itemize}
\begin{figure}[!h]
\centering
\begin{minipage}[]{0.46\textwidth}
  \centering
  \includegraphics[width = \textwidth]{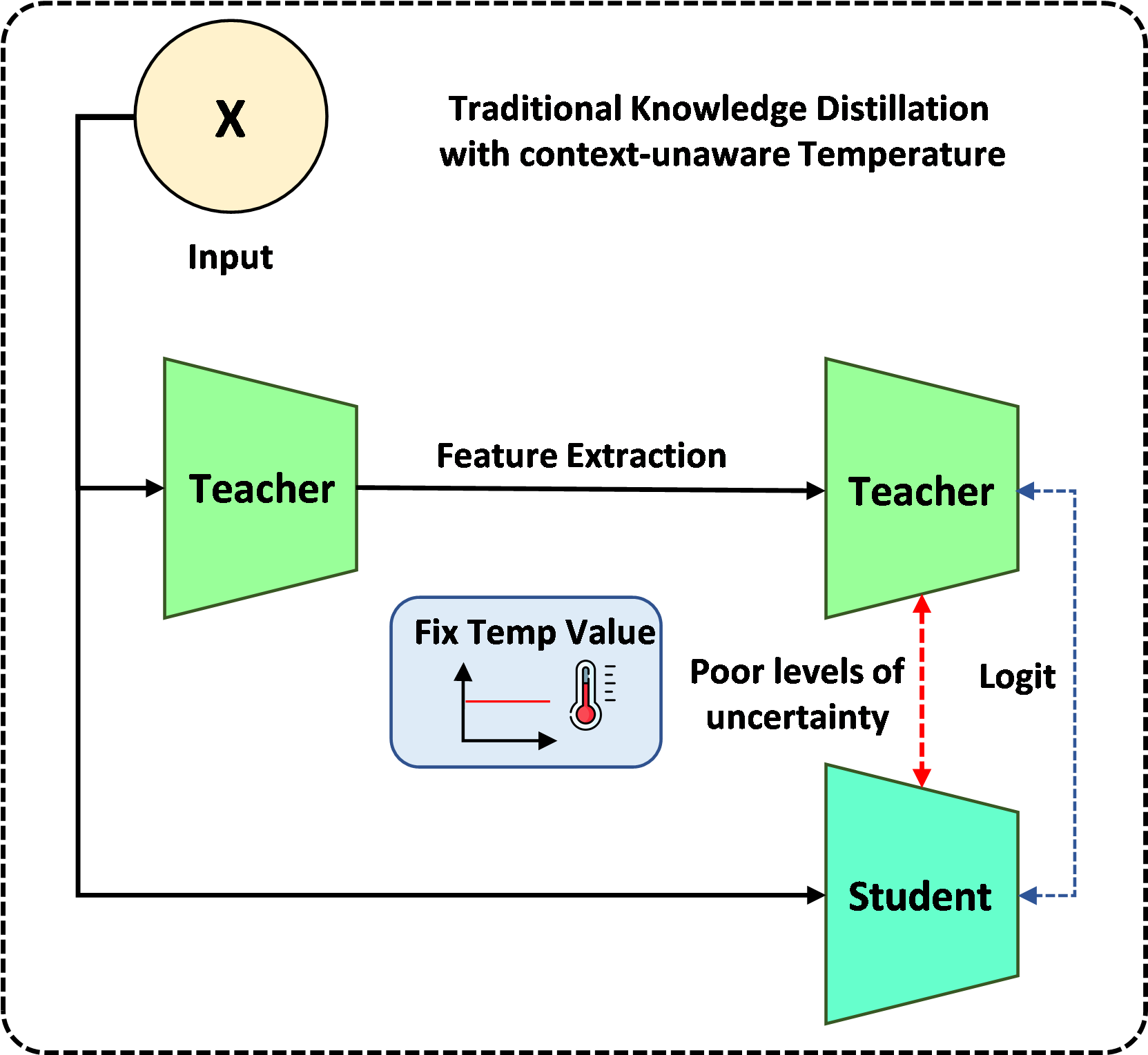}
  
  % Adds some space between the image and text
    % \begin{center}
    % \textbf{Pass density}    
    % \end{center}
\end{minipage}
\begin{minipage}[]{0.46\textwidth}
  \centering
  \includegraphics[width = \textwidth]{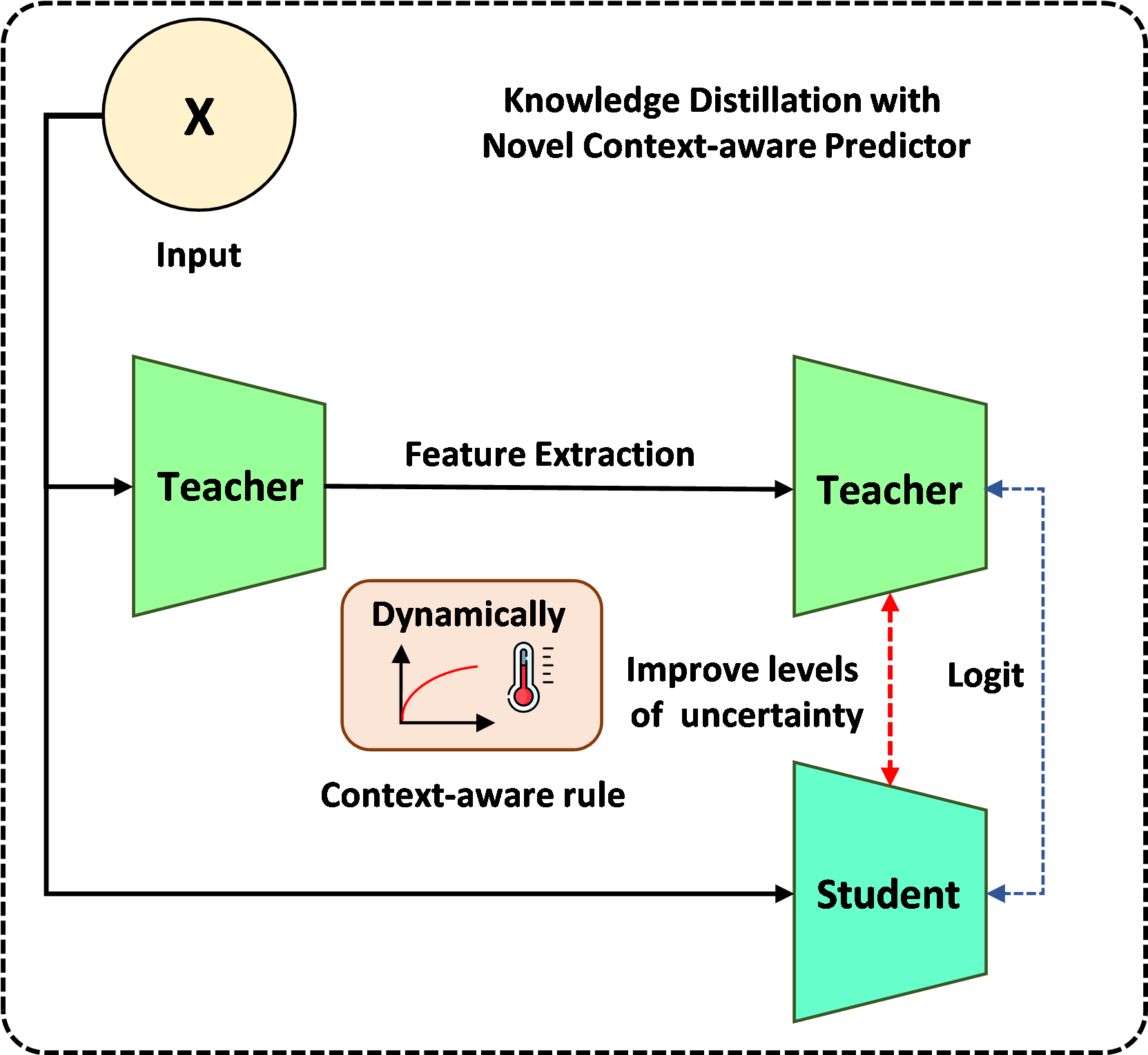}
 
 % Adds some space between the image and text
    % \begin{center}
    % \textbf{Duel density}    
    % \end{center}
\end{minipage}
   
  \caption{Knowledge distillation: (a) Constant temperature and (b) Context-aware adjusted temperature}
  \label{fig:11.png}
\end{figure}

\subsubsection{Novel Context-aware Adaptive Temperature Scaling}
A constant temperature fails to account for varying uncertainty levels in medical images, limiting its ability to adapt to complex or ambiguous regions. A dynamic temperature approach, adjusting based on data uncertainty, could improve the distillation process by focusing on more challenging regions. To address the uncertainty in medical images, we propose novel context-aware predictor dynamically adjust the temperature values during KD. Traditional methods use a context-unaware temperature to transfer knowledge from the teacher model (DenseNet201) to the student model (ResNet152V2), but they fail to consider uncertainty or ambiguity in medical images. We utilized Context-aware Adaptive Temperature Scaling which help to:
\begin{enumerate}
    \item \textbf{Handling Uncertainty:} Medical images often contain noise or ambiguous features. context-aware predictor soft decision boundaries, enhancing the student model's robustness to uncertainty.
    
    \item \textbf{Context-aware Adaptive Distillation:} Rather than relying solely on the teacher's soft labels, context-aware predictor adjusts the knowledge transfer process based on the teacher's confidence levels.
\end{enumerate}
\begin{itemize}
    \item \textbf{Context-aware Rule}
    \begin{itemize}
        \item \textbf{Rule-1:} IF image is noisy AND teacher’s confidence is low, THEN increase \( T \) (softer labels).
        \item \textbf{Rule-2:} IF image is clear AND teacher’s confidence is high, THEN decrease \( T \) (harder labels).
        \item \textbf{Rule-3:} IF disease is complex, THEN give more weight to teacher prediction.
    \end{itemize}
\end{itemize}
For example, when an image is noisy and the teacher confidence is low, the temperature is increased, allowing for softer labels that help the student model learn from more uncertain data. Conversely, if the image is clear and the teacher confidence is high, the temperature is decreased to produce harder labels, ensuring that the student focuses on more confident predictions. Additionally, if the disease in question is complex, the rules assign more weight to the teacher predictions to ensure that critical information is emphasized. These adaptive rules help the model better handle uncertainty, improving both robustness and accuracy, especially in medical image analysis where data quality and complexity vary significantly. Fig 6(b) display the Context-aware adaptive temperature scaling.
\subsection{Overview of Context-aware Adaptive Temperature Scaling within the Knowledge Distillation Framework}
Brain tumor detection is a critical task in medical imaging, where early and accurate identification can significantly improve patient outcomes. However, the complexity and variability of medical images especially when dealing with multi-disease scenarios pose significant challenges for conventional CNN models. In brain tumor detection, the diversity in tumor types, imaging conditions, and patient-specific factors requires robust and adaptable models. Standard Knowledge Distillation (KD) techniques, which rely on a context-unaware temperature to transfer knowledge from the teacher model to the student model, fail to address the inherent uncertainty and variability in medical images, leading to suboptimal performance. To overcome these challenges, we propose a context-aware predictor rule-based adaptive temperature scaling mechanism that enhances the KD framework, making it more suited to medical image analysis, particularly for complex multi-disease brain tumor detection. In our approach, we incorporate context-aware predictor into the KD process to dynamically adjust the temperature parameter based on various factors, including image quality, disease complexity, teacher prediction confidence, and feature importance. Traditional KD methods use a context-unaware temperature value throughout the training process, which does not adapt to the changing levels of uncertainty in the data. For instance, medical images often contain noise, blurriness, and ambiguous features that can confuse the model. Context-aware predictor addresses these issues by allowing the temperature to vary depending on the specific characteristics of the image. If the image quality is poor, or if the teacher model is uncertain about its predictions, the context-aware predictor increases the temperature to allow for softer labels. This helps the student model learn from ambiguous data more effectively. Conversely, when the image is clear and the teacher model predictions are confident, the temperature is lowered, resulting in harder labels and encouraging the student model to focus on more reliable information.
  Moreover, we enhance the KD process by employing ACO for teacher and student model selection. ACO intelligently explores the space of possible teacher-student model combinations by guiding the search process using pheromone-based exploration. Unlike simpler approaches that run models sequentially and select the top performers, ACO dynamically adapts over iterations, taking into account both the performance of each model and its previous successes. This adaptive approach ensures that the best teacher-student model pair is chosen for distillation, maximizing the efficiency of the training process. By selecting the most appropriate models for each stage of distillation, we can better conserve computational resources while improving model accuracy. In our experiments, where we utilized 16 pre-trained models, ACO played a crucial role in selecting the optimal pairings, ensuring that the most relevant and effective models were used for knowledge transfer. Fig 7 presents the overview of proposed context-aware adaptive temperature scaling framework in KD. Our use of adaptive feature guidance through student-teacher distillation shares similarities with recent fusion frameworks for robust generalization across domains \cite{zhang2025reproducible}.
\begin{figure}[h]
    \centering
    \includegraphics[width = 15cm]{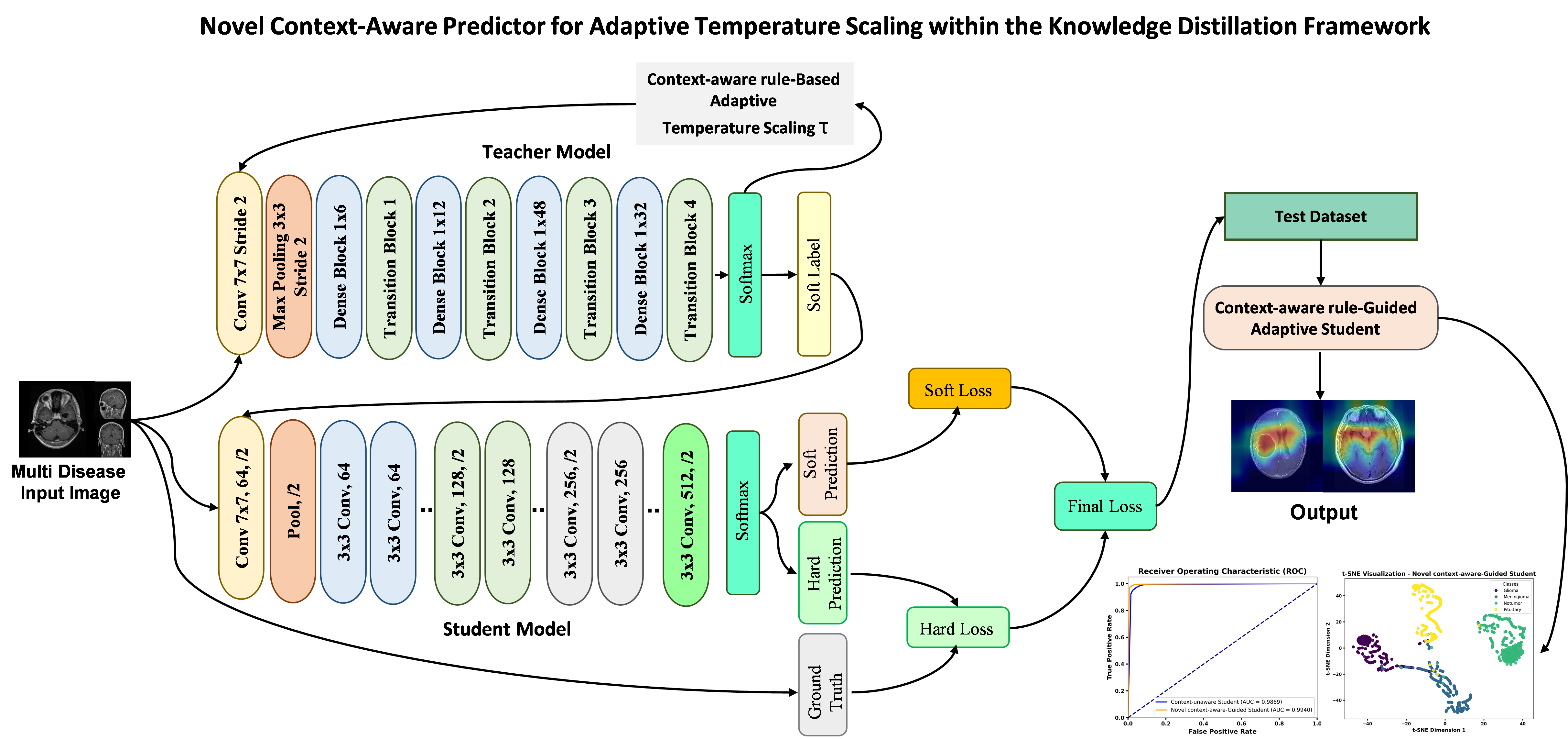}
    \caption{Overview of Proposed Context-aware Adaptive temperature scaling framework in Knowledge distillation}
    \label{fig:se.png}
\end{figure}

In this work, we address the challenge of adjusting temperature scaling dynamically based on various factors inherent to the input data, such as image noise, disease complexity, and teacher model confidence. These factors contribute to the uncertainty in the model's predictions, and our approach leverages uncertainty-based scaling to improve the performance and robustness of KD frameworks, especially in the context of noisy or ambiguous medical imaging data.
\subsection{Running Example}
\subsubsection{Problem: Uniform Temperature Scaling in KD}
In traditional KD, the teacher model provides soft targets using a constant temperature scaling factor \( \tau \). The probability distribution over classes is computed as:
\begin{equation}
P_T^{(\tau)}(x) = \text{softmax}\left(\frac{z_T(x)}{\tau}\right)
\end{equation}
where:
\begin{itemize}
  \item \( z_T(x) \) denotes the teacher logits for input \( x \),
  \item \( \tau \) is a context-unaware temperature scalar.
\end{itemize}

However, medical images often exhibit varying levels of uncertainty due to factors like noise, occlusion, and anatomical ambiguity. Using a context-unaware \( \tau \) limits the teacher adaptability to such data variability, leading to suboptimal student performance.

\subsubsection{Proposed Solution: Context-Aware Rule-Based Temperature Scaling}
We propose a dynamic temperature scaling strategy, where \( \tau \) is a function of the image uncertainty:
\begin{equation}
\tau(x) = 1 + \alpha \cdot U(x)
\end{equation}
\begin{itemize}
  \item \( U(x) \) is an uncertainty score (e.g., entropy of predictions, ensemble variance),
  \item \( \alpha \) is a scaling hyperparameter.
\end{itemize}
This results in a context-aware softmax:
\begin{equation}
P_T^{(\tau(x))}(x) = \text{softmax}\left(\frac{z_T(x)}{\tau(x)}\right)
\end{equation}

\subsubsection*{Numerical Example}
Consider teacher logits for a sample:
\[
z_T(x) = [2.0, 0.5, -1.0]
\]
Assume:
\begin{itemize}
  \item context-unaware temperature: \( \tau = 2 \),
  \item Context-aware: \( U(x) = 0.3 \), \( \alpha = 2 \Rightarrow \tau(x) = 1 + 2 \cdot 0.3 = 1.6 \).
\end{itemize}

\noindent \textbf{Constant Temperature Scaling:}
\begin{align*}
\frac{z_T(x)}{\tau} &= \left[\frac{2.0}{2}, \frac{0.5}{2}, \frac{-1.0}{2}\right] = [1.0, 0.25, -0.5] \\
P_T^{(2)}(x) &= \text{softmax}([1.0, 0.25, -0.5]) \approx [0.61, 0.27, 0.12]
\end{align*}

\noindent \textbf{Context-Aware Scaling:}
\begin{align*}
\frac{z_T(x)}{\tau(x)} &= \left[\frac{2.0}{1.6}, \frac{0.5}{1.6}, \frac{-1.0}{1.6}\right] \approx [1.25, 0.31, -0.625] \\
P_T^{(1.6)}(x) &= \text{softmax}([1.25, 0.31, -0.625]) \approx [0.65, 0.23, 0.12]
\end{align*}
\subsubsection{Impact}
By introducing a context-aware scaling factor based on input uncertainty, the proposed rule-based temperature scaling enhances the flexibility and robustness of the KD framework, especially in the context of noisy or ambiguous medical imaging data. The incorporation of sparse attention and temporal context, as proposed in \cite{zhang2025sparse}, also inspired our use of contextual cues for adaptive temperature assignment in uncertain regions.
\section{Implementation and results}
We implemented Context-aware-based adaptive temperature scaling in KD, adjusting the temperature based on image quality and teacher confidence. ACO has used for selecting the best teacher-student model pairs from sixteen pre-trained models. Experiments on a multi-disease brain tumor detection dataset showed significant improvements in accuracy and robustness. The dynamic temperature adjustment enhanced handling of noisy data, and ACO optimized model selection, leading to better detection performance across various tumor types.
\subsection{Hyperparameter and experimental settings}
We conducted our experiments using Python on a Windows 10 platform with high-performance computing graphics. The model was trained with the following hyperparameters \cite{khan2025detection}: batch size of 64, 30 epochs, a learning rate of 0.001, and the Adam optimizer. To evaluate the model, we used standard performance metrics. For model visualization, we applied GRADCAM to interpret decision-making, and T-SNE was used for dimensionality reduction and visualization of high-dimensional data. Statistical tests were performed to validate the robustness of the results, ensuring a comprehensive evaluation of model performance.
\subsection{Model evaluation metrics}\label{sec5}
The efficacy of the suggested methodology in accurately classifying tumor and nontumor MRI images is being evaluated. A range of performance metrics, including accuracy (Eq-5), precision (Eq-6), recall (Eq-7), and f1-score (Eq-8) have been employed. Furthermore, the confusion matrix is also shown to illustrate the predicted and actual results of the brain tumor diagnostic from the proposed framework. The following formulas were applied to test data in order to calculate the overall model performance from training to testing. In the following equations, TP stands for true positive, FP for false positive, TN for true negative, and FN for false negative. 
\begin{equation}
\text{Accuracy} = \frac{TP + TN}{TP + FP + TN + FN} \tag{5}
\end{equation}

\begin{equation}
\text{Precision} = \frac{TP}{TP + FP} \tag{6}
\end{equation}

\begin{equation}
\text{Recall} = \frac{TP}{TP + FN} \tag{7}
\end{equation}

\begin{equation}
\text{F1 Score} = 2 \times \frac{\text{Precision} \times \text{Recall}}{\text{Precision} + \text{Recall}} \tag{8}
\end{equation}
\subsection{TRIPOD Checklist for Compliance Reporting}\label{sec5}
The TRIPOD checklist presenting in this section, which highlights the reporting elements and related sections in this study. It guarantees openness and conformity to accepted guidelines for creating and assessing prediction models in medical research.
\begin{itemize}
  \item 1. Title and Abstract
   \begin{itemize}
    \item 1.1 Title: Describes the prediction model and its application.
   \item 1.2 Abstract: Clearly states the aim, methods, results, and conclusions in a structured format.
   \end{itemize}
  \item 2 Introduction
    \begin{itemize}
    \item 2.1 Background: The context for using medical imaging (brain tumor and gastrointestinal data) for diagnosis is established.
   \item 2.2 Problem Formulation: The issue with conventional knowledge distillation methods and how proposed approach addresses the limitations.
   \end{itemize}
   \item 3 Methods
     \begin{itemize}
    \item 3.1 Model Development
         \begin{itemize}
    \item 3.1.1 Purpose: Describes the purpose of using the Context-aware Student model (CASM) with knowledge distillation.
   \item 3.1.2 Methods of Model Building: Integration of Ant Colony Optimization (ACO) for teacher-student model selection, along with the context-aware temperature adjustment mechanism.
   \item 3.1.3 Feature Selection: Details are provided on the datasets used (e.g., Kaggle MRI dataset, Figshare MRI dataset, GastroNet dataset), which are relevant to the features of interest for tumor classification.
   \end{itemize}
   \item 3.2 Data Collection and Preprocessing
           \begin{itemize}
    \item 3.2.1 Data Collection: The MRI datasets (Kaggle, Figshare, GastroNet) are described, along with their specific contents (e.g., types of brain tumors).
   \item 3.2.2 Data Preprocessing: Techniques like resizing, rescaling, rotation, and flipping are employed.
   \end{itemize}
   \item 3.3 Model Evaluation: Evaluation metrics include accuracy, precision, recall, F1-score, and confusion matrices.
   \item 3.4 Statistical Analysis: Use of standard performance metrics such as AUC-ROC, PR curves, and statistical tests to validate the results.
   \item 4 Results
    \begin{itemize}
    \item 4.1 Model Performance: Clear reporting of model performance with accuracy rates across multiple datasets.
   \item 4.2 Validation: Use of multiple datasets for validation (e.g., Kaggle, Figshare, and GastroNet).
   \item 4.3 Comparison with SOTA Methods: The CASM outperforms traditional models, including DenseNet and ResNet architectures.
   \end{itemize}
   \item 5 Discussion
       \begin{itemize}
    \item 5.1 Interpretation of Results: Highlights that the proposed model significantly improves classification accuracy and robustness, especially in handling noisy data.
   \item 5.2 Limitations: The potential limitation in performance for Meningioma and the need for future work to address it.
   \item 5.3 Practical Implications: Discusses the real-world applicability of the proposed model in medical imaging, particularly for clinical deployment.
   \end{itemize}
   \item 6 Conclusions
          \begin{itemize}
    \item 6.1 Summary: The model achieves high accuracy, enhancing the ability to handle complex, multi-disease brain tumor detection and gastrointestinal imaging.
   \item 6.2 Future Work: The approach will be extended to more datasets and clinical settings.
   \end{itemize}
   \item 7 Ethical Considerations
         \begin{itemize}
    \item 7.1 Ethics Approval: The study mentions approval and consent, stating that it was approved where necessary.
   \end{itemize}
   \end{itemize}
\end{itemize}

\subsection{Dataset pre-processing}
Dataset pre-processing is a crucial step in preparing data for effective model training, especially in the context of medical image analysis. This process involves several techniques \cite{khan2025optimized}, including resizing, rescaling, rotation, and flipping, which collectively enhance the robustness and diversity of the dataset. Resizing images ensures uniformity in dimensions by allowing models to process input efficiently without any distortion. Rescaling adjusts the pixel intensity values by normalizing them to a specific range [0-1], which can significantly improve the convergence of neural networks. Rotation introduces variations in the orientation of images by enabling models to become invariant to changes in perspective, while flipping both horizontally and vertically assists to simulate different viewpoints and minimize overfitting. By employing these pre-processing techniques, we not only enrich the dataset but also contribute to a more generalized model that performs better across a variety of real-world scenarios. This comprehensive approach efficiently prepares the data by establishing a strong basis for subsequent training and improving the overall model performance. Fig 8 displays the preprocess and augmented dataset samples of Kaggle multiclass dataset. 
\begin{figure}[h]
    \centering
    \includegraphics[width = 10cm]{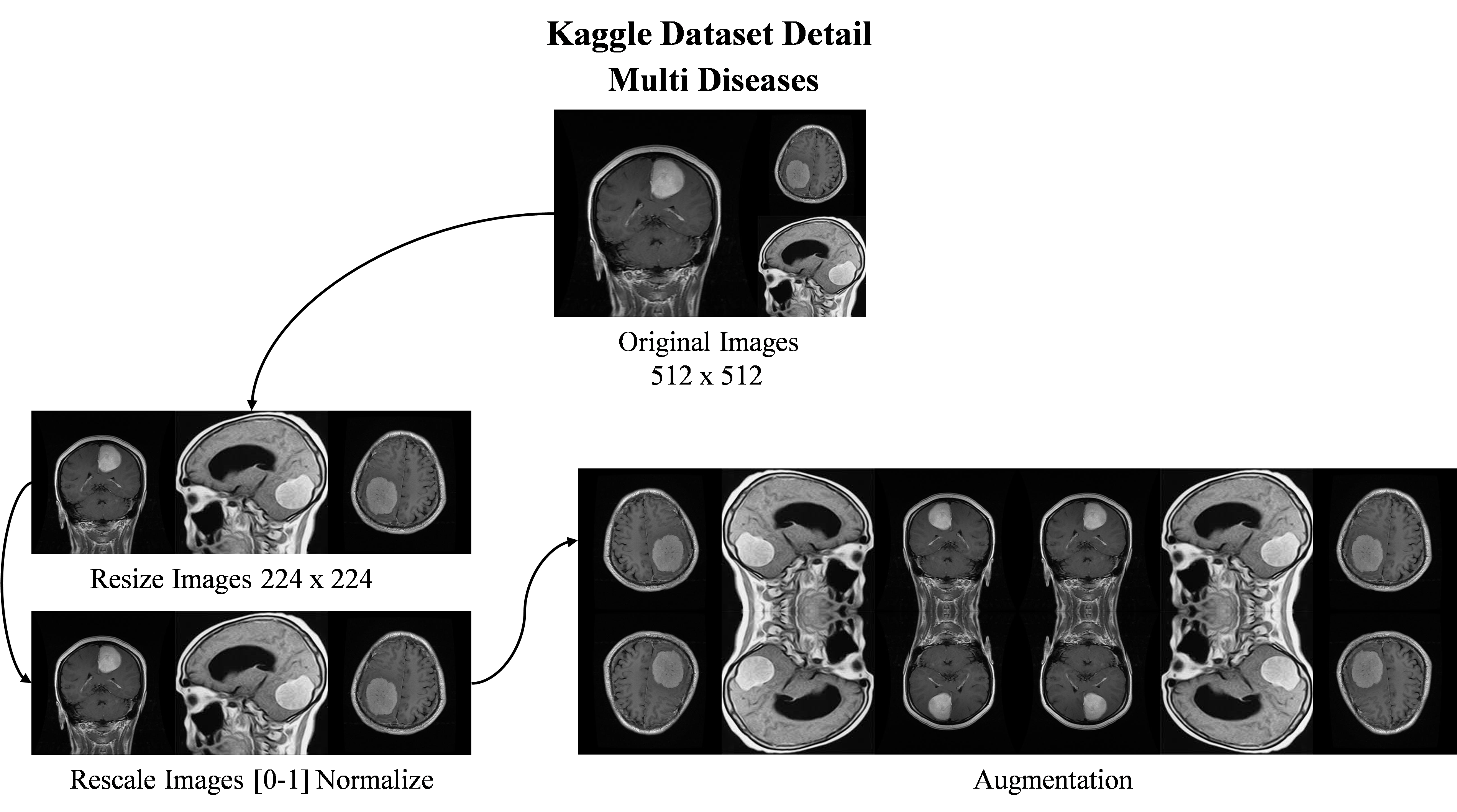}
    \caption{Overview of preprocess and augmented dataset samples}
    \label{fig:se.png}
\end{figure}
\subsection{Classification performance of Context-aware Student model: Kaggle Multi MRI}
In this section, we present the classification performance of the CASM on the Kaggle Multi MRI dataset. Table ~\ref{T1} provides a comprehensive performance comparison of the model using the original context-unaware temperature and the Context-aware temperature. The results are presented in terms of precision, recall, F1-score, and overall accuracy for each class within the dataset. The Original (constant Temp) Student model achieves an overall accuracy of 96.33. Examining the class-specific performance, the model demonstrates exceptional proficiency in identifying No-tumor cases by achieving above 99.00 score in all key performance metrices. For Pituitary tumors, the model also performs well, with a 95.53 precision, 99.67 recall, and 97.55 F1-score. While a still strong but slightly lower performance is observed for Glioma class with 95.05 precision, 96.00 recall, and 95.52 F1-score. The model finds difficulty to detect Meningioma class compared to other three classes by indicating 94.81 precision, 89.54 recall, and 92.10 F1-score. The lower recall for Meningioma suggests a potential area for future improvement by indicating that the model may miss a higher proportion of these cases compared to other tumor types. 
   
   The CASM demonstrates significant improvements across various metrics compared to the model utilizing a context-unaware temperature setting. With an overall accuracy of 98.01, the student model surpasses the performance of the prior model. In terms of class-specific results, it achieves an enhanced precision score of 98.55 for Glioma but it shows slightly lower recall and F1 scores with 94.00 recall and 95.22 F1 score compared to the earlier model. In the Meningioma class, there is a notable improvement in both recall and F1 score metrics, with recall increasing from 89.54 to 96.41, and the F1 score rising from 92.10 to 94.25. However, the precision values experienced a slight decrease by falling from 94.81 to 94.19 when compared to the values achieved by the prior model with a context-unaware temperature setting. The student model also excels in identifying the No-tumor class by attaining a score above 99.50 precision score in all key metrices by marking an improvement over the context-unaware temperature-based model. In the Pituitary class, the model shows slightly lower performance in recall, with a score reduced from 99.67 to 98.67, but still demonstrates enhancements in both precision and F1 scores by reaching 98.34 and 98.50, respectively. Overall, the implementation of Context-aware temperature adjustments results in substantial gains in accurately classifying various tumor types and normal cases within the Kaggle MRI dataset, underscoring the effectiveness of the proposed approach in enhancing diagnostic accuracy in medical image analysis.
\begin{table}[h!]
\centering
\caption{Class-wise performance analysis: Constant temperature and Context-aware Adaptive temperature: Kaggle Multiclass}
\begin{tabular}{llcccc}
\hline
\textbf{Model} & \textbf{Class} & \textbf{Precision} & \textbf{Recall} & \textbf{F1-Score} & \textbf{Accuracy} \\ \hline
\multicolumn{6}{c}{\textbf{Constant Temperature}} \\ 
\textbf{Original (context-unaware) Student} & Glioma & 95.05 & 96.00 & 95.52 & 96.33 \\ 
 & Meningioma & 94.81 & 89.54 & 92.10 &  \\ 
 & No-tumor & 99.01 & 99.26 & 99.14 &  \\ 
 & Pituitary & 95.53 & 99.67 & 97.55 &  \\ 
\multicolumn{6}{c}{\textbf{Context-aware Guided}} \\ 
\textbf{Context-aware Guided Student} & Glioma & 98.55 & 94.00 & 95.22 & 98.01 \\ 
 & Meningioma & 94.19 & 96.41 & 94.25 &  \\ 
 & No-tumor & 99.75 & 99.51 & 99.63 &  \\ 
 & Pituitary & 98.34 & 98.67 & 98.50 &  \\ \hline
\end{tabular}
\label{T1}
\end{table}
The confusion matrix is a valuable tool for evaluating the performance of classification models, as it visually represents the compared of predictions generated by the model to the actual class labels. It allows us to identify not only the number of correctly classified instances but also the types of misclassifications that occur. Fig 9 presents the confusion matrix visualizations for both the context-unaware temperature and CASM. In panel (a), the original (constant temp) student model was tested on the Kaggle Multiclass dataset, which comprised a total of 1,311 images. It successfully classified 1,263 images correctly by resulting in 48 misclassified samples. These results indicate that there is potential for further improvement in performance. In contrast, Fig (b) illustrates the performance of the CASM, which was also tested on the same dataset. The proposed model achieved higher accuracy by correctly classifying 1,285 images while misclassifying only 25 samples. These findings highlight the effectiveness of the Context-aware approach in enhancing classification accuracy and reducing misclassification within the dataset.
\begin{figure}[!h]
\centering
\begin{minipage}[]{0.46\textwidth}
  \centering
  \includegraphics[width = \textwidth]{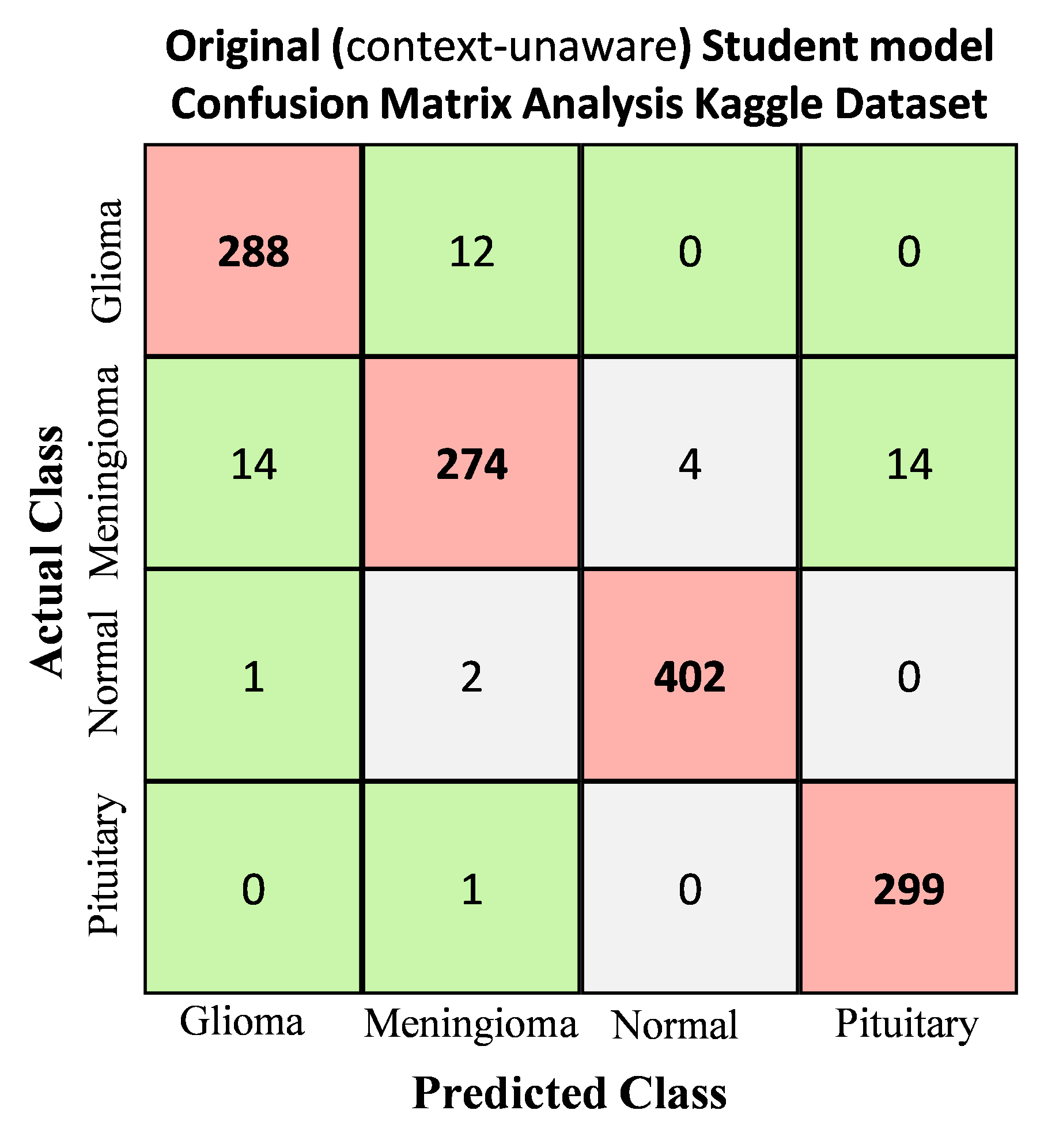}
  
  % Adds some space between the image and text
    % \begin{center}
    % \textbf{Pass density}    
    % \end{center}
\end{minipage}
\begin{minipage}[]{0.46\textwidth}
  \centering
  \includegraphics[width = \textwidth]{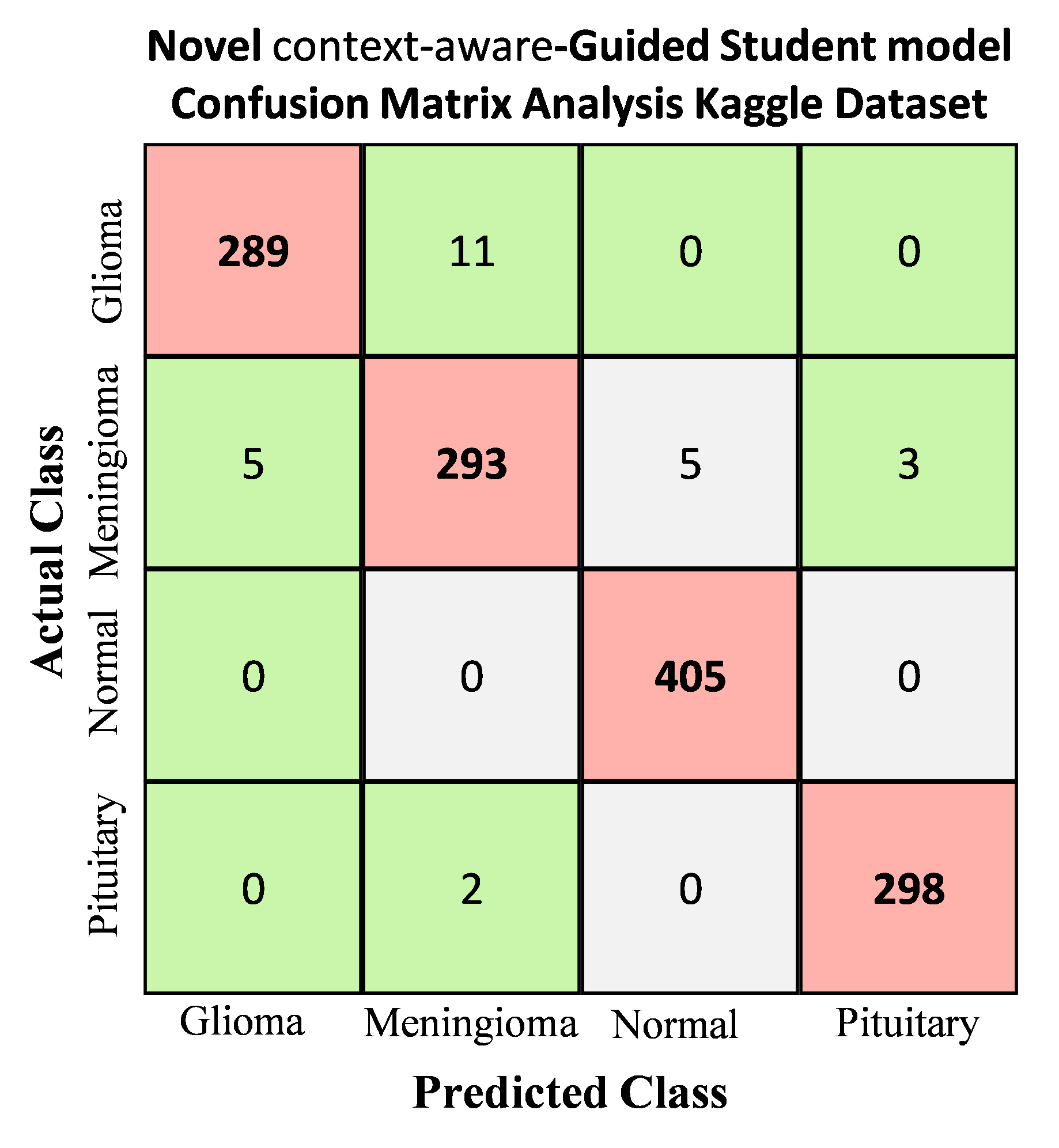}
 
 % Adds some space between the image and text
    % \begin{center}
    % \textbf{Duel density}    
    % \end{center}
\end{minipage}
   
  \caption{Confusion matric visualization: (a) Constant temperature and (b) Context-aware Adaptive temperature: Kaggle Multiclass}
  \label{fig:11.png}
\end{figure}\\
The area under the curve-receiver operating characteristics curve (AUC-ROC) and precision recall (PR) curves are essential tools for evaluating the model performance as they provide insights into the trade-off between true positive rates and false positive rates, as well as the balance between precision and recall across various thresholds. Fig 10 presents a performance analysis of the Original (context-unaware) Student model and the CASM on the Kaggle Multiclass dataset using AUC-ROC and PR curves.  In Fig 10(a), the AUC-ROC curve for the context-aware model demonstrates a higher AUC value of 0.9940 compared to the original model AUC value of 0.9869 by indicating a marginal improvement in overall classification performance. Similarly, Fig 10(b) displays the PR curves, where the Context-aware model achieves a higher AP score of 0.9894 compared to the original model AP value of 0.9751. This improvement in AP suggests the Context-aware model is better at balancing precision and recall, especially at higher thresholds. Both curves indicate that the CASM exhibits superior performance in discriminating between the different classes in the Kaggle Multiclass dataset.
\begin{figure}[!h]
\centering
\begin{minipage}[]{0.46\textwidth}
  \centering
  \includegraphics[width = \textwidth]{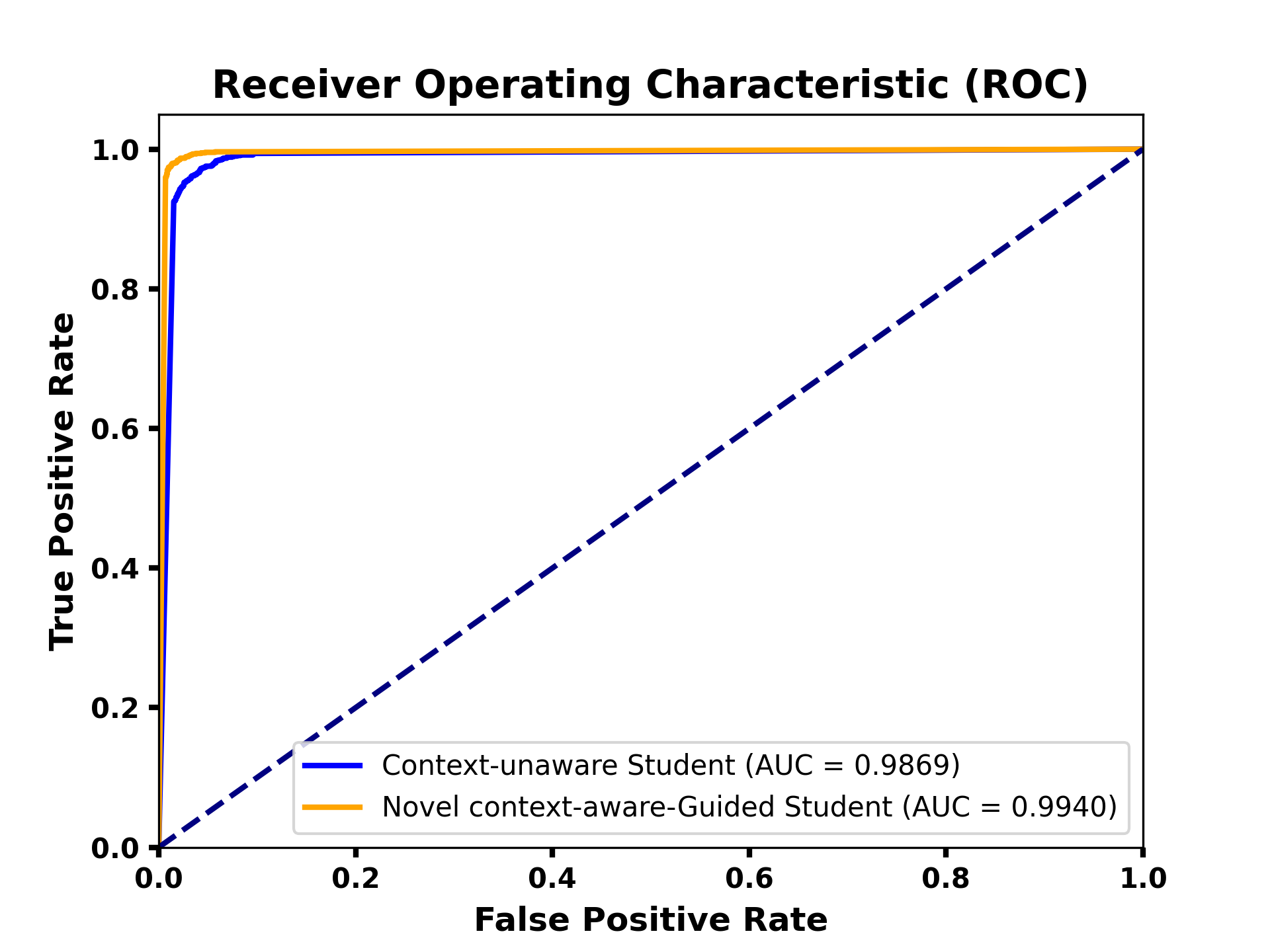}
  
  % Adds some space between the image and text
    % \begin{center}
    % \textbf{Pass density}    
    % \end{center}
\end{minipage}
\begin{minipage}[]{0.46\textwidth}
  \centering
  \includegraphics[width = \textwidth]{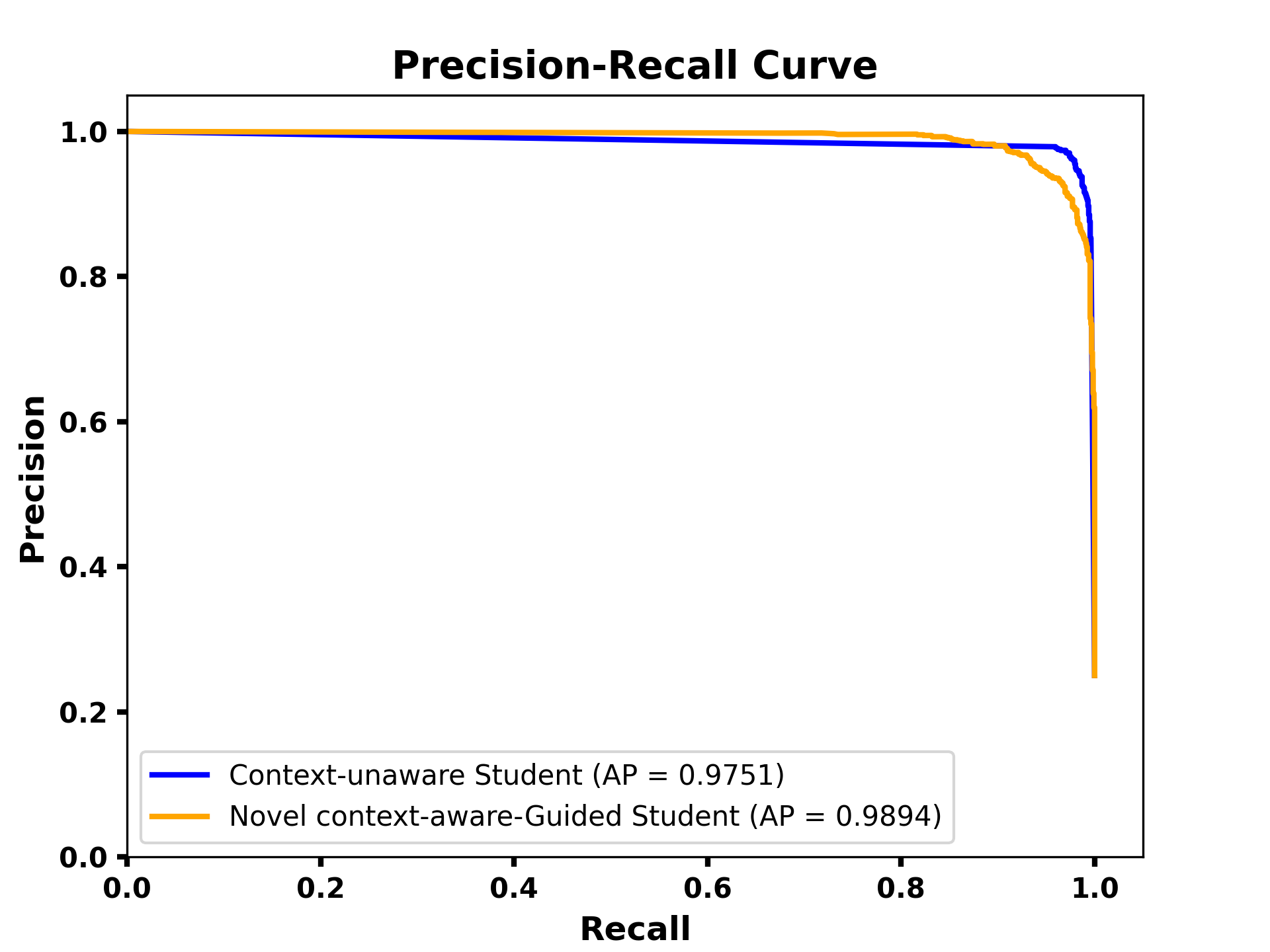}
 
 % Adds some space between the image and text
    % \begin{center}
    % \textbf{Duel density}    
    % \end{center}
\end{minipage}
   
  \caption{Performance analysis with: (a) AUC-ROC curve and (b) PR Curve: Kaggle Multiclass}
  \label{fig:11.png}
\end{figure}
\subsection{Performance evaluation of Context-aware Student with existing model}
In this section, we will analyze the performance of various pre-trained models including DenseNet201-Base  \cite{huang2017densely} and ResNet152V2-Base \cite{he2016identity}, across all key performance indicators. Both of these models exhibit superior performance compared to other pre-trained models as outlined in Table \ref{T2}. DenseNet201 stands out as the top-performing base model, achieving the highest accuracy at 95.33. Following closely is ResNet152V2-Base, which demonstrates the second-highest accuracy at 95.19. The lowest accuracy is recorded by the VGG19-Base model, at 89.47, while the remaining pre-trained models fall within the accuracy range of 90.61 to 94.58. In terms of recall, DenseNet201 again leads with the highest score of 96.11, while ResNet152V2 follows with a recall score of 94.86. Other models also performed well but showed lower recall scores compared to these top base models, with values ranging from 88.91 to 94.37. This high-performance positions DenseNet201 as the teacher model and ResNet152V2 as the student model in our proposed approach. Ultimately, our proposed CASM surpasses all pre-trained and base models, achieving an exceptional accuracy score of 98.01 along with the highest recall score of 97.89. Furthermore, it exhibits improved performance in precision and F1-score compared to other models. This indicates that the integration of Context-aware-Guided rules significantly enhances performance beyond the capabilities of the individual pre-trained models alone.

Moreover, CASM significantly outperforms all VIT baselines across all metrics. Compared to Vit-B16-Base (93.52, 93.09, 93.16, 93.53), Vit-B32-Base (95.73, 95.65, 95.64, 95.51), Vit-L16-Base (95.12, 94.93, 94.97, 94.63), and Vit-L32-Base (95.90, 95.49, 95.55, 95.82), the CASM achieves a superior score of 98.01 in all categories, demonstrating a clear performance advantage over the VIT baselines.
\begin{table}[ht]
\centering
\caption{Comparison of Context-aware with existing pre-trained: Kaggle Multiclass}
\begin{tabular}{lcccc}
\hline
\textbf{Method} & \textbf{Accuracy} & \textbf{F1-Score} & \textbf{Recall} & \textbf{Precision} \\
\hline
DenseNet121-Base & 92.82 & 92.45 & 92.37 & 92.87 \\
DenseNet169-Base & 94.58 & 94.22 & 94.37 & 94.34 \\
DenseNet201-Base & 95.33 & 96.07 & 96.11 & 96.10 \\
ResNet50V2-Base  & 93.66 & 93.30 & 93.29 & 93.40 \\
ResNet101V2-Base & 93.13 & 92.67 & 92.68 & 92.78 \\
ResNet152V2-Base & 95.19 & 94.90 & 94.86 & 95.10 \\
MobileNetV1-Base & 93.82 & 93.43 & 93.37 & 93.62 \\
MobileNetV2-Base & 92.60 & 92.20 & 92.30 & 92.23 \\
VGG16-Base       & 90.61 & 89.93 & 89.85 & 90.80 \\
VGG19-Base       & 89.47 & 88.84 & 88.91 & 88.93 \\
InceptionV3-Base & 92.29 & 91.84 & 91.85 & 91.91 \\
Xception-Base    & 91.99 & 91.47 & 91.48 & 91.52 \\
Vit-B16-Base     & 93.52 & 93.09 & 93.16 & 93.53 \\
Vit-B32-Base     & 95.73 & 95.65 & 95.64 & 95.51 \\
Vit-L16-Base     & 95.12 & 94.93 & 94.97 & 94.63 \\
Vit-L32-Base     & 95.90 & 95.49 & 95.55 & 95.82 \\
Context-aware Student & 98.01 & 97.90 & 97.89 & 97.95 \\
\hline
\end{tabular}
\label{T2}
\end{table}

\subsection{Interpretability analysis of Context-aware Student model}
GRADCAM has used to enhance the interpretability of the CASM by visualizing which parts of the image influence the model predictions. This is crucial in medical image analysis, especially for tasks like brain tumor detection, where understanding the model decision-making process is essential. GRADCAM helps highlight important image areas, showing how context-aware Guided adjustments to the temperature parameter impact the student model learning, thus improving transparency in the model predictions.
\subsubsection{GRADCAM visualization}
In order to better understand the model decision-making process, we employ the Grad-CAM \cite{khan2025multi} visualization technique to highlight the regions of interest that the CASM focuses on when making predictions. Grad-CAM computes gradients of the target class with respect to the final convolutional layer, producing heatmaps that indicate the areas of the input image that contribute most significantly to the classification outcome. To evaluate the impact of the context-aware temperature mechanism on the model's interpretability, we compare the attention maps generated with and without this mechanism. Specifically, we analyze how the dynamic adjustment of attention using the context-aware temperature leads to different heatmaps, as compared to a constant temperature approach. This comparison enables us to evaluate whether the context-aware mechanism enhances the model's ability to focus on the relevant tumor areas, leading to more accurate and interpretable predictions.

By incorporating both qualitative visualizations, this analysis not only enhances our understanding of the model's reliability but also presents the confidence in applying it to clinical tasks.

GRADCAM (Gradient-weighted Class Activation Mapping)  \cite{khan2025multi} is employed to visualize the regions of interest that the CASM focuses on when making its predictions. This technique serves to bridge the gap between model performance and human interpretability by enabling us to understand the specific areas in the input images that influence the decisions of the model. By computing gradients of the target class with respect to the final convolutional layer, GRADCAM highlights the salient features in the input image that contribute most significantly to the classification outcome. This visualization aids in identifying if the model is focusing on relevant regions pertinent to tumor detection, thereby providing insights into its reliability and dependability. By applying GRAD-CAM as a visualization technique, we are not only able to validate the predictions of the model but also enhance its confidence in application for clinical tasks. Fig 11 presents the Kaggle multiclass dataset GRADCAM visualization.
\begin{figure}[h]
    \centering
    \includegraphics[width = 11cm]{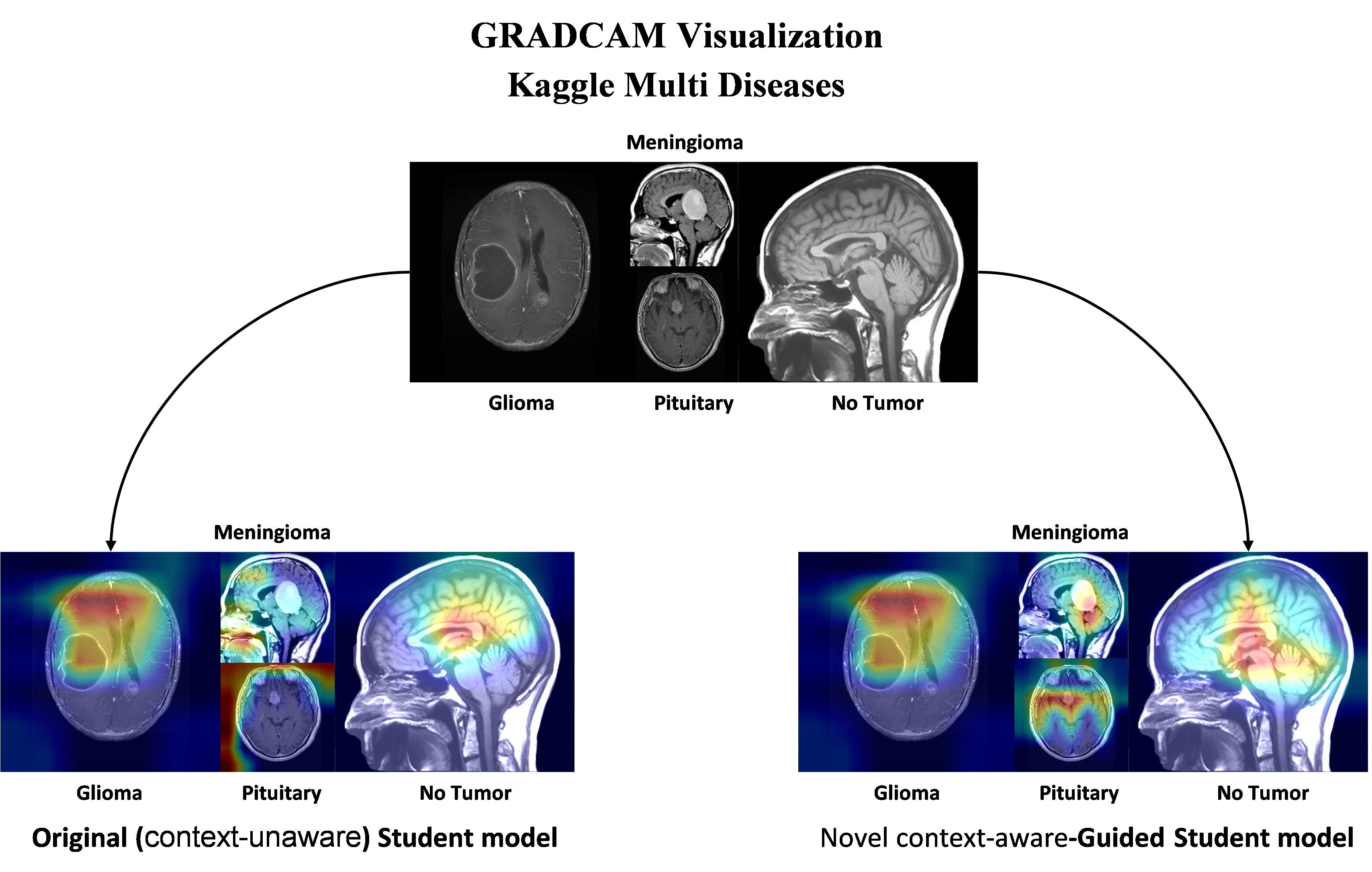}
    \caption{Kaggle Multiclass Dataset GRADCAM Visualization}
    \label{fig:se.png}
\end{figure}
\subsubsection{T-SNE statistical test}
Furthermore, we employed T-SNE for dimensionality reduction and visualization of the high-dimensional feature space \cite{mucca2024dimensionality}, providing insights into the relationships between the feature representations learned by the student model. The statistical tests applied to the T-SNE results helped assess the effectiveness of the Context-aware temperature scaling in improving the model performance as illustrated in Fig 12. So, GRAD-CAM and T-SNE have abled to both interpret the model decisions and analyze its performance in a more comprehensive, data-driven manner, ensuring that the Context-aware-Guided adjustments were enhancing the accuracy and robustness of the student model.
\begin{figure}[!h]
\centering
\begin{minipage}[]{0.46\textwidth}
  \centering
  \includegraphics[width = \textwidth]{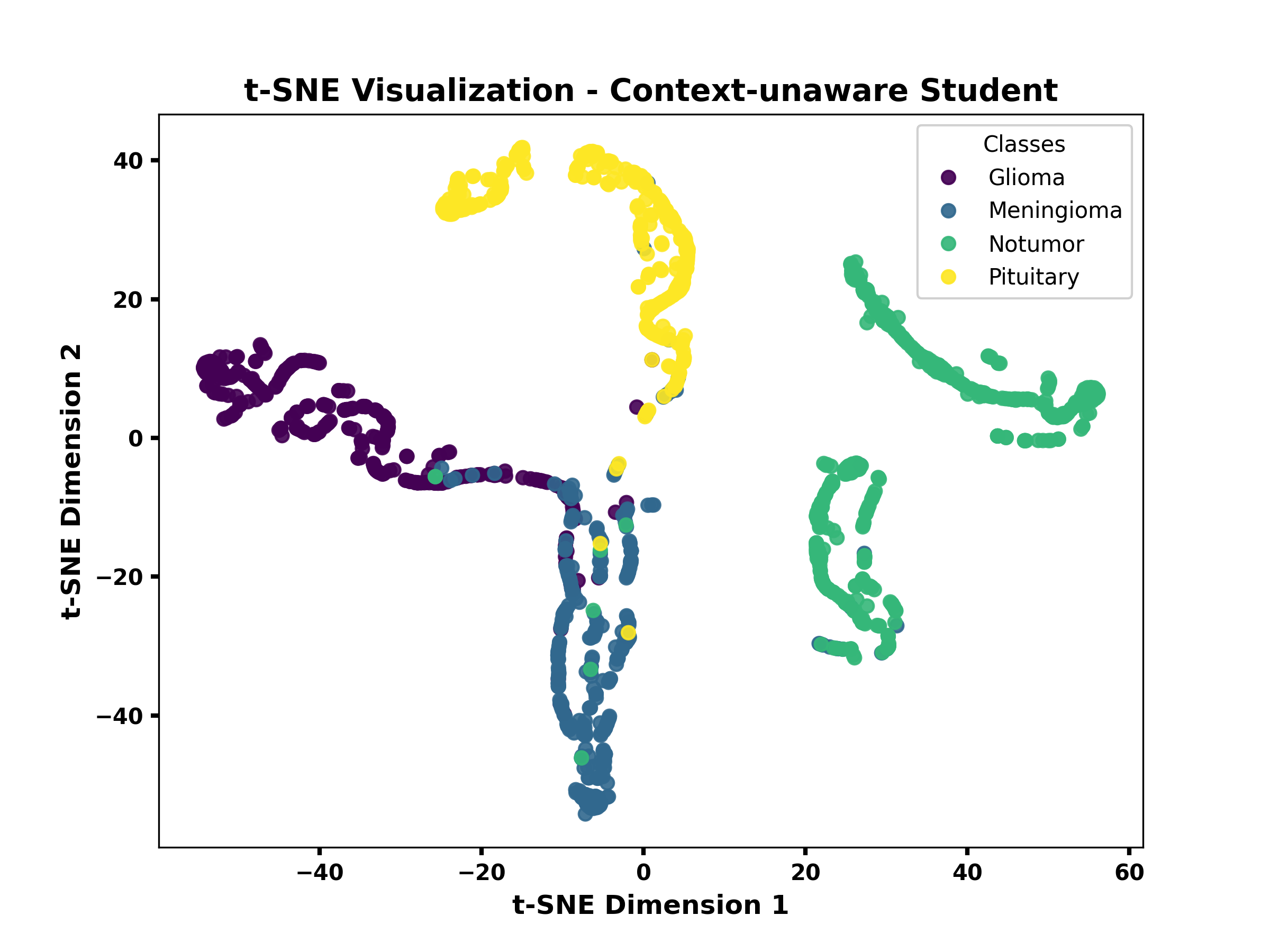}
  
  % Adds some space between the image and text
    % \begin{center}
    % \textbf{Pass density}    
    % \end{center}
\end{minipage}
\begin{minipage}[]{0.46\textwidth}
  \centering
  \includegraphics[width = \textwidth]{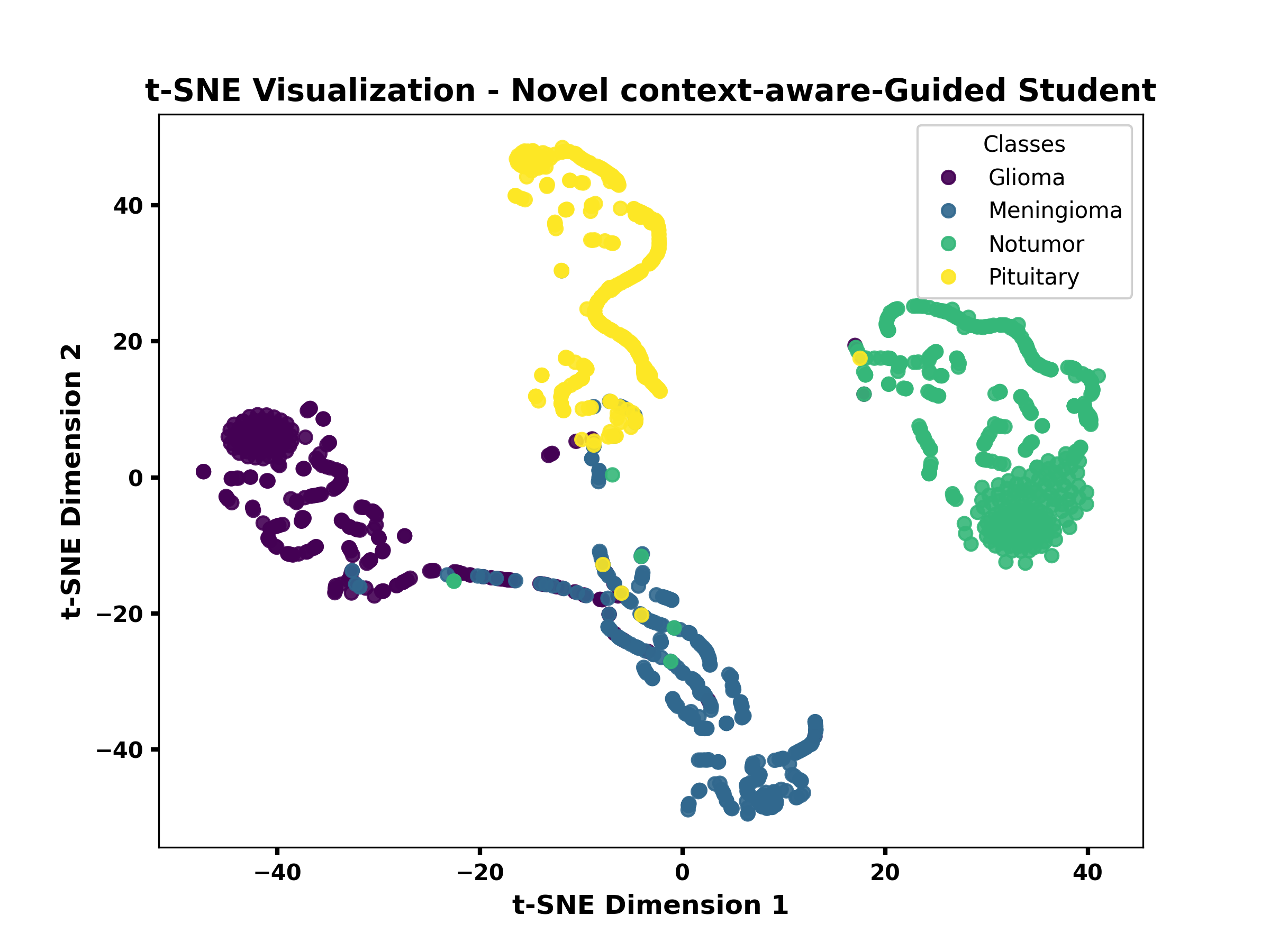}
 
 % Adds some space between the image and text
    % \begin{center}
    % \textbf{Duel density}    
    % \end{center}
\end{minipage}
  \caption{T-SNE performance analysis with:  (a) Constant temperature and (b) Context-aware Adaptive temperature: Kaggle Multiclass}
  \label{fig:11.png}
\end{figure}
\subsection{Performance evaluation of Context-aware Student model with previous SOTA models}
The comparative analysis of the proposed CASM against various existing state-of-the-art (SOTA) methods on the Kaggle Multiclass dataset is presented in Table \ref{T3}. Several existing methods, such as CNN \cite{minarno2021convolutional} models, Deep CNN \cite{srinivas2022deep} models, and Geometric Mean Ensemble demonstrated \cite{tariq2025transforming} similar accuracy performance of 96.00. The Cross Bayesian attention with correlated inception and residual learning (CB-CIRL-Net) \cite{vijayalakshmi2025cross} achieves slightly higher performance, with an accuracy of 96.28. The discrete cosine transformation (DCT) \cite{chen2024robust} based Fusion method demonstrates strong performance at 97.24 but still did not surpass the performance of the proposed method. Our proposed CASM achieves a competitive accuracy of 97.95 by outperforming the existing methods. The higher performance of the proposed method offers a strong and effective solution for the multiclass classification problem.
\begin{table}[ht]
\centering
\caption{Comparison of Context-aware with existing SOTA model: Kaggle Multiclass}
\begin{tabular}{llc}
\hline
\textbf{References} & \textbf{Methods} & \textbf{Accuracy Performance (\%)} \\
\hline
\cite{minarno2021convolutional} & CNN models & 96.00 \\
\cite{chen2024robust} & DCT-based Fusion & 97.24 \\
\cite{srinivas2022deep} & Deep CNN models & 96.00 \\
\cite{vijayalakshmi2025cross} & CB-CIRL Net & 96.28 \\
\cite{tariq2025transforming} & Geometric Mean Ensemble & 96.00 \\
Our & Proposed CASM & 97.95 \\
\hline
\end{tabular}
\label{T3}
\end{table}
\subsection{Generalize performance evaluation of Context-aware Student model with Additional test}
Validating the performance of the CASM with additional tests, such as intra-test using other multi-class MRI dataset and generalized testing with a gastro dataset, is crucial for ensuring the robustness and adaptability of the model across different applications and domains \cite{zhao2024digan}. Intra-test validation using multi-class MRI data allows for assessing how well the model performs when dealing with complex and varied medical images, ensuring that the Context-aware based temperature scaling mechanism can handle different levels of uncertainty and image quality within the same domain. On the other hand, generalized testing with a gastro dataset helps evaluate the model ability to transfer its learned knowledge to a new, distinct application. This ensures that the model does not overfit to the specific features of the original dataset and can generalize effectively across different types of medical data. Overall, these additional tests help demonstrate the model flexibility, reliability, and potential for real-world deployment across various medical imaging tasks.
\subsubsection{Dataset Collection}
Fig 13 displays sample images from Figshare brain tumor \cite{khan2025optimize} dataset. In a standard brain MRI image, healthy brain structures appear normal. However, when dealing with brain tumors, we observe changes in tissue density, manifesting as varying signal intensities in the images. These changes often include tumor masses, areas of consolidation, and variations in texture. These anomalies tend to localize within specific brain regions, frequently affecting particular intellectual parts. To maintain the fair evaluation, we have utilized the same dataset distribution. Three primary categories of brain tumors include Figshare dataset. Among these, Glioma tumors account for a total of 1283 images, Meningioma tumors contribute to 566 cases, and Pituitary tumors make up 714 cases. A comprehensive GI endoscopic image dataset was created by combining curated gastrointestinal imaging datasets \cite{khan2025optimize}. The proposed model for identifying GI ailments was trained and tested using this dataset. In order to depict various infection classifications, the dataset 396 images were divided into three groups of images such as Colorectal Cancer (139), Esophagitis (107), and Pylorus (150).
\begin{figure}[h]
    \centering
    \includegraphics[width = 12cm]{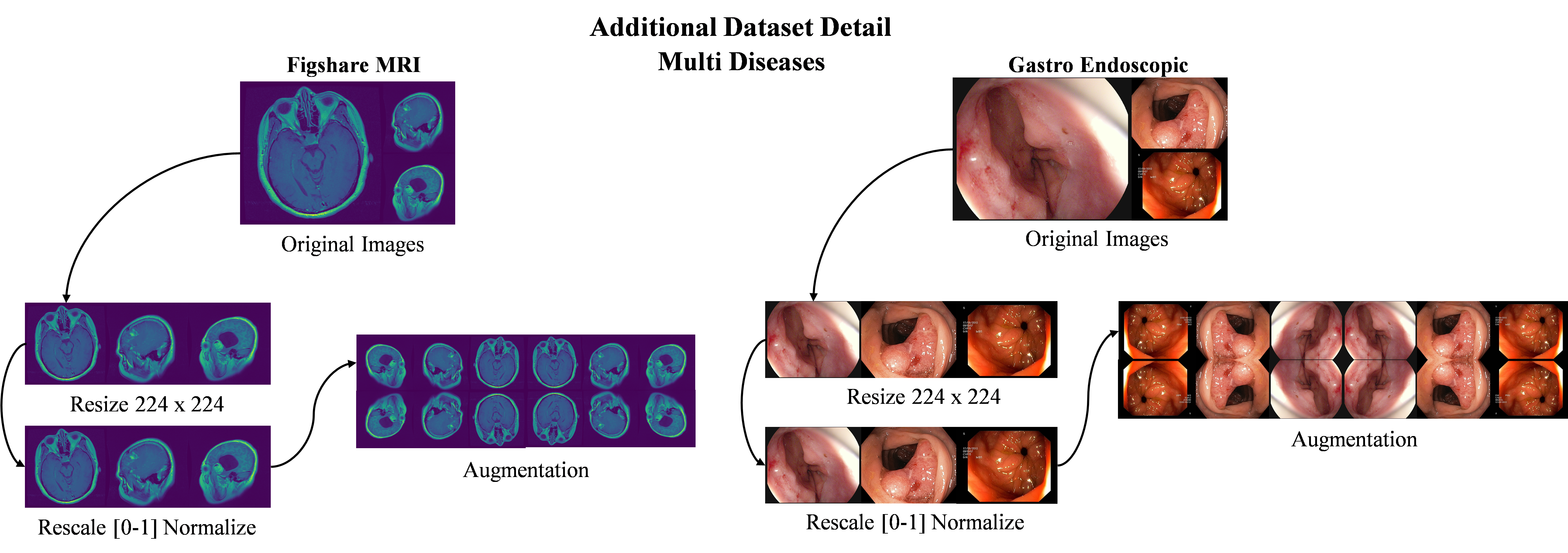}
    \caption{Additional multiclass dataset sample visualization}
    \label{fig:se.png}
\end{figure}
\subsection{Classification performance of Context-aware Student model: Figshare Multi MRI}
This section delves into the class-wise performance analysis of the CASM on the Figshare Multi class MRI dataset by comparing it with the original (context-unaware) student model. Table \ref{T4} presents the key performance indicators for both models across three types of tumor classes. The original (context-unaware) student model in (a) achieves an overall accuracy of 90.03. For Glioma, it exhibits 85.32 precision score, 83.21 recall value, and an 84.32 F1-score. Meningioma classification yields 84.21 precision value, 85.23 recall score, and an 86.47 F1-score, while Pituitary tumor detection results in 92.26 precision score, 90.25 recall value, and a 94.25 F1-score. In comparison to the original student model with a context-unaware temperature, the CASM in (b) demonstrates improved performance with an overall accuracy of 92.81. For the Glioma class, both precision and F1-score have increased by achieving values of 88.97 and 85.40, respectively, although the model shows a slightly lower recall score for this class. Regarding Meningioma, although the F1-score experiences a slight reduction, the precision and recall values are improved to 89.18 and 88.01, respectively. Notably, the CASM in Pituitary tumors exhibits substantial improvements across all key evaluation metrics, with all values exceeding to 95.00. These results underscore the efficacy of the dynamic temperature adjustment introduced by the context-aware rule-based approach by leading to a significant performance enhancement for the Figshare MRI dataset.
\begin{table}[ht]

\centering
\caption{Class wise performance analysis: (a) Constant temperature and (b) Context-aware Adaptive temperature: Figshare Multiclass}
\begin{tabular}{llcccc}
\hline
\textbf{Model} & \textbf{Class} & \textbf{Precision} & \textbf{Recall} & \textbf{F1-Score} & \textbf{Accuracy} \\
\hline
\multirow{3}{*}{Original (context-unaware) Student} & Glioma     & 85.32 & 83.21 & 84.32 & 90.03 \\
                                              & Meningioma & 84.21 & 85.23 & 86.47 &      \\
                                              & Pituitary  & 92.26 & 90.25 & 94.25 &      \\
\hline
\multirow{3}{*}{Context-aware Student}    & Glioma     & 88.97 & 82.11 & 85.40 & 92.81 \\
                                              & Meningioma & 89.18 & 88.01 & 83.33 &      \\
                                              & Pituitary  & 95.26 & 97.31 & 96.28 &      \\
\hline
\end{tabular}
\label{T4}
\end{table}
The confusion matrix visualizations for the context-unaware temperature student model and the dynamically adjusted context-aware rule model focus on Figshare Multiclass dataset are depicted in Fig 14, which comprises a total of 612 test images. In Fig 14 (a), the context-unaware Temperature student model correctly classified 551 samples but misclassified 61 by indicating a need for improvement in its classification capabilities. In contrast, Fig 14 (b) illustrates the performance of the dynamically adjusted context-aware rule model, which successfully detected 568 samples, thereby reducing the misclassification error to only 44 samples. This significant reduction in misclassifications demonstrates the effectiveness of the context-aware rule based approach in enhancing the capability of the model to accurately classify tumor types within the dataset.
\begin{figure}[!h]
\centering
\begin{minipage}[]{0.46\textwidth}
  \centering
  \includegraphics[width = \textwidth]{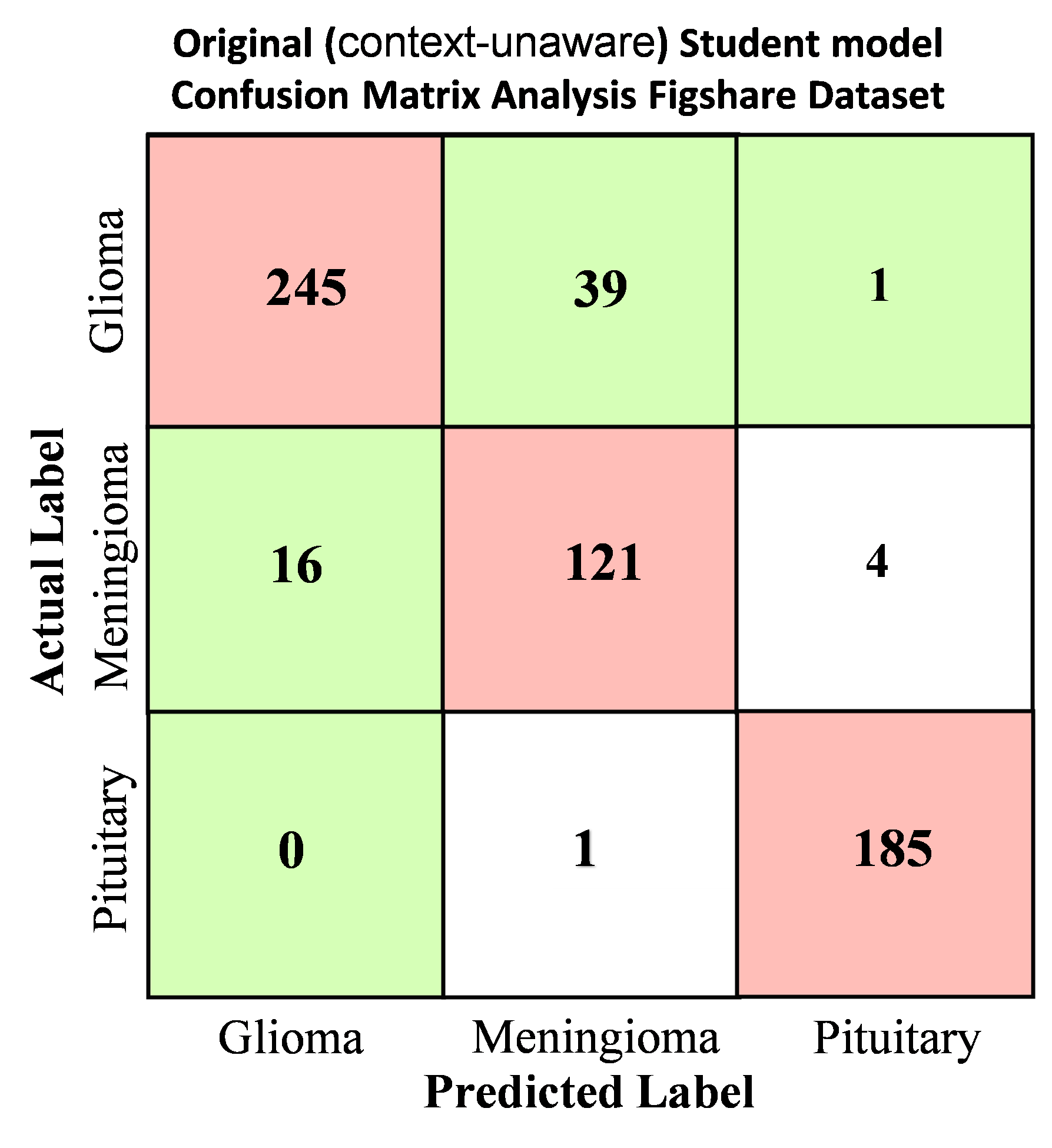}
  
  % Adds some space between the image and text
    % \begin{center}
    % \textbf{Pass density}    
    % \end{center}
\end{minipage}
\begin{minipage}[]{0.46\textwidth}
  \centering
  \includegraphics[width = \textwidth]{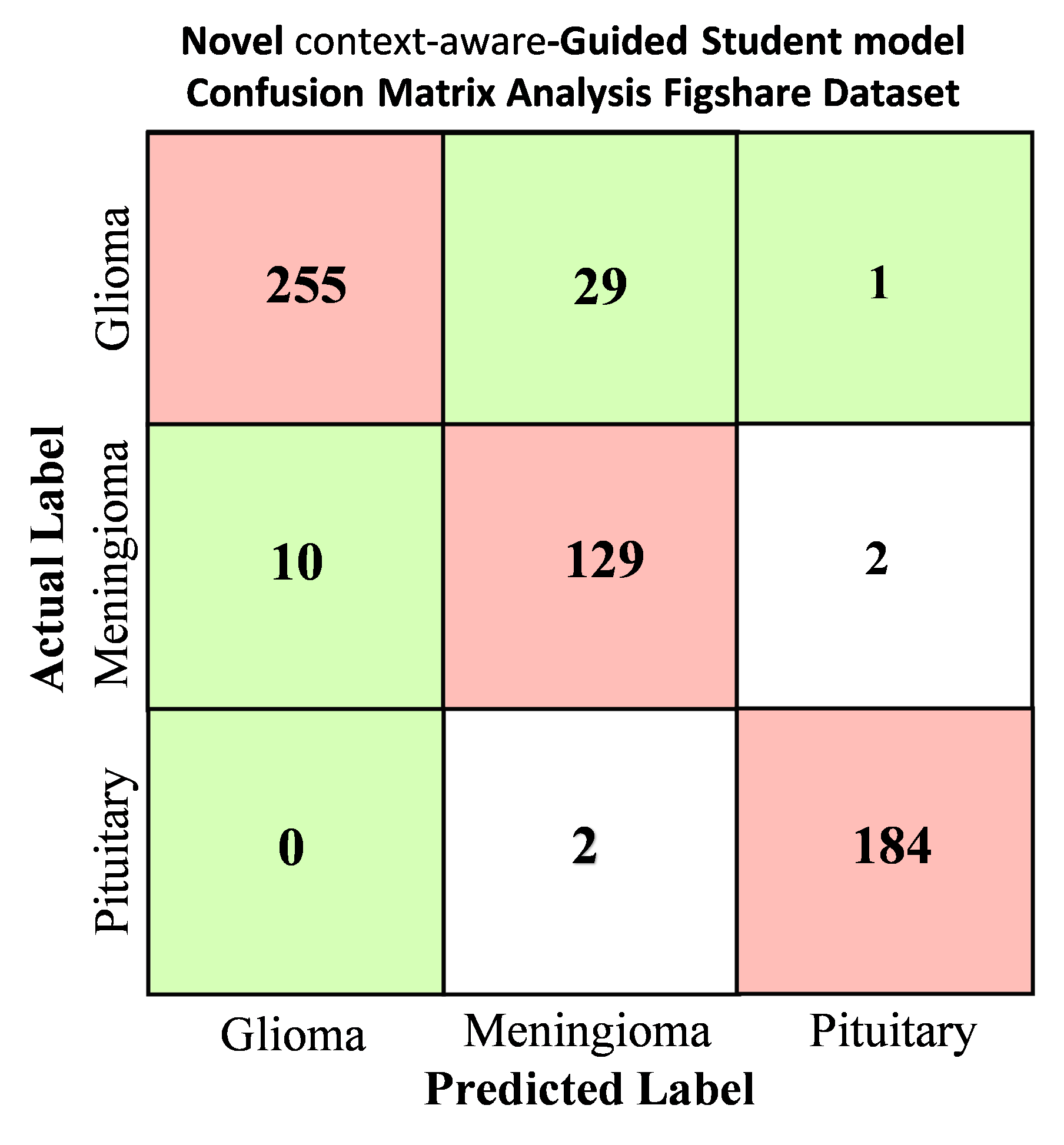}
 
 % Adds some space between the image and text
    % \begin{center}
    % \textbf{Duel density}    
    % \end{center}
\end{minipage}
  \caption{Confusion matric visualization: (a) Constant temperature and (b) Context-aware Adaptive temperature: Figshare Multiclass}
  \label{fig:11.png}
\end{figure}\\
The performance of the Original (context-unaware) Student model and the CASM is analyzed on the Figshare Multiclass dataset using AUC-ROC and PR curves, as depicted in Fig 15. The AUC-ROC curves in Fig 15(a) show that the Context-aware model achieves a superior AUC value of 0.9626 compared to the original student model showing 0.9570 AUC value, indicating a marginal improvement in overall performance. The PR curves in Fig 15(b) illustrate a more noticeable difference, with the Context-aware model achieving a significantly higher AP value of 0.9335 compared to the AP value of original model at 0.9218. This suggests that the Context-aware model exhibits a better balance between precision and recall, particularly at higher thresholds. These results highlight the advantages of the dynamic temperature adjustment in the Context-aware model for the Figshare Multiclass dataset.
\begin{figure}[!h]
\centering
\begin{minipage}[]{0.46\textwidth}
  \centering
  \includegraphics[width = \textwidth]{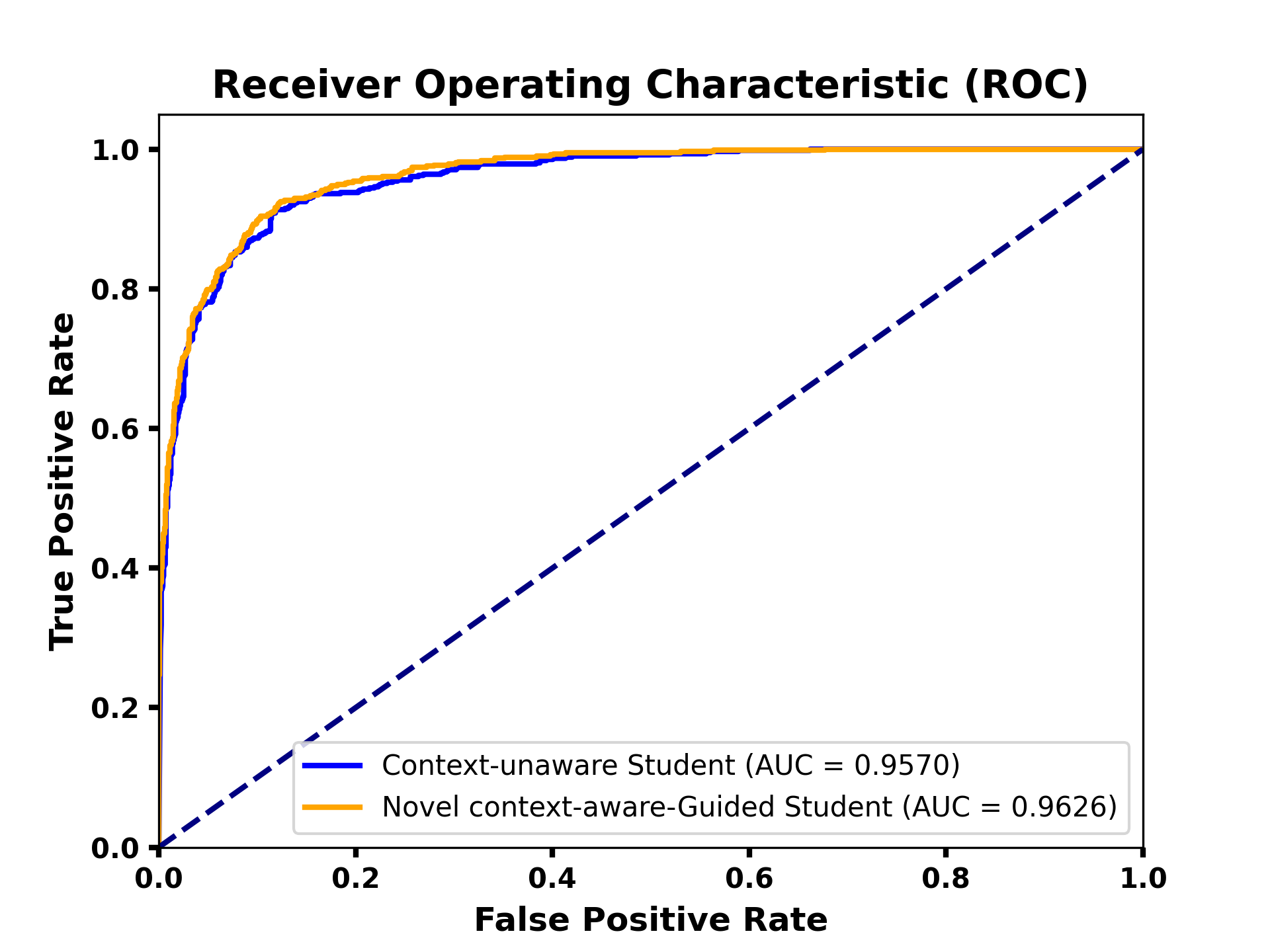}
  
  % Adds some space between the image and text
    % \begin{center}
    % \textbf{Pass density}    
    % \end{center}
\end{minipage}
\begin{minipage}[]{0.46\textwidth}
  \centering
  \includegraphics[width = \textwidth]{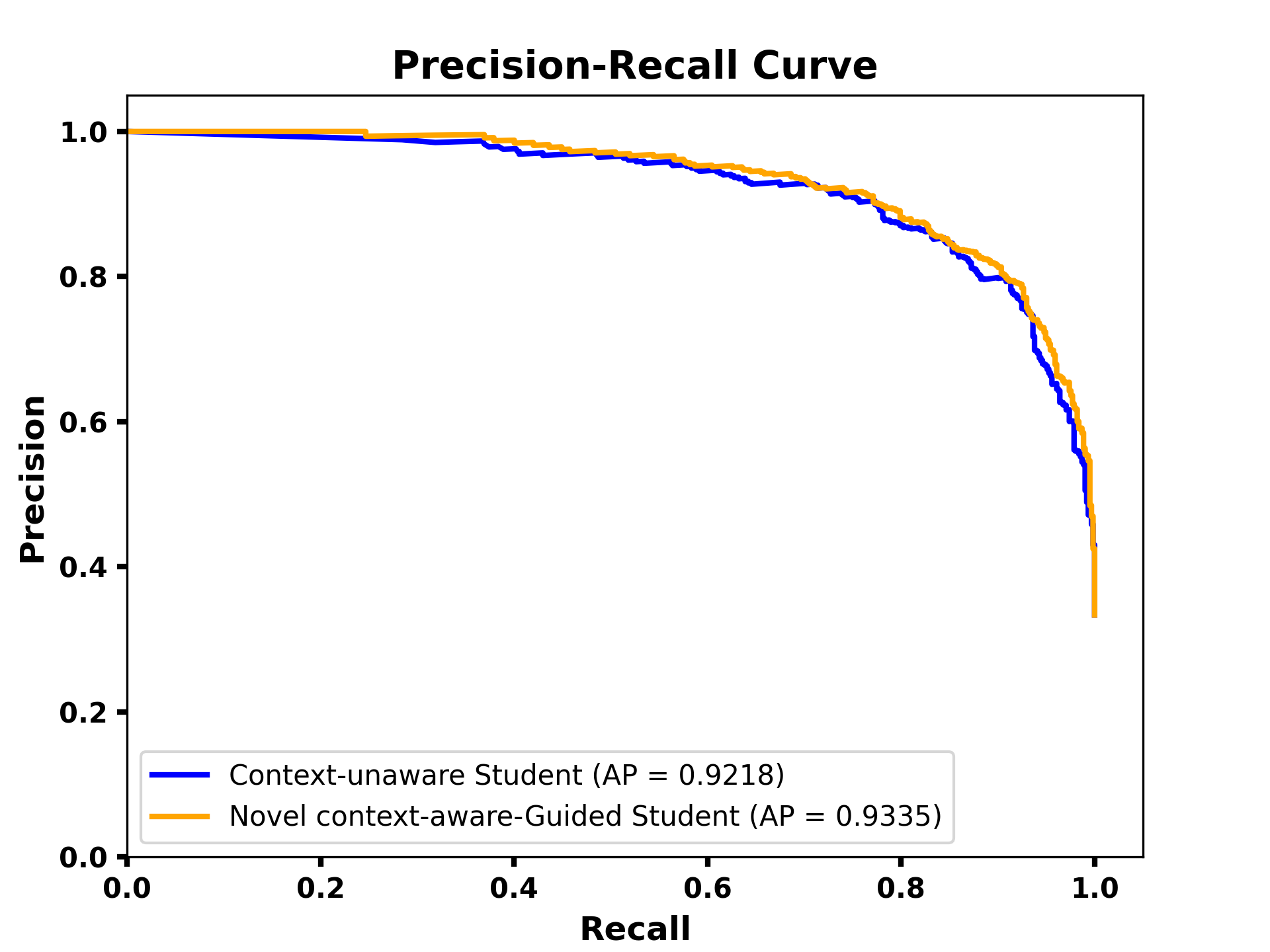}
 
 % Adds some space between the image and text
    % \begin{center}
    % \textbf{Duel density}    
    % \end{center}
\end{minipage}
  \caption{Performance analysis with: (a) AUC-ROC curve and (b) PR Curve: Figshare Multiclass}
  \label{fig:11.png}
\end{figure}
The performance comparison of the proposed CASM with individual pre-trained models is summarized in Table \ref{T5}. We have selected the same teacher and student models used in the primary dataset experimental case. DenseNet201 is considered as the teacher model, and it achieves the highest accuracy among all pre-trained models at 89.86. The ResNet152V2 student model performs slightly lower, while DenseNet169 ranks as the second-highest performing model after DenseNet201. Overall, the accuracy achieved by our proposed CASM is the highest among all models by reaching 92.81. The performance of the other models falls within the range of 80.90 to 85.78, which is lower than that of the proposed model and the selected teacher and student models. A similar trend is observed in the recall scores, where the recall value of the proposed method is comparable to that of the DenseNet201 teacher model, but the overall accuracy of the proposed model exceeds that of the teacher model. In terms of precision, both the selected teacher and student models remain the top performers among the other pre-trained models, with scores of 86.10 and 84.73, respectively. However, they still do not surpass the performance of the proposed CASM, which achieves an impressive precision score of 91.65. A similar trend is observed in the F1-score, where the results align closely with the observed precision and accuracy performance scores, further demonstrating the effectiveness of the proposed method.
\begin{table}[ht]
\centering
\caption{Comparison of Context-aware with existing pre-trained: Figshare Multiclass}
\begin{tabular}{lcccc}
\hline
\textbf{Method} & \textbf{Accuracy} & \textbf{F1-Score} & \textbf{Recall} & \textbf{Precision} \\
\hline
DenseNet121-Base & 85.78 & 85.00 & 85.81 & 84.47 \\
DenseNet169-Base & 88.64 & 84.23 & 86.18 & 83.93 \\
DenseNet201-Base & 89.86 & 89.07 & 89.11 & 86.10 \\
ResNet50V2-Base  & 81.37 & 79.79 & 79.77 & 79.81 \\
ResNet101V2-Base & 82.18 & 80.79 & 81.26 & 80.41 \\
ResNet152V2-Base & 86.80 & 86.27 & 86.06 & 84.73 \\
MobileNetV1-Base & 81.90 & 79.34 & 80.62 & 79.02 \\
MobileNetV2-Base & 83.98 & 82.19 & 82.13 & 82.43 \\
VGG16-Base       & 84.80 & 84.58 & 86.84 & 84.52 \\
VGG19-Base       & 83.82 & 82.85 & 84.11 & 82.14 \\
InceptionV3-Base & 81.69 & 80.68 & 81.19 & 80.35 \\
Xception-Base    & 80.90 & 75.60 & 75.68 & 78.53 \\
Context-aware Student & 92.81 & 91.07 & 89.11 & 91.65 \\
\hline
\end{tabular}
\label{T5}
\end{table}
\subsubsection{Classification performance of Context-aware Student model: Gastro Multi Endoscopic}
The class-wise performance analysis comparing the Original (context-unaware) Student model and the dynamically adjusted CASM on the Gastro Multiclass dataset is discussed in Table \ref{T6}. The Original student model achieves an overall accuracy of 94.94 by demonstrating perfect score in all key evaluation metrices for Colorectal Cancer class. For Esophagitis, it achieves 90.00 precision with a lower recall score of 88.26 by resulting an F1-score of 91.26, while for Pylorus class, it shows precision value of 93.45, recall value of 96.65, and the value of F1-score at 95.10. The Context-aware model demonstrates a +1.26 improvement in overall accuracy over the original student model by reaching 96.20 value. It maintains perfect performance for Colorectal Cancer class and exhibits increased precision and F1-score for Esophagitis with values 95.00 and 92.68, respectively. It shows a lower recall rate but still higher than the original student model. While the F1-score score is slightly reduced from 95.10 to 95.08 in the context-aware predictor model, but the precision and recall are still improved with values reach to 93.55 and 96.67, respectively. The improved overall accuracy highlights the benefits of the context-aware based adjustment in the proposed model.
\begin{table}[ht]
\centering
\caption{Class wise performance analysis: (a) Constant temperature and (b) Context-aware Adeptvie Temperature: Gastro Multiclass}
\begin{tabular}{llcccc}
\hline
\textbf{Dataset} & \textbf{Class} & \textbf{Precision} & \textbf{Recall} & \textbf{F1-Score} & \textbf{Accuracy} \\
\hline
\multirow{3}{*}{Original (context-unaware) Student} & Colorectal Cancer & 1.000 & 1.000 & 1.000 & 94.94 \\
& Esophagitis & 90.00 & 88.26 & 91.26 &  \\
& Pylorus & 93.45 & 96.65 & 95.10 &  \\
\hline
\multirow{3}{*}{Context-aware Student} & Colorectal Cancer & 1.000 & 1.000 & 1.000 & 96.20 \\
& Esophagitis & 95.00 & 90.48 & 92.68 &  \\
& Pylorus & 93.55 & 96.67 & 95.08 &  \\
\hline
\end{tabular}
\label{T6}
\end{table}
The confusion matrix visualizations for both the context-unaware temperature original student model and the dynamically adjusted context-aware rule based model on the Gastro Multiclass endoscopic images dataset are illustrated in Fig 16. This dataset comprises 79 test images, which are categorized into three classes including colorectal cancer, pylorus, and esophagitis. Fig 16(a) visualizes the confusion matrix of original student model with accurately classifying 75 samples but made 4 misclassifications. In contrary, Fig 16(b) demonstrates the confusion matrix of the dynamically adjusted context-aware rule model, which correctly identified 76 samples by reducing misclassifications to only 3. This significant reduction in misclassification errors underscores the effectiveness of the context-aware rule-based approach in improving the classification accuracy of proposed model on gastro dataset.
\begin{figure}[!h]
\centering
\begin{minipage}[]{0.46\textwidth}
  \centering
  \includegraphics[width = \textwidth]{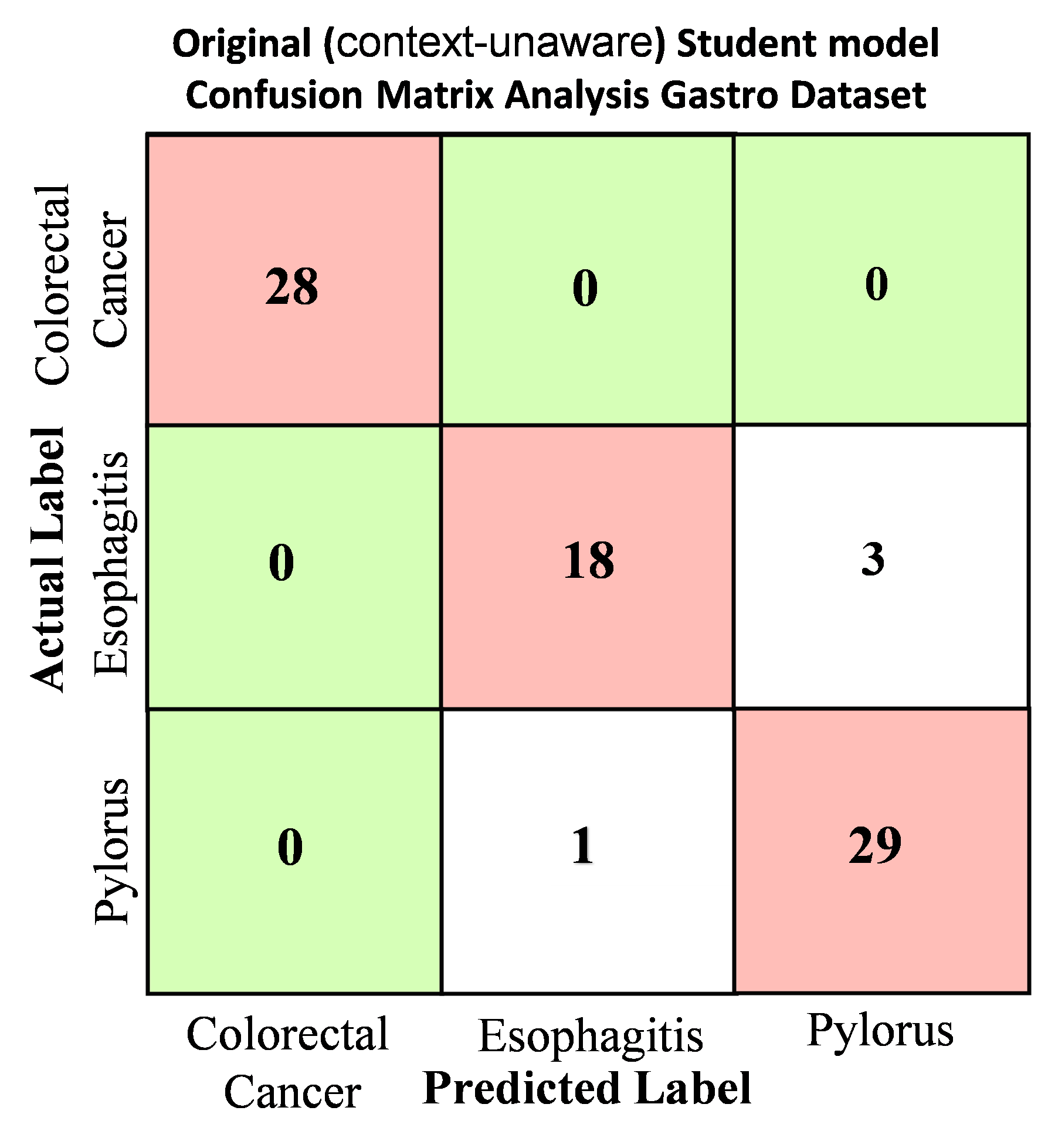}
  
  % Adds some space between the image and text
    % \begin{center}
    % \textbf{Pass density}    
    % \end{center}
\end{minipage}
\begin{minipage}[]{0.46\textwidth}
  \centering
  \includegraphics[width = \textwidth]{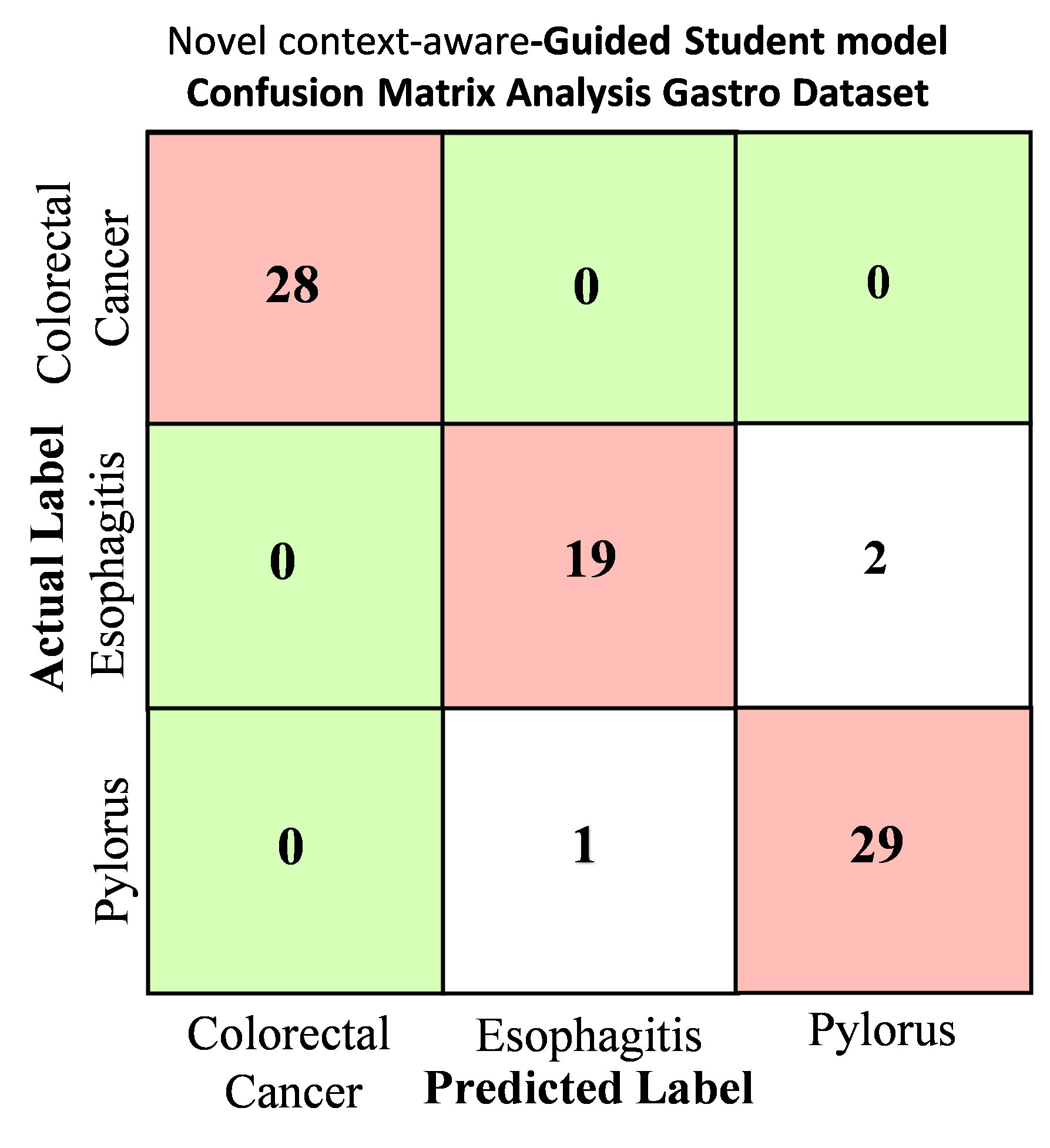}
 
 % Adds some space between the image and text
    % \begin{center}
    % \textbf{Duel density}    
    % \end{center}
\end{minipage}
  \caption{Confusion matric visualization: (a) Constant temperature and (b) Context-aware Adaptive temperature: Gastro Multiclass}
  \label{fig:11.png}
\end{figure}\\
The AUC-ROC and PR curves for original (context-unaware) Student model and the CASM on the Gastro Multiclass dataset are further analyzed in Fig 17. The AUC-ROC curves in Fig 17(a) reveal that both models achieve high AUC values, with the original student model at 0.9737 and the Context-aware model slightly higher at 0.9803. The PR curves in Fig 17(b) demonstrate a similar trend, with both models exhibiting strong performance. The Original model achieves an AP value of 0.9248 while the Context-aware model achieves a notably higher score of 0.9477 by indicating better performance in balancing precision and recall across varying thresholds. This further reinforces the effectiveness of the context-aware rule-based approach.
\begin{figure}[!h]
\centering
\begin{minipage}[]{0.46\textwidth}
  \centering
  \includegraphics[width = \textwidth]{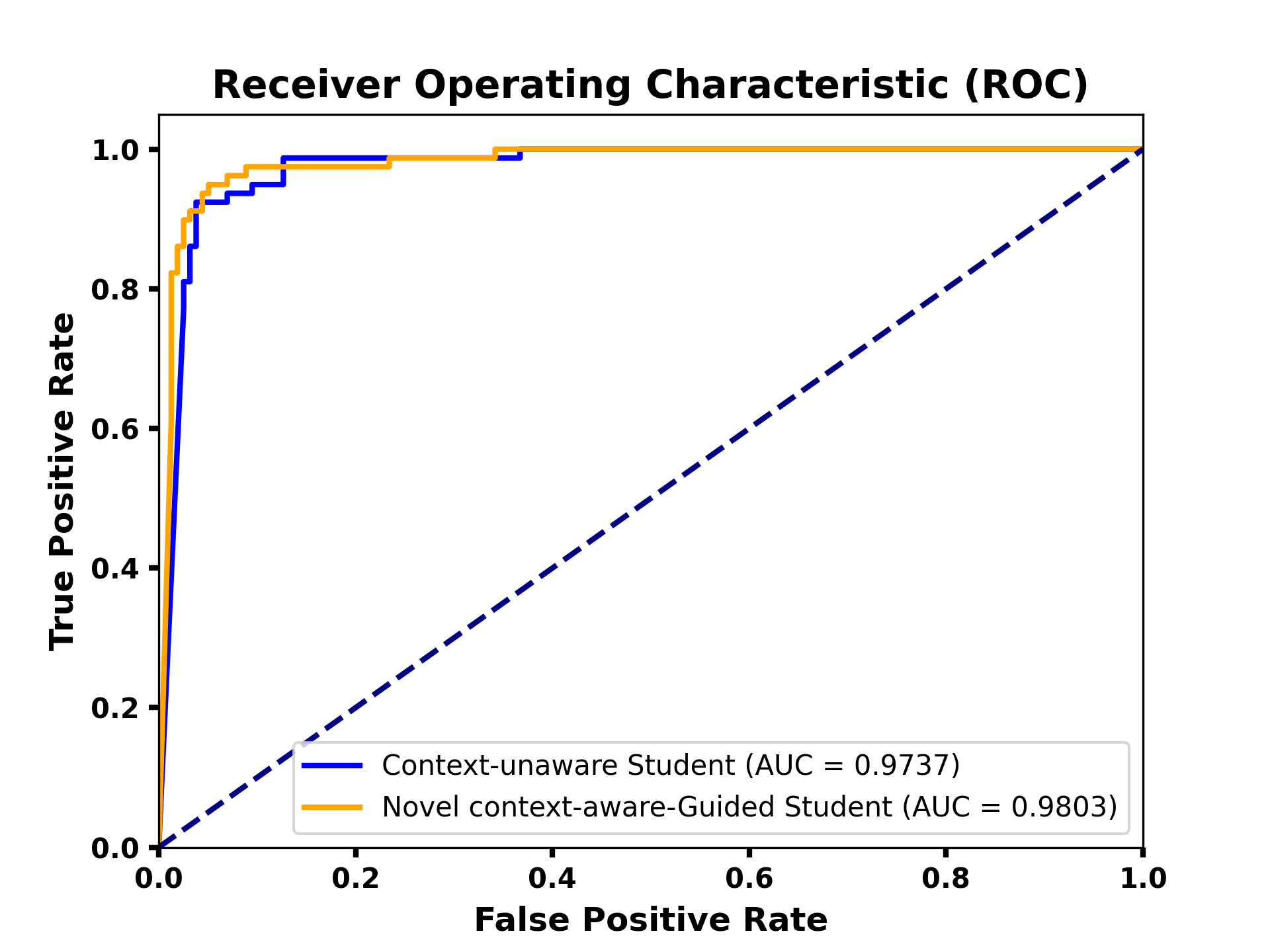}
  
  % Adds some space between the image and text
    % \begin{center}
    % \textbf{Pass density}    
    % \end{center}
\end{minipage}
\begin{minipage}[]{0.46\textwidth}
  \centering
  \includegraphics[width = \textwidth]{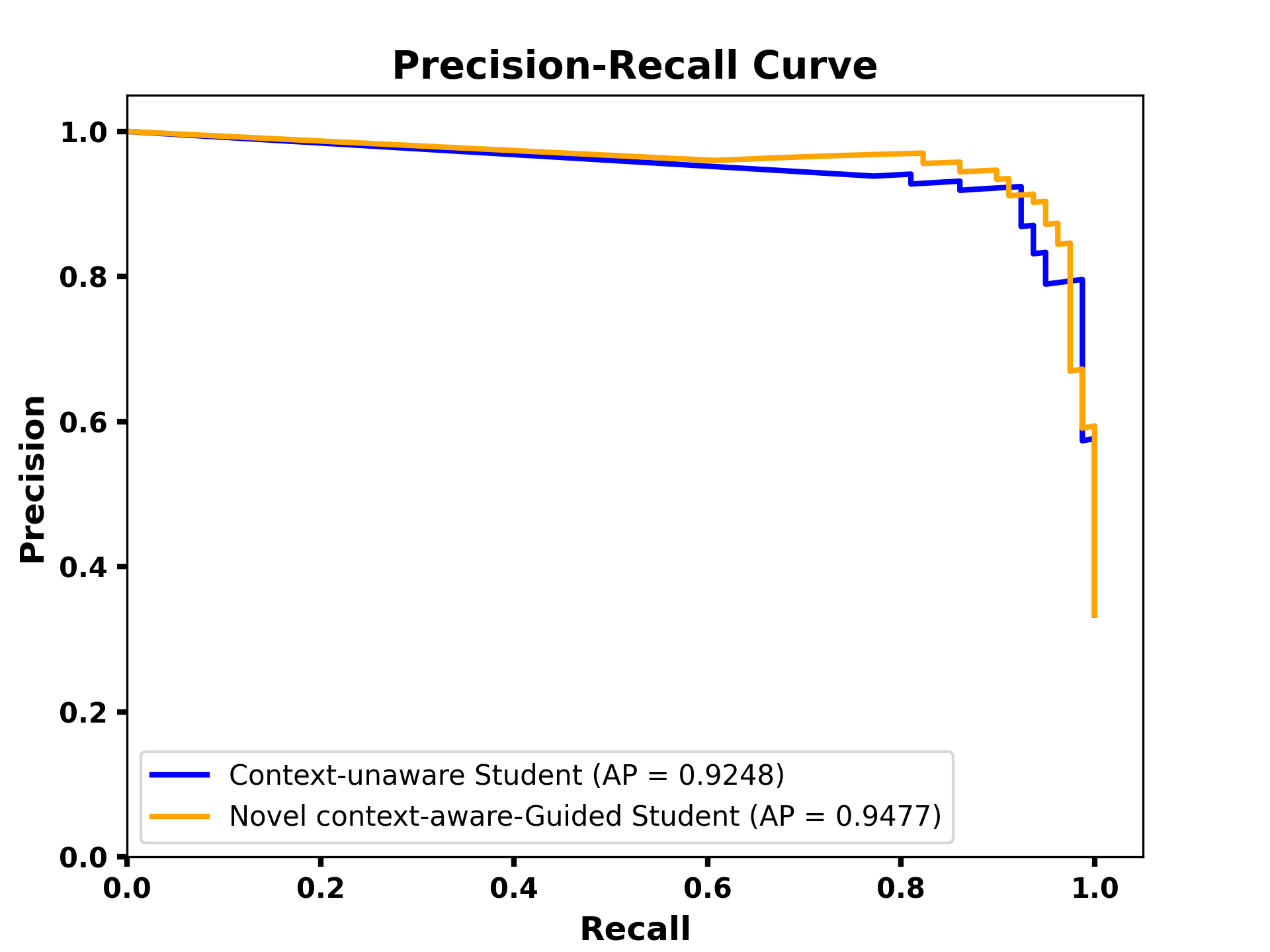}
 
 % Adds some space between the image and text
    % \begin{center}
    % \textbf{Duel density}    
    % \end{center}
\end{minipage}
  \caption{Performance analysis with: (a) AUC-ROC curve and (b) PR Curve: Gastro Multiclass}
  \label{fig:11.png}
\end{figure}
The performance of CASM with existing pre-trained models on the Gastro Multiclass dataset is detailed in Table \ref{T7}. Among the pre-trained models, DenseNet201 demonstrates the highest accuracy at 93.67, precision at 93.44, and recall at 93.01 by making it the top-performing teacher model. ResNet152V2 follows as the second-best performing model and it is chosen as student model with an accuracy of 92.40, precision of 92.12, and recall of 92.30. The other pre-trained models exhibit lower performance across all these key metrics. However, the proposed CASM surpasses all individual pre-trained models by achieving an accuracy of 96.20, precision of 96.18, and recall of 95.71. The F1-scores exhibit a similar trend to precision and recall scores, as the F1 score is defined as the harmonic mean of precision and recall by providing a single metric that balances both metrics in assessing model performance. This balance reflects the superior performance of the proposed method across both precision and recall metrics by leading to its outperformance over all other models in the evaluation. This superior performance reinforces the efficacy of the context-aware rule-based approach in enhancing classification performance on the Gastro Multiclass dataset.
\begin{table}[ht]
\centering
\caption{Comparison of Context-aware with existing pre-trained: Gastro Multiclass}
\begin{tabular}{lcccc}
\hline
\textbf{Method} & \textbf{Accuracy} & \textbf{F1-Score} & \textbf{Recall} & \textbf{Precision} \\
\hline
DenseNet121-Base & 91.13 & 90.72 & 91.26 & 90.58 \\
DenseNet169-Base & 91.14 & 90.44 & 89.94 & 91.83 \\
DenseNet201-Base & 93.67 & 93.20 & 93.01 & 93.44 \\
ResNet50V2-Base  & 88.60 & 88.24 & 88.49 & 88.15 \\
ResNet101V2-Base & 86.07 & 85.51 & 86.34 & 85.44 \\
ResNet152V2-Base & 92.40 & 92.18 & 92.30 & 92.12 \\
MobileNetV1-Base & 84.81 & 84.41 & 84.76 & 85.11 \\
MobileNetV2-Base & 82.27 & 81.69 & 83.01 & 82.81 \\
VGG16-Base       & 84.81 & 84.16 & 85.23 & 84.84 \\
VGG19-Base       & 84.81 & 83.67 & 83.80 & 83.57 \\
InceptionV3-Base & 89.87 & 89.59 & 89.04 & 90.60 \\
Xception-Base    & 86.07 & 85.49 & 86.82 & 86.38 \\
Context-aware Student & 96.20 & 95.92 & 95.71 & 96.18 \\
\hline
\end{tabular}
\label{T7}
\end{table}
Fig 18 demonstrates the GRADCAM visualization of the proposed model applied to both MRI and gastro images. In Fig 18, the GRADCAM heatmaps for MRI images highlight the key regions in the mid and cross-sectional views that the model focuses on for classification. These heatmaps indicate how the model attends to critical anatomical structures, emphasizing areas that are crucial for accurate diagnosis. Similarly, in Fig 18, GRADCAM is applied to gastro images, where the heatmaps reveal the regions of the endoscopic images that are most influential in the model decision-making process.
\begin{figure}[h]
    \centering
    \includegraphics[width = 12cm]{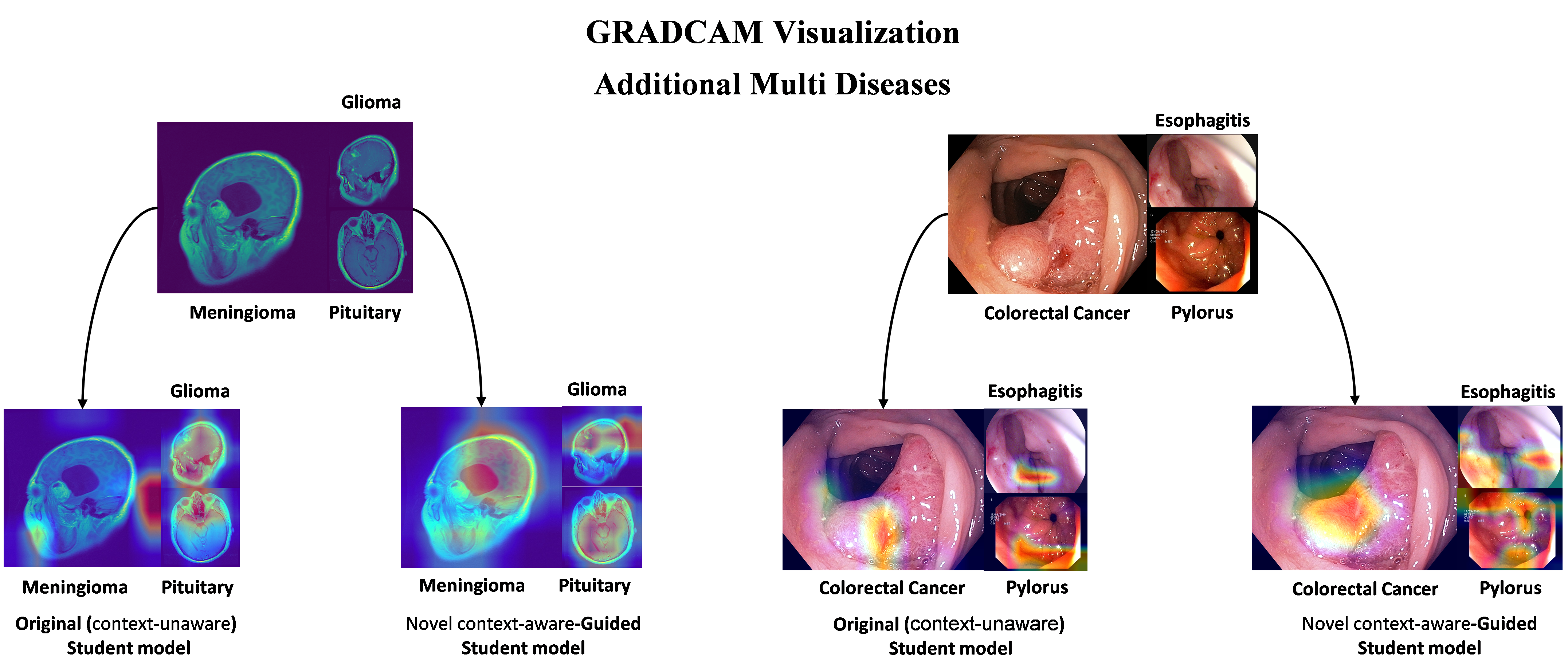}
    \caption{Additional Multiclass Dataset GRADCAM Visualization}
    \label{fig:se.png}
\end{figure}

\subsection{Ablation Study}
To assess the effectiveness of using  ACO for selecting optimal teacher–student model pairs within the  KD framework, we conducted a comprehensive ablation study \cite{khan2025optimized}. This study aimed to evaluate the potential advantages of ACO-driven strategies in comparison to conventional model selection methods. Specifically, our objective was to determine how well the ACO-based approach enhanced the performance of student models compared to traditional selection techniques.

Table \ref{T10} presents a performance comparison between the proposed context-aware student model and other configurations, including the teacher model, a student model without distillation, and a student model distilled using a constant temperature. The context-aware student model outperforms all others across all metrics—accuracy, F1-score, recall, and precision—achieving the highest accuracy of 98.01\% and F1-score of 97.90\%. In contrast, the student model without distillation shows reduced performance, highlighting the importance of knowledge transfer. Moreover, the distilled student model using a constant temperature performs similarly to the teacher model but still falls short of the context-aware design. These results demonstrate the effectiveness of incorporating context in enhancing the distillation process and improving model performance.
\begin{table}[ht]
\centering
\caption{Performance Comparison of Context-aware model and other configurations}
\begin{tabular}{lcccc}
\hline
\textbf{Method} & \textbf{Accuracy} & \textbf{F1-Score} & \textbf{Recall} & \textbf{Precision}  \\
\hline
Teacher  & 95.33 & 96.07 & 96.11 & 96.10 \\
Student (WO Distillation) & 95.19 & 94.90 & 94.86 & 95.10 \\
Student (Distilled with constant temp) & 96.33 & 96.07 & 96.11 & 96.10 \\
Context-aware Student & 98.01 & 97.90 & 97.89 & 97.95 \\
\hline
\end{tabular}
\label{T10}
\end{table}

Table \ref{T11} presents a performance comparison of the proposed model, "Context-aware Student," under different noise conditions. The model's accuracy is evaluated across four scenarios: with Gaussian noise, Salt and Pepper noise, Uniform noise, and without any added noise. The results show that the highest accuracy of 98.01\% is achieved when no noise is introduced to the model. Among the noise types, the Uniform noise results in the highest accuracy at 96.33\%, followed by the Gaussian noise with 95.90\%. The model performs least well with Salt and Pepper noise, yielding an accuracy of 95.19\%. This comparison demonstrates the impact of different noise types on the performance of the Context-aware Student model.
\begin{table}[ht]
\centering
\caption{Performance Comparison of Proposed Model after Adding Noise}
\renewcommand{\arraystretch}{1.3}
\begin{tabular}{llc}
\hline
\textbf{Method} & \textbf{Approach} & \textbf{Accuracy} \\
\hline
Context-aware Student (Our) & Gaussian Noise & 95.90 \\
Context-aware Student (Our) & Salt and Pepper Noise & 95.19 \\
Context-aware Student (Our) & Uniform Noise & 96.33 \\
Context-aware Student (Our) & Without added Noise & 98.01 \\
\hline
\end{tabular}
\label{T11}
\end{table}

We utilized a diverse pool of 16 pre-trained models, carefully selected to represent a wide range of architectures and learning behaviors. This allowed us to explore 240 possible teacher–student combinations, providing a robust dataset for our experiment. The selection of these models was based on varying complexities, capabilities, and generalization abilities, ensuring that the study covered a broad spectrum of potential scenarios.
To carry out the assessment, we compared three distinct model selection strategies:
\subsubsection{Strategies}
\begin{itemize}
\item \textit{ACO-Based Selection (Our)}: In this approach \cite{wu2024application}, we employed ACO to dynamically search for the best teacher–student model pairings. The algorithm iteratively adjusted model pairings based on feedback from the student model's performance, simulating a biological optimization process to identify the most effective pairings for knowledge transfer.\\
\item \textit{Random Selection:} In contrast, this method randomly \cite{rimal2024accuracy} selected teacher–student pairs from the pool of models. While simple and fast, this strategy did not incorporate any optimization and relied solely on chance.\\
\item \textit{Particle Swarm Optimization:} This approach used predefined heuristics to guide the selection of teacher–student pairs \cite{jain2024exhaustive}. The heuristics considered factors such as performance, and compatibility, but lacked the adaptability and iterative learning process offered by the ACO method.\\
\item \textit{Grid Search:} This method \cite{rimal2024accuracy} exhaustively searches through a manually specified subset of the hyperparameter space. While more systematic than random selection, Grid Search is computationally expensive as it evaluates every possible combination of hyperparameters, which may lead to increased evaluation time without necessarily improving model performance.
\end{itemize}

Each of these strategies was evaluated by comparing the performance of the student models across several key metrics, including accuracy, training efficiency, and generalization. The results of this study were used to determine the potential of ACO for improving the teacher–student selection process and how it stands against conventional methods in the KD framework. The findings will provide valuable insights into the feasibility and advantages of applying optimization techniques like ACO to enhance the KD process.

\begin{table}[ht]
\centering

\caption{Comparison of Model Selection Strategies for Teacher-Student Selection}
\begin{tabular}{lcc}
\hline
\textbf{Selection Method} & \textbf{Student Accuracy (\%)} & \textbf{Model Evaluations} \\
\hline
Random Selection          & 91.12                          & 1                         \\
Particle Swarm Optimization   & 93.45                          & 240                       \\
Grid Search               & 94.87                          & 1000                      \\
\textbf{ACO (Proposed)}   & \textbf{96.33}                 & \textbf{47}               \\
\hline
\end{tabular}
\label{T8}
\end{table}

Table ~\ref{T8} presents that the ACO-based approach achieved the highest accuracy (96.33\%), outperforming both random and exhaustive strategies. While the exhaustive method explored every possible pair, it required a significantly larger number of evaluations. In contrast, ACO achieved better performance with only ~20\% of the evaluations compared to exhaustive search. This efficiency is made possible by ACO's adaptive learning mechanism, which favors high-potential combinations based on iterative pheromone updates. Grid Search, while systematic, resulted in fewer evaluations than Particle Swarm Optimization but was still computationally expensive compared to ACO, achieving a performance of 94.87\%.
\subsection{Clinical application}
\begin{itemize}
    \item Improved Data Management and Interpretation
    \begin{itemize}
    \item Ensuring that medical data is accurately managed and interpreted is a challenge for hospitals and clinics.
    \item The proposed context-aware adaptive student model, with distilled instructor learning, helps improve how medical professionals interpret the model outputs, facilitating better decision-making.
\end{itemize}
 \item Better Performance than Pre-trained Models:
    \begin{itemize}
    \item The model outperforms pre-trained models and traditional knowledge distillation learning, making it more reliable and effective for clinical use.
    \item This allows hospitals to leverage a more advanced tool that could enhance diagnostic accuracy.
\end{itemize}
\item Practical Implementation in Hospital Monitoring Systems:
   \begin{itemize}
    \item The model is designed to be integrated into existing clinical infrastructures, such as hospital monitoring systems, without requiring substantial hardware upgrades.
    \item Its strong yet privacy requirements make it ideal for real-world settings with resource privacy issue limitations due to its Student-Teacher learning.
\end{itemize}
\item Training and Integration:
   \begin{itemize}
    \item Proper training for medical staff and smooth integration of the model into existing systems are key to overcoming any challenges.
    \item The model’s ease of integration ensures minimal disruption to current workflows, making the transition to using the model relatively straightforward
\end{itemize}
\end{itemize}
\subsection{Discussion}

In this study, we introduced the CASM within the KD framework, which leverages a context-aware rule to adaptively adjust the temperature during the distillation process. The model has shown impressive performance across multiple datasets, achieving 98.0\% accuracy on the Kaggle dataset, 92.81\% accuracy on the Figshare MRI dataset, and 96.20\% accuracy on the gastro dataset. These results demonstrate the model ability to adapt and perform robustly across diverse medical imaging modalities, even in the presence of noisy, ambiguous, and challenging data.

However, while the CASM model exhibits strong performance in most scenarios, there are limitations that need to be considered. Although the model adjusts the temperature based on the uncertainty of the data, this adjustment may not be sufficient in certain extreme cases where the data quality is extremely poor or where the disease presents unusual features that the model has not encountered during training. For example, in the case of MRI images with very low contrast or unusual image artifacts, the context-aware predictor might fail to adapt adequately, leading to reduced performance. The current model reliance on context-based adjustments may be limited when faced with extreme deviations from the types of conditions seen during training, especially in rare or poorly represented diseases.

Moreover, while the CASM model demonstrates robustness across diverse datasets, its ability to handle uncertainty (e.g., when both image quality and disease complexity are low) may still require further refinement. Although it adjusts temperature based on teacher confidence and image quality, there could be scenarios where the model might overfit to specific image characteristics, thus limiting its generalizability in some real-world medical imaging applications.

Furthermore, while the model is trained on well-curated datasets, real-world clinical environments often involve much noisier, unstructured data, which could impact the performance of our approach. The temperature adjustment mechanism, which is based on uncertainty estimation, may not always be capable of filtering out all the noise in the data, potentially leading to incorrect predictions, particularly in cases of data corruption or when images are heavily distorted.

\section{Conclusion and future work}\label{sec6}
This research introduces the CASM, a novel approach within KD framework, aimed at overcoming the challenges of uncertainty and noise in medical image classification. By utilizing context-aware predictor to dynamically adjust the temperature during the distillation process, our model demonstrates superior performance and adaptability across different medical imaging datasets. The integration of ACO further enhances the model by selecting optimal teacher-student pairs, thereby improving performance while reducing computational demands one by one model testing. This combination of context-aware predictor and ACO ensures efficient knowledge transfer, making the model particularly well-suited for real-world applications in the medical field. In comparison to traditional context-unaware-temperature distillation methods, our approach offers significant improvements by adapting to varying levels of uncertainty in medical images. This dynamic adaptation allows the CASM to focus on critical regions of the images, even in the presence of low-quality data, ultimately leading to more reliable predictions. The application of this model to a wide range of datasets, including multi-disease brain tumor detection, MRI, and gastro data, highlights its robustness and potential for deployment in clinical settings. CASM provides a significant contribution to the field of medical image classification by improving the efficiency and accuracy of knowledge transfer, making it a promising tool for addressing the complexities of real-world medical imaging. The model achieved an impressive accuracy of 98.01\% on the Kaggle dataset, 92.81\% on MRI images, and 96.20\% on gastro images, showcasing its capability to handle diverse imaging conditions and effectively learn from ambiguous data. Future work will focus on extending this approach to additional datasets and clinical applications, further solidifying its value in the medical domain.

\section*{Declarations}
\bmhead{Ethics approval and consent to participate}
Approved and Not applicable
\bmhead{Consent for publication}
Not applicable
\bmhead{Data availability}
Data will be made available on request.
\bmhead{Funding}
No funding
\bmhead{Declaration of competing interest}
The authors declare that they have no known competing financial interests or personal relationships that could have appeared to influence the work reported in this paper.
\bmhead{CRediT authorship contribution statement}
Saif Ur Rehman Khan \& Muhammed Nabeel Asim: Conceptualization, Data curation, Methodology, Software, Validation, Writing original draft \& Formal analysis. Sebastian Vollmer: Conceptualization, Funding acquisition, Supervision. Andreas Dengel: review \& editing. 

\bibliography{sn-bibliography}% common bib file


\begin{thebibliography}{55}
\ifx \bisbn   \undefined \def \bisbn  #1{ISBN #1}\fi
\ifx \binits  \undefined \def \binits#1{#1}\fi
\ifx \bauthor  \undefined \def \bauthor#1{#1}\fi
\ifx \batitle  \undefined \def \batitle#1{#1}\fi
\ifx \bjtitle  \undefined \def \bjtitle#1{#1}\fi
\ifx \bvolume  \undefined \def \bvolume#1{\textbf{#1}}\fi
\ifx \byear  \undefined \def \byear#1{#1}\fi
\ifx \bissue  \undefined \def \bissue#1{#1}\fi
\ifx \bfpage  \undefined \def \bfpage#1{#1}\fi
\ifx \blpage  \undefined \def \blpage #1{#1}\fi
\ifx \burl  \undefined \def \burl#1{\textsf{#1}}\fi
\ifx \doiurl  \undefined \def \doiurl#1{\url{https://doi.org/#1}}\fi
\ifx \betal  \undefined \def \betal{\textit{et al.}}\fi
\ifx \binstitute  \undefined \def \binstitute#1{#1}\fi
\ifx \binstitutionaled  \undefined \def \binstitutionaled#1{#1}\fi
\ifx \bctitle  \undefined \def \bctitle#1{#1}\fi
\ifx \beditor  \undefined \def \beditor#1{#1}\fi
\ifx \bpublisher  \undefined \def \bpublisher#1{#1}\fi
\ifx \bbtitle  \undefined \def \bbtitle#1{#1}\fi
\ifx \bedition  \undefined \def \bedition#1{#1}\fi
\ifx \bseriesno  \undefined \def \bseriesno#1{#1}\fi
\ifx \blocation  \undefined \def \blocation#1{#1}\fi
\ifx \bsertitle  \undefined \def \bsertitle#1{#1}\fi
\ifx \bsnm \undefined \def \bsnm#1{#1}\fi
\ifx \bsuffix \undefined \def \bsuffix#1{#1}\fi
\ifx \bparticle \undefined \def \bparticle#1{#1}\fi
\ifx \barticle \undefined \def \barticle#1{#1}\fi
\bibcommenthead
\ifx \bconfdate \undefined \def \bconfdate #1{#1}\fi
\ifx \botherref \undefined \def \botherref #1{#1}\fi
\ifx \url \undefined \def \url#1{\textsf{#1}}\fi
\ifx \bchapter \undefined \def \bchapter#1{#1}\fi
\ifx \bbook \undefined \def \bbook#1{#1}\fi
\ifx \bcomment \undefined \def \bcomment#1{#1}\fi
\ifx \oauthor \undefined \def \oauthor#1{#1}\fi
\ifx \citeauthoryear \undefined \def \citeauthoryear#1{#1}\fi
\ifx \endbibitem  \undefined \def \endbibitem {}\fi
\ifx \bconflocation  \undefined \def \bconflocation#1{#1}\fi
\ifx \arxivurl  \undefined \def \arxivurl#1{\textsf{#1}}\fi
\csname PreBibitemsHook\endcsname

%%% 1
\bibitem[\protect\citeauthoryear{Boutry and et~al.}{2022}]{boutry2022evolution}
\begin{barticle}
\bauthor{\bsnm{Boutry}, \binits{J.}},
\bauthor{\bsnm{al.}}:
\batitle{The evolution and ecology of benign tumors}.
\bjtitle{Biochimica et Biophysica Acta (BBA)-Reviews on Cancer}
\bvolume{1877}(\bissue{1}),
\bfpage{188643}
(\byear{2022})
\end{barticle}
\endbibitem

%%% 2
\bibitem[\protect\citeauthoryear{Khan et~al.}{2025}]{khan2025optimizing}
\begin{bchapter}
\bauthor{\bsnm{Khan}, \binits{Z.}},
\bauthor{\bsnm{Khan}, \binits{S.U.R.}},
\bauthor{\bsnm{Bilal}, \binits{O.}},
\bauthor{\bsnm{Raza}, \binits{A.}},
\bauthor{\bsnm{Ali}, \binits{G.}}:
\bctitle{Optimizing cervical lesion detection using deep learning with particle swarm optimization}.
In: \bbtitle{2025 6th International Conference on Advancements in Computational Sciences (ICACS)},
pp. \bfpage{1}--\blpage{7}
(\byear{2025}).
\bcomment{IEEE}
\end{bchapter}
\endbibitem

%%% 3
\bibitem[\protect\citeauthoryear{Alemany and et~al.}{2021}]{alemany2021late}
\begin{barticle}
\bauthor{\bsnm{Alemany}, \binits{M.}},
\bauthor{\bsnm{al.}}:
\batitle{Late effects of cancer treatment: consequences for long-term brain cancer survivors}.
\bjtitle{Neuro-Oncology Practice}
\bvolume{8}(\bissue{1}),
\bfpage{18}--\blpage{30}
(\byear{2021})
\end{barticle}
\endbibitem

%%% 4
\bibitem[\protect\citeauthoryear{Hekmat et~al.}{2025}]{hekmat2025differential}
\begin{botherref}
\oauthor{\bsnm{Hekmat}, \binits{A.}},
\oauthor{\bsnm{Zuping}, \binits{Z.}},
\oauthor{\bsnm{Bilal}, \binits{O.}},
\oauthor{\bsnm{Khan}, \binits{S.U.R.}}:
Differential evolution-driven optimized ensemble network for brain tumor detection.
International Journal of Machine Learning and Cybernetics,
1--26
(2025)
\end{botherref}
\endbibitem

%%% 5
\bibitem[\protect\citeauthoryear{Rathi and et~al.}{2022}]{rathi2022influence}
\begin{barticle}
\bauthor{\bsnm{Rathi}, \binits{S.}},
\bauthor{\bsnm{al.}}:
\batitle{The influence of the blood–brain barrier in the treatment of brain tumours}.
\bjtitle{Journal of Internal Medicine}
\bvolume{292}(\bissue{1}),
\bfpage{3}--\blpage{30}
(\byear{2022})
\end{barticle}
\endbibitem

%%% 6
\bibitem[\protect\citeauthoryear{Lee}{2021}]{lee2021neurologic}
\begin{barticle}
\bauthor{\bsnm{Lee}, \binits{E.Q.}}:
\batitle{Neurologic complications of cancer therapies}.
\bjtitle{Current Neurology and Neuroscience Reports}
\bvolume{21},
\bfpage{1}--\blpage{12}
(\byear{2021})
\end{barticle}
\endbibitem

%%% 7
\bibitem[\protect\citeauthoryear{Ostrom and et~al.}{2018}]{ostrom2018epidemiology}
\begin{barticle}
\bauthor{\bsnm{Ostrom}, \binits{Q.T.}},
\bauthor{\bsnm{al.}}:
\batitle{Epidemiology of intracranial gliomas}.
\bjtitle{Intracranial Gliomas Part I-Surgery}
\bvolume{30},
\bfpage{1}--\blpage{11}
(\byear{2018})
\end{barticle}
\endbibitem

%%% 8
\bibitem[\protect\citeauthoryear{Swati and et~al.}{2019}]{swati2019content}
\begin{barticle}
\bauthor{\bsnm{Swati}, \binits{Z.N.K.}},
\bauthor{\bsnm{al.}}:
\batitle{Content-based brain tumor retrieval for mr images using transfer learning}.
\bjtitle{IEEE Access}
\bvolume{7},
\bfpage{17809}--\blpage{17822}
(\byear{2019})
\end{barticle}
\endbibitem

%%% 9
\bibitem[\protect\citeauthoryear{Ganesh et~al.}{2024}]{ganesh2024multi}
\begin{botherref}
\oauthor{\bsnm{Ganesh}, \binits{S.}},
\oauthor{\bsnm{Kannadhasan}, \binits{S.}},
\oauthor{\bsnm{Jayachandran}, \binits{A.}}:
Multi class robust brain tumor with hybrid classification using dta algorithm.
Heliyon
\textbf{10}(1)
(2024)
\end{botherref}
\endbibitem

%%% 10
\bibitem[\protect\citeauthoryear{Jyothi and Singh}{2023}]{jyothi2023deep}
\begin{barticle}
\bauthor{\bsnm{Jyothi}, \binits{P.}},
\bauthor{\bsnm{Singh}, \binits{A.R.}}:
\batitle{Deep learning models and traditional automated techniques for brain tumor segmentation in mri: a review}.
\bjtitle{Artificial Intelligence Review}
\bvolume{56}(\bissue{4}),
\bfpage{2923}--\blpage{2969}
(\byear{2023})
\end{barticle}
\endbibitem

%%% 11
\bibitem[\protect\citeauthoryear{Malakouti et~al.}{2024}]{malakouti2024machine}
\begin{botherref}
\oauthor{\bsnm{Malakouti}, \binits{S.M.}},
\oauthor{\bsnm{Menhaj}, \binits{M.B.}},
\oauthor{\bsnm{Suratgar}, \binits{A.A.}}:
Machine learning and transfer learning techniques for accurate brain tumor classification.
Clinical eHealth
(2024)
\end{botherref}
\endbibitem

%%% 12
\bibitem[\protect\citeauthoryear{Saboor and et~al.}{2024}]{saboor2024ddfc}
\begin{barticle}
\bauthor{\bsnm{Saboor}, \binits{A.}},
\bauthor{\bsnm{al.}}:
\batitle{Ddfc: deep learning approach for deep feature extraction and classification of brain tumors using magnetic resonance imaging in e-healthcare system}.
\bjtitle{Scientific Reports}
\bvolume{14}(\bissue{1}),
\bfpage{6425}
(\byear{2024})
\end{barticle}
\endbibitem

%%% 13
\bibitem[\protect\citeauthoryear{Rajput and et~al.}{2024}]{rajput2024transfer}
\begin{barticle}
\bauthor{\bsnm{Rajput}, \binits{I.S.}},
\bauthor{\bsnm{al.}}:
\batitle{A transfer learning-based brain tumor classification using magnetic resonance images}.
\bjtitle{Multimedia Tools and Applications}
\bvolume{83}(\bissue{7}),
\bfpage{20487}--\blpage{20506}
(\byear{2024})
\end{barticle}
\endbibitem

%%% 14
\bibitem[\protect\citeauthoryear{Salih and Abdulazeez}{2024}]{salih2024fusion}
\begin{barticle}
\bauthor{\bsnm{Salih}, \binits{M.S.}},
\bauthor{\bsnm{Abdulazeez}, \binits{A.M.}}:
\batitle{A fusion-based deep approach for enhanced brain tumor classification}.
\bjtitle{Journal of Soft Computing and Data Mining}
\bvolume{5}(\bissue{1}),
\bfpage{183}--\blpage{193}
(\byear{2024})
\end{barticle}
\endbibitem

%%% 15
\bibitem[\protect\citeauthoryear{AG and et~al.}{2024}]{ag2024robust}
\begin{barticle}
\bauthor{\bsnm{AG}, \binits{B.}},
\bauthor{\bsnm{al.}}:
\batitle{Robust brain tumor classification by fusion of deep learning and channel-wise attention mode approach}.
\bjtitle{BMC Medical Imaging}
\bvolume{24}(\bissue{1}),
\bfpage{147}
(\byear{2024})
\end{barticle}
\endbibitem

%%% 16
\bibitem[\protect\citeauthoryear{Zhou and et~al.}{2024}]{zhou2024research}
\begin{botherref}
\oauthor{\bsnm{Zhou}, \binits{Y.}},
\oauthor{\bsnm{al.}}:
Research on multi-scale feature fusion network algorithm based on brain tumor medical image classification.
Computers, Materials \& Continua
\textbf{79}(3)
(2024)
\end{botherref}
\endbibitem

%%% 17
\bibitem[\protect\citeauthoryear{Khan et~al.}{2025}]{khan2025optimized}
\begin{barticle}
\bauthor{\bsnm{Khan}, \binits{S.U.R.}},
\bauthor{\bsnm{Asif}, \binits{S.}},
\bauthor{\bsnm{Zhao}, \binits{M.}},
\bauthor{\bsnm{Zou}, \binits{W.}},
\bauthor{\bsnm{Li}, \binits{Y.}},
\bauthor{\bsnm{Li}, \binits{X.}}:
\batitle{Optimized deep learning model for comprehensive medical image analysis across multiple modalities}.
\bjtitle{Neurocomputing}
\bvolume{619},
\bfpage{129182}
(\byear{2025})
\end{barticle}
\endbibitem

%%% 18
\bibitem[\protect\citeauthoryear{Zhang et~al.}{2024}]{zhang2024cenn}
\begin{barticle}
\bauthor{\bsnm{Zhang}, \binits{H.}},
\bauthor{\bsnm{Huang}, \binits{H.}},
\bauthor{\bsnm{Zhao}, \binits{P.}},
\bauthor{\bsnm{Zhu}, \binits{X.}},
\bauthor{\bsnm{Yu}, \binits{Z.}}:
\batitle{Cenn: Capsule-enhanced neural network with innovative metrics for robust speech emotion recognition}.
\bjtitle{Knowledge-Based Systems}
\bvolume{304},
\bfpage{112499}
(\byear{2024})
\end{barticle}
\endbibitem

%%% 19
\bibitem[\protect\citeauthoryear{Habib et~al.}{2024}]{habib2024comprehensive}
\begin{botherref}
\oauthor{\bsnm{Habib}, \binits{G.}},
\oauthor{\bsnm{Kaleem}, \binits{S.M.}},
\oauthor{\bsnm{Rouf}, \binits{T.}},
\oauthor{\bsnm{Lall}, \binits{B.}}, et al.:
A comprehensive review of knowledge distillation in computer vision.
arXiv preprint arXiv:2404.00936
(2024)
\end{botherref}
\endbibitem

%%% 20
\bibitem[\protect\citeauthoryear{Khan}{2025}]{khan2025multi}
\begin{barticle}
\bauthor{\bsnm{Khan}, \binits{S.U.R.}}:
\batitle{Multi-level feature fusion network for kidney disease detection}.
\bjtitle{Computers in Biology and Medicine}
\bvolume{191},
\bfpage{110214}
(\byear{2025})
\end{barticle}
\endbibitem

%%% 21
\bibitem[\protect\citeauthoryear{Khan et~al.}{2025}]{khan2025ai}
\begin{botherref}
\oauthor{\bsnm{Khan}, \binits{S.U.R.}},
\oauthor{\bsnm{Asim}, \binits{M.N.}},
\oauthor{\bsnm{Vollmer}, \binits{S.}},
\oauthor{\bsnm{Dengel}, \binits{A.}}:
Ai-driven diabetic retinopathy diagnosis enhancement through image processing and salp swarm algorithm-optimized ensemble network.
arXiv preprint arXiv:2503.14209
(2025)
\end{botherref}
\endbibitem

%%% 22
\bibitem[\protect\citeauthoryear{Baloch and Krim}{2007}]{baloch2007flexible}
\begin{barticle}
\bauthor{\bsnm{Baloch}, \binits{S.H.}},
\bauthor{\bsnm{Krim}, \binits{H.}}:
\batitle{Flexible skew-symmetric shape model for shape representation, classification, and sampling}.
\bjtitle{IEEE transactions on image processing}
\bvolume{16}(\bissue{2}),
\bfpage{317}--\blpage{328}
(\byear{2007})
\end{barticle}
\endbibitem

%%% 23
\bibitem[\protect\citeauthoryear{Koitka and Friedrich}{2016}]{koitka2016traditional}
\begin{bchapter}
\bauthor{\bsnm{Koitka}, \binits{S.}},
\bauthor{\bsnm{Friedrich}, \binits{C.M.}}:
\bctitle{Traditional feature engineering and deep learning approaches at medical classification task of imageclef 2016}.
In: \bbtitle{CLEF (Working Notes)}.
\bpublisher{Citeseer}, \blocation{???}
(\byear{2016})
\end{bchapter}
\endbibitem

%%% 24
\bibitem[\protect\citeauthoryear{Shahzad and et~al.}{2024}]{shahzad2024enhancing}
\begin{botherref}
\oauthor{\bsnm{Shahzad}, \binits{I.}},
\oauthor{\bsnm{al.}}:
Enhancing asd classification through hybrid attention-based learning of facial features.
Signal, Image and Video Processing,
1--14
(2024)
\end{botherref}
\endbibitem

%%% 25
\bibitem[\protect\citeauthoryear{Singha et~al.}{2021}]{singha2021deep}
\begin{bbook}
\bauthor{\bsnm{Singha}, \binits{A.}},
\bauthor{\bsnm{Thakur}, \binits{R.S.}},
\bauthor{\bsnm{Patel}, \binits{T.}}:
\bbtitle{Deep Learning Applications in Medical Image Analysis},
pp. \bfpage{293}--\blpage{350}
(\byear{2021})
\end{bbook}
\endbibitem

%%% 26
\bibitem[\protect\citeauthoryear{Ghiasi et~al.}{2022}]{ghiasi2022simple}
\begin{barticle}
\bauthor{\bsnm{Ghiasi}, \binits{M.B.}},
\bauthor{\bsnm{Moezzi}, \binits{M.}},
\bauthor{\bsnm{Kashi}, \binits{A.}}:
\batitle{A simple low phase noise class-f lc oscillator}.
\bjtitle{Circuits, Systems, and Signal Processing}
\bvolume{41}(\bissue{6}),
\bfpage{3041}--\blpage{3049}
(\byear{2022})
\end{barticle}
\endbibitem

%%% 27
\bibitem[\protect\citeauthoryear{Archana and Jeevaraj}{2024}]{archana2024deep}
\begin{barticle}
\bauthor{\bsnm{Archana}, \binits{R.}},
\bauthor{\bsnm{Jeevaraj}, \binits{P.E.}}:
\batitle{Deep learning models for digital image processing: a review}.
\bjtitle{Artificial Intelligence Review}
\bvolume{57}(\bissue{1}),
\bfpage{11}
(\byear{2024})
\end{barticle}
\endbibitem

%%% 28
\bibitem[\protect\citeauthoryear{Zahid and et~al.}{2022}]{zahid2022brainnet}
\begin{barticle}
\bauthor{\bsnm{Zahid}, \binits{U.}},
\bauthor{\bsnm{al.}}:
\batitle{Brainnet: optimal deep learning feature fusion for brain tumor classification}.
\bjtitle{Computational Intelligence and Neuroscience}
\bvolume{2022}(\bissue{1}),
\bfpage{1465173}
(\byear{2022})
\end{barticle}
\endbibitem

%%% 29
\bibitem[\protect\citeauthoryear{Kibriya and et~al.}{2022}]{kibriya2022novel}
\begin{barticle}
\bauthor{\bsnm{Kibriya}, \binits{H.}},
\bauthor{\bsnm{al.}}:
\batitle{A novel and effective brain tumor classification model using deep feature fusion and famous machine learning classifiers}.
\bjtitle{Computational Intelligence and Neuroscience}
\bvolume{2022}(\bissue{1}),
\bfpage{7897669}
(\byear{2022})
\end{barticle}
\endbibitem

%%% 30
\bibitem[\protect\citeauthoryear{Öksüz et~al.}{2022}]{oksuz2022brain}
\begin{barticle}
\bauthor{\bsnm{Öksüz}, \binits{C.}},
\bauthor{\bsnm{Urhan}, \binits{O.}},
\bauthor{\bsnm{Güllü}, \binits{M.K.}}:
\batitle{Brain tumor classification using the fused features extracted from expanded tumor region}.
\bjtitle{Biomedical Signal Processing and Control}
\bvolume{72},
\bfpage{103356}
(\byear{2022})
\end{barticle}
\endbibitem

%%% 31
\bibitem[\protect\citeauthoryear{Khan and et~al.}{2023}]{khan2023deep}
\begin{barticle}
\bauthor{\bsnm{Khan}, \binits{M.A.}},
\bauthor{\bsnm{al.}}:
\batitle{Deep-net: Fine-tuned deep neural network multi-features fusion for brain tumor recognition}.
\bjtitle{Comput Mater Contin}
\bvolume{76}(\bissue{3}),
\bfpage{3029}--\blpage{3047}
(\byear{2023})
\end{barticle}
\endbibitem

%%% 32
\bibitem[\protect\citeauthoryear{Agrawal and et~al.}{2024}]{agrawal2024multifenet}
\begin{barticle}
\bauthor{\bsnm{Agrawal}, \binits{T.}},
\bauthor{\bsnm{al.}}:
\batitle{Multifenet: Multi-scale feature scaling in deep neural network for the brain tumour classification in mri images}.
\bjtitle{International Journal of Imaging Systems and Technology}
\bvolume{34}(\bissue{1}),
\bfpage{22956}
(\byear{2024})
\end{barticle}
\endbibitem

%%% 33
\bibitem[\protect\citeauthoryear{Liu and et~al.}{2024}]{liu2024multi}
\begin{barticle}
\bauthor{\bsnm{Liu}, \binits{X.}},
\bauthor{\bsnm{al.}}:
\batitle{Multi-scale feature fusion for prediction of idh1 mutations in glioma histopathological images}.
\bjtitle{Computer Methods and Programs in Biomedicine}
\bvolume{248},
\bfpage{108116}
(\byear{2024})
\end{barticle}
\endbibitem

%%% 34
\bibitem[\protect\citeauthoryear{Shahin et~al.}{2023}]{shahin2023mbtfcn}
\begin{barticle}
\bauthor{\bsnm{Shahin}, \binits{A.I.}},
\bauthor{\bsnm{Aly}, \binits{W.}},
\bauthor{\bsnm{Aly}, \binits{S.}}:
\batitle{Mbtfcn: A novel modular fully convolutional network for mri brain tumor multi-classification}.
\bjtitle{Expert Systems with Applications}
\bvolume{212},
\bfpage{118776}
(\byear{2023})
\end{barticle}
\endbibitem

%%% 35
\bibitem[\protect\citeauthoryear{Sobhaninia and et~al.}{2023}]{sobhaninia2023brain}
\begin{barticle}
\bauthor{\bsnm{Sobhaninia}, \binits{Z.}},
\bauthor{\bsnm{al.}}:
\batitle{Brain tumor segmentation by cascaded multiscale multitask learning framework based on feature aggregation}.
\bjtitle{Biomedical Signal Processing and Control}
\bvolume{85},
\bfpage{104834}
(\byear{2023})
\end{barticle}
\endbibitem

%%% 36
\bibitem[\protect\citeauthoryear{Miles and Mikolajczyk}{2024}]{miles2024understanding}
\begin{bchapter}
\bauthor{\bsnm{Miles}, \binits{R.}},
\bauthor{\bsnm{Mikolajczyk}, \binits{K.}}:
\bctitle{Understanding the role of the projector in knowledge distillation}.
In: \bbtitle{Proceedings of the AAAI Conference on Artificial Intelligence},
vol. \bseriesno{38},
pp. \bfpage{4233}--\blpage{4241}
(\byear{2024})
\end{bchapter}
\endbibitem

%%% 37
\bibitem[\protect\citeauthoryear{Yuan et~al.}{2024}]{yuan2024student}
\begin{barticle}
\bauthor{\bsnm{Yuan}, \binits{M.}},
\bauthor{\bsnm{Lang}, \binits{B.}},
\bauthor{\bsnm{Quan}, \binits{F.}}:
\batitle{Student-friendly knowledge distillation}.
\bjtitle{Knowledge-Based Systems}
\bvolume{296},
\bfpage{111915}
(\byear{2024})
\end{barticle}
\endbibitem

%%% 38
\bibitem[\protect\citeauthoryear{Ding et~al.}{2024}]{ding2024generous}
\begin{barticle}
\bauthor{\bsnm{Ding}, \binits{Y.}},
\bauthor{\bsnm{Yang}, \binits{G.}},
\bauthor{\bsnm{Yin}, \binits{S.}},
\bauthor{\bsnm{Zhang}, \binits{J.}},
\bauthor{\bsnm{Fang}, \binits{X.}},
\bauthor{\bsnm{Yang}, \binits{W.}}:
\batitle{Generous teacher: Good at distilling knowledge for student learning}.
\bjtitle{Image and Vision Computing}
\bvolume{150},
\bfpage{105199}
(\byear{2024})
\end{barticle}
\endbibitem

%%% 39
\bibitem[\protect\citeauthoryear{Hao et~al.}{2023}]{hao2023revisit}
\begin{barticle}
\bauthor{\bsnm{Hao}, \binits{Z.}},
\bauthor{\bsnm{Guo}, \binits{J.}},
\bauthor{\bsnm{Han}, \binits{K.}},
\bauthor{\bsnm{Hu}, \binits{H.}},
\bauthor{\bsnm{Xu}, \binits{C.}},
\bauthor{\bsnm{Wang}, \binits{Y.}}:
\batitle{Revisit the power of vanilla knowledge distillation: from small scale to large scale}.
\bjtitle{Advances in Neural Information Processing Systems}
\bvolume{36},
\bfpage{10170}--\blpage{10183}
(\byear{2023})
\end{barticle}
\endbibitem

%%% 40
\bibitem[\protect\citeauthoryear{Khan et~al.}{2025}]{khan2025detection}
\begin{barticle}
\bauthor{\bsnm{Khan}, \binits{S.U.R.}},
\bauthor{\bsnm{Zhao}, \binits{M.}},
\bauthor{\bsnm{Li}, \binits{Y.}}:
\batitle{Detection of mri brain tumor using residual skip block based modified mobilenet model}.
\bjtitle{Cluster Computing}
\bvolume{28}(\bissue{4}),
\bfpage{248}
(\byear{2025})
\end{barticle}
\endbibitem

%%% 41
\bibitem[\protect\citeauthoryear{Wu et~al.}{2024}]{wu2024application}
\begin{barticle}
\bauthor{\bsnm{Wu}, \binits{S.}},
\bauthor{\bsnm{Dong}, \binits{A.}},
\bauthor{\bsnm{Li}, \binits{Q.}},
\bauthor{\bsnm{Wei}, \binits{W.}},
\bauthor{\bsnm{Zhang}, \binits{Y.}},
\bauthor{\bsnm{Ye}, \binits{Z.}}:
\batitle{Application of ant colony optimization algorithm based on farthest point optimization and multi-objective strategy in robot path planning}.
\bjtitle{Applied Soft Computing}
\bvolume{167},
\bfpage{112433}
(\byear{2024})
\end{barticle}
\endbibitem

%%% 42
\bibitem[\protect\citeauthoryear{Huang and et~al.}{2017}]{huang2017densely}
\begin{bchapter}
\bauthor{\bsnm{Huang}, \binits{G.}},
\bauthor{\bsnm{al.}}:
\bctitle{Densely connected convolutional networks}.
In: \bbtitle{Proceedings of the IEEE Conference on Computer Vision and Pattern Recognition}
(\byear{2017})
\end{bchapter}
\endbibitem

%%% 43
\bibitem[\protect\citeauthoryear{He et~al.}{2016}]{he2016identity}
\begin{botherref}
\oauthor{\bsnm{He}, \binits{K.}},
\oauthor{\bsnm{Zhang}, \binits{X.}},
\oauthor{\bsnm{Ren}, \binits{S.}},
\oauthor{\bsnm{Sun}, \binits{J.}}:
Identity mappings in deep residual networks.
Proceedings of the IEEE conference on computer vision and pattern recognition,
5987--5995
(2016)
\end{botherref}
\endbibitem

%%% 44
\bibitem[\protect\citeauthoryear{Zhang et~al.}{2025a}]{zhang2025reproducible}
\begin{barticle}
\bauthor{\bsnm{Zhang}, \binits{H.}},
\bauthor{\bsnm{Zhao}, \binits{P.}},
\bauthor{\bsnm{Tang}, \binits{G.}},
\bauthor{\bsnm{Li}, \binits{Z.}},
\bauthor{\bsnm{Yuan}, \binits{Z.}}:
\batitle{Reproducible and generalizable speech emotion recognition via an intelligent fusion network}.
\bjtitle{Biomedical Signal Processing and Control}
\bvolume{109},
\bfpage{107996}
(\byear{2025})
\end{barticle}
\endbibitem

%%% 45
\bibitem[\protect\citeauthoryear{Zhang et~al.}{2025b}]{zhang2025sparse}
\begin{barticle}
\bauthor{\bsnm{Zhang}, \binits{H.}},
\bauthor{\bsnm{Huang}, \binits{H.}},
\bauthor{\bsnm{Zhao}, \binits{P.}},
\bauthor{\bsnm{Yu}, \binits{Z.}}:
\batitle{Sparse temporal aware capsule network for robust speech emotion recognition}.
\bjtitle{Engineering Applications of Artificial Intelligence}
\bvolume{144},
\bfpage{110060}
(\byear{2025})
\end{barticle}
\endbibitem

%%% 46
\bibitem[\protect\citeauthoryear{Mu{\c{c}}a et~al.}{2024}]{mucca2024dimensionality}
\begin{botherref}
\oauthor{\bsnm{Mu{\c{c}}a}, \binits{M.}},
\oauthor{\bsnm{Kap{\c{c}}iu}, \binits{R.}}, et al.:
Dimensionality reduction: A comparative review using rbm, kpca, and t-sne for micro-expressions recognition.
International Journal of Advanced Computer Science \& Applications
\textbf{15}(1)
(2024)
\end{botherref}
\endbibitem

%%% 47
\bibitem[\protect\citeauthoryear{Minarno and et~al.}{2021}]{minarno2021convolutional}
\begin{botherref}
\oauthor{\bsnm{Minarno}, \binits{A.E.}},
\oauthor{\bsnm{al.}}:
Convolutional neural network with hyperparameter tuning for brain tumor classification.
Kinetik: Game Technology, Information System, Computer Network, Computing, Electronics, and Control
(2021)
\end{botherref}
\endbibitem

%%% 48
\bibitem[\protect\citeauthoryear{Srinivas and et~al.}{2022}]{srinivas2022deep}
\begin{barticle}
\bauthor{\bsnm{Srinivas}, \binits{C.}},
\bauthor{\bsnm{al.}}:
\batitle{Deep transfer learning approaches in performance analysis of brain tumor classification using mri images}.
\bjtitle{Journal of Healthcare Engineering}
\bvolume{2022}(\bissue{1}),
\bfpage{3264367}
(\byear{2022})
\end{barticle}
\endbibitem

%%% 49
\bibitem[\protect\citeauthoryear{Tariq and et~al.}{2025}]{tariq2025transforming}
\begin{botherref}
\oauthor{\bsnm{Tariq}, \binits{A.}},
\oauthor{\bsnm{al.}}:
Transforming brain tumor detection empowering multi-class classification with vision transformers and efficientnetv2.
IEEE Access
(2025)
\end{botherref}
\endbibitem

%%% 50
\bibitem[\protect\citeauthoryear{Vijayalakshmi and Anand}{2025}]{vijayalakshmi2025cross}
\begin{botherref}
\oauthor{\bsnm{Vijayalakshmi}, \binits{B.}},
\oauthor{\bsnm{Anand}, \binits{S.}}:
Cross prior bayesian attention with correlated inception and residual learning for brain tumor classification using mr images (cb-cirl net).
Journal of Neuroscience Methods,
110392
(2025)
\end{botherref}
\endbibitem

%%% 51
\bibitem[\protect\citeauthoryear{Chen and et~al.}{2024}]{chen2024robust}
\begin{barticle}
\bauthor{\bsnm{Chen}, \binits{W.}},
\bauthor{\bsnm{al.}}:
\batitle{A robust approach for multi-type classification of brain tumor using deep feature fusion}.
\bjtitle{Frontiers in Neuroscience}
\bvolume{18},
\bfpage{1288274}
(\byear{2024})
\end{barticle}
\endbibitem

%%% 52
\bibitem[\protect\citeauthoryear{Zhao et~al.}{2024}]{zhao2024digan}
\begin{barticle}
\bauthor{\bsnm{Zhao}, \binits{P.}},
\bauthor{\bsnm{Liu}, \binits{X.}},
\bauthor{\bsnm{Yue}, \binits{Z.}},
\bauthor{\bsnm{Zhao}, \binits{Q.}},
\bauthor{\bsnm{Liu}, \binits{X.}},
\bauthor{\bsnm{Deng}, \binits{Y.}},
\bauthor{\bsnm{Wu}, \binits{J.}}:
\batitle{Digan breakthrough: Advancing diabetic data analysis with innovative gan-based imbalance correction techniques}.
\bjtitle{Computer Methods and Programs in Biomedicine Update}
\bvolume{5},
\bfpage{100152}
(\byear{2024})
\end{barticle}
\endbibitem

%%% 53
\bibitem[\protect\citeauthoryear{Khan et~al.}{2025}]{khan2025optimize}
\begin{barticle}
\bauthor{\bsnm{Khan}, \binits{S.U.R.}},
\bauthor{\bsnm{Asif}, \binits{S.}},
\bauthor{\bsnm{Zhao}, \binits{M.}},
\bauthor{\bsnm{Zou}, \binits{W.}},
\bauthor{\bsnm{Li}, \binits{Y.}}:
\batitle{Optimize brain tumor multiclass classification with manta ray foraging and improved residual block techniques}.
\bjtitle{Multimedia Systems}
\bvolume{31}(\bissue{1}),
\bfpage{1}--\blpage{27}
(\byear{2025})
\end{barticle}
\endbibitem

%%% 54
\bibitem[\protect\citeauthoryear{Rimal et~al.}{2024}]{rimal2024accuracy}
\begin{barticle}
\bauthor{\bsnm{Rimal}, \binits{Y.}},
\bauthor{\bsnm{Sharma}, \binits{N.}},
\bauthor{\bsnm{Alsadoon}, \binits{A.}}:
\batitle{The accuracy of machine learning models relies on hyperparameter tuning: student result classification using random forest, randomized search, grid search, bayesian, genetic, and optuna algorithms}.
\bjtitle{Multimedia Tools and Applications}
\bvolume{83}(\bissue{30}),
\bfpage{74349}--\blpage{74364}
(\byear{2024})
\end{barticle}
\endbibitem

%%% 55
\bibitem[\protect\citeauthoryear{Jain et~al.}{2024}]{jain2024exhaustive}
\begin{botherref}
\oauthor{\bsnm{Jain}, \binits{R.}},
\oauthor{\bsnm{Sarvakar}, \binits{K.}},
\oauthor{\bsnm{Patel}, \binits{C.}},
\oauthor{\bsnm{Mishra}, \binits{S.}}:
An exhaustive examination of deep learning algorithms: present patterns and prospects for the future.
GRENZE International Journal of Engineering and Technology (forthcoming)
(2024)
\end{botherref}
\endbibitem

\end{thebibliography}

\end{document}